\documentclass[proceedings]{uai2025arxiv} 
                        
\usepackage[american]{babel}

\usepackage{natbib} 
\usepackage{mathtools} 
\usepackage{booktabs} 
\usepackage{tikz} 

\usepackage{enumitem}
\usepackage{hyperref}
\usepackage{graphicx}

\usepackage{nicefrac}       
\usepackage{microtype}      
\usepackage{float}
\usepackage{multirow}
\usepackage{colortbl}
\usepackage{changepage}
\usepackage{comment}
\usepackage{adjustbox}
\usepackage{array}

\usepackage{xcolor}
\usepackage[utf8]{inputenc}
\usepackage{url}
\usepackage{amsmath,amssymb}
\usepackage{booktabs}

\usepackage{marvosym}

\newcommand{\K}{ \kappa }
\newcommand{\GP}{ \text{GP} }
\newcommand{\X}{ \mathcal{X} }
\renewcommand{\div}{ D }
\newcommand{\D}{ \mathcal{D} }

\newcommand{\fout}{ f^l }

\edef\normalE{\the\mathcode`E}
\begingroup\lccode`~=`E \lowercase{\endgroup
  \def~}{\mathbb{\mathchar\normalE}}
\mathcode`E="8000

\usepackage{authblk}

\newcommand{\std}[1]{{\scriptsize{$\pm #1$}}}
\def\gW{\mathcal{W}}

\title{Revisiting the Equivalence of Bayesian Neural Networks and Gaussian Processes: On the Importance of Learning Activations}

\author[1,2]{Marcin Sendera\textsuperscript{*}}
\author[ ]{Amin Sorkhei}
\author[1]{\href{mailto:<t.kusmierczyk@uj.edu.pl>?Subject=Your UAI 2025 paper}{Tomasz Ku\'smierczyk\textsuperscript{*}}{}}
\affil[1]{%
    Jagiellonian University
}
\affil[2]{%
Mila, Universit\'e de Montr\'eal
}
  
\begin{document}
\maketitle

\begingroup
\renewcommand\thefootnote{\fnsymbol{footnote}}
\footnotetext[1]{Equal contribution to implementation.}
\endgroup

\begin{abstract}
Gaussian Processes (GPs) provide a convenient framework for specifying function-space priors, making them a natural choice for modeling uncertainty. In contrast, Bayesian Neural Networks (BNNs) offer greater scalability and extendability but lack the advantageous properties of GPs. This motivates the development of BNNs capable of replicating GP-like behavior. However, existing solutions are either limited to specific GP kernels or rely on heuristics.

We demonstrate that trainable activations are crucial for effective mapping of GP priors to wide BNNs. Specifically, we leverage the closed-form 2-Wasserstein distance for efficient gradient-based optimization of reparameterized priors and activations. Beyond learned activations, we also introduce trainable periodic activations that ensure global stationarity by design, and functional priors conditioned on GP hyperparameters to allow efficient model selection.

Empirically, our method consistently outperforms existing approaches or matches performance of the heuristic methods, while offering stronger theoretical foundations.
\end{abstract}

\section{Introduction}
\label{sec:intro}

Function-space priors for BNNs offer a better way of specifying beliefs on data modeled by the model. Unlike priors on model parameters (i.e. weights), which lack interpretability and are hard to specify, they ultimately lead to more intuitive and meaningful representation of prior knowledge~\citep{sun2018functional}.
Function-space priors can be conveniently specified in terms of  GPs. Instead of specifying complex distributions over model parameters, GPs allow for defining priors over entire functions, through the choice of kernel capturing requirements such as smoothness or periodicity. 

Finding a GP that matches a wide BNN is straightforward. A classic result by \cite{neal96a, williams1996computing}, later extended to deep neural networks by \cite{lee2017deep, matthews2018gaussian}, shows that wide layers in neural networks behave \textit{a priori} like GPs. 
This is achieved by identifying GP kernels matching the covariances of (B)NNs.

In contrast, finding BNNs that exhibit the desired behavior of GPs is a notoriously difficult problem, with solutions or approximations existing for only a few GP kernels. For example, \cite{meronen2020stationary} recently found a solution for the popular Matérn kernel using the Wiener-Khinchin theorem. However, the proposed BNN activation assumes binary white noise priors on the model weights, which severely hinders posterior learning. Consequently, an approximate solution with Gaussian priors must be used, ultimately leading to suboptimal performance, as we demonstrate in Sec.~\ref{sec:closed_form}.

The challenge of imposing function-space \textit{a priori} behavior on BNNs previously has been addressed using gradient-based optimization. In particular, \cite{flam2017mapping}, \cite{flam2018characterizing}, and especially \cite{tran2022all} attempted to solve this by learning priors on parameters. However, to achieve sufficient fidelity with simple priors (Gaussian or hierarchical), their approach requires deep networks, which presents a challenge for posterior learning due to the lack of effective algorithms for learning posteriors in deep and wide BNNs. On the other hand, by using normalizing flows to model weight priors they often can match target GP prior (see Sec.~\ref{sec:tran}). Nevertheless, imposing different priors for each weight invalidates the theoretical assumptions regarding BNN convergence, rendering this approach merely a heuristic.

\begin{figure}
    \centering
    \includegraphics[width=0.75\linewidth]{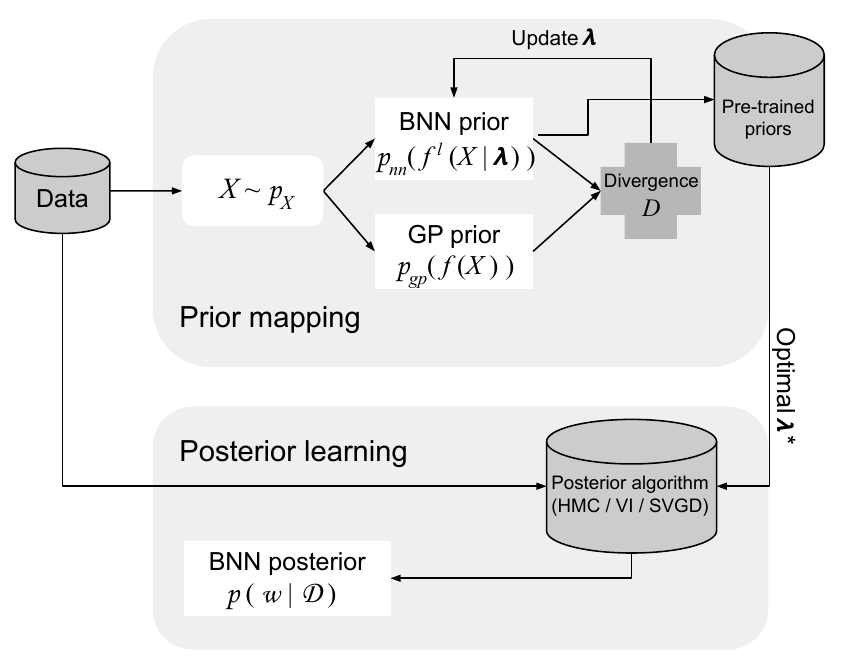}
    \caption{Schematic view of our approach.}
    \label{fig:gp2bnn}
\end{figure}

Fitting only weight (and biases) priors is insufficient for matching BNN and GP priors. 
Hence, we propose learning parametric activations as well.
We draw inspiration from previous work demonstrating that function-space priors can be adjusted by altering activations \citep{neal96a} and also from research utilizing activations for improved uncertainty estimation \citep{morales2020activation,Postels2020HiddenUncertainty}.
In fact, the same GP-like behavior can often be realized by different combinations of parameter priors and activations, as seen in the BNN covariance formulation (Eq.~\ref{eq:layer_cov}) and learning both jointly offers a greater flexibility.
In particular,
our method (see Fig.~\ref{fig:gp2bnn}) transfers priors from GPs to BNNs by matching their function-space distributions using the closed-form 2-Wasserstein loss and by learning activations in addition to priors on weights. It achieves faithful function-space priors using just shallow BNNs. 
In contrast to~\cite{tran2022all}, it is backed by theoretical results ensuring that the learned BNNs asymptotically converge to GPs, and compared to \cite{meronen2020stationary}, it can work with arbitrary kernels. Finally, it relies on simple weight priors, enabling efficient posterior inference.

Within the proposed framework, we propose two novel extensions. First, by conditioning the BNN's functional priors (weights and activation) on GP hyperparameters -- implemented via hypernetworks -- we facilitate efficient model selection, e.g., allow prior hyperparameters optimization. Second, building on the result of \cite{meronen2021periodic} that periodic activations induce stationarity, we propose trainable periodic activations and show how such activations can be learned in practice, ensuring global stationarity in BNNs.

Our contributions extend the existing literature on BNN–GP equivalence in several ways: (1) we introduce a practical approach for imposing GP-like function-space priors on BNNs; (2) we employ hypernetworks to condition these priors on GP hyperparameters; (3) we introduce trainable periodic activations; and (4) we empirically demonstrate that even shallow BNNs can achieve the desired properties, offering an alternative to baselines using deep models.
Note that although parametric activations and hypernetworks have been used previously, we are the first to employ and combine them for this particular task. Moreover, although our design of trainable periodic activations  builds on the previous result that periodic activations induce stationarity, it is novel. Finally, despite our focus on the specification of priors, we also validate our pre-trained priors by learning posteriors, employing both Hamiltonian Monte Carlo (HMC) and scalable Stochastic Variational Inference (SVI) to demonstrate empirical improvements in predictive performance and uncertainty quantification.

In the Appendix, the results presented in the paper are supplemented with a description of related work, 
a discussion of computational advantages of our approach,
a detailed overview of the experimental settings, and additional plots and numerical data.

Our code we made publicly available~\footnote{\href{https://github.com/gmum/bnn-functional-priors}{https://github.com/gmum/bnn-functional-priors}}.

\section{Preliminaries}
\label{sec:preliminaries}

\label{sec:convergence}
Behavior akin to GPs has been identified in wide neural networks with a single hidden layer by \cite{neal96a,williams1996computing} and later generalized for deep architectures by \cite{lee2017deep,matthews2018gaussian} who considered stacking multiple such the wide layers. \\
Let's consider a network defined as:
\begin{align*}
f_i^{0}(x) &= \sum_j^{I} w^{0}_{ij} x_j + b^{0}_i, \quad i=1, \dots, H_0; \\
\fout(x) &= \sum_j^{H_{0}} w^{l}_{ij} \phi(f^{0}_j(x)) + b^{l}
\end{align*}
where \(x\) denotes an \(I\)-dimensional input, and \(f^l(x)\) represents output at the \(l\)-th layer.
In a BNN, \(z_{ij}(x) = w^{l}_{ij} \phi(f^{0}_j(x))\) is a random variable. Assuming that the weights \(w_{ij}\) share a common prior and also the biases \(b_i\) have independent but common prior, the output \(\fout(x)\) becomes a sum of independent and identically distributed (i.i.d.) random variables. By the Central Limit Theorem, this results in \(\fout(x)\) converging to a Gaussian distribution, implying that the BNN converges to a GP in the limit of infinite width.

The covariance between two inputs, \(x\) and \(x'\), for the BNN at the (output) layer \(l\) is given by:
\begin{align}\small
\text{Cov}(\fout(x), \fout(x')) &= {\sigma^l_b}^2 + {\sigma^l_w}^2\mathbb{E}_{f^{0}_j}[\phi(f^{0}_j(x)) \phi(f^{0}_j(x'))], \label{eq:layer_cov}
\end{align}
where \(\sigma^l_b\) and \(\sigma^l_w\) are respectively the variances of the biases and weights at layer \(l\) and the expectation is taken over distributions for $w^0$ and $b^0$. 
Hence, the BNN functional prior corresponds to a GP with kernel \(\mathcal{\kappa}_f^l(x, x') = \text{Cov}(\fout(x), \fout(x'))\).

In this work, 
we assume zero-centered GPs. Therefore, for the distributions on weights and biases, we take \(\mathbb{E}[w] = 0\) and \(\mathbb{E}[b] = 0\). To ensure that the variance remains stable as the layer width increases 
we scale
\(\text{Var}[w^l_{ij}] = \frac{{\sigma^l_w}^2}{H_0}\).


\section{Method}

\subsection{Transfer of Priors from GP to BNN}
\label{sec:inverse}

Finding a GP equivalent to a pre-specified BNN can be easily done by a Monte Carlo estimate of Eq.~\ref{eq:layer_cov}. However, we focus on the significantly more challenging inverse problem of imposing behavior akin to a $\GP$ on a BNN. The task is to \emph{identify appropriate priors on $w^l$, $b^l$, $w^{0}$, $b^{0}$, and the activation $\phi$ in Eq.~\ref{eq:layer_cov} so that the covariance induced by the BNN aligns with the desired GP kernel $\kappa$}. Ideally, we aim to have $\fout \sim \GP(0, \K)|_{\X}$, meaning that the distribution of the BNN  outputs (pre-likelihood) would closely match that of the GP across the input space $\X$, i.e., $p_{nn}(\fout) = p_{gp}(\fout)$. 

In practice, however, we settle for approximate matching over finite index sets~\citep{shi2018spectral,sun2018functional}, where the densities over BNN and GP outputs should approximately align as $p_{nn}(\fout(X)) \approx p_{gp}(\fout(X))$. $X \sim p_X$ ($X \in \X$) can be understood as sampling sets of inputs $\{x\}$ at which we observe the functions. 

The matching between BNN and GP, we formulate as an optimization task:
{\small
\[
p^*_{nn} = \text{argmin}_{p_{nn}} \frac{1}{S} \sum_{X \sim p_X} \div(p_{nn}(\fout(X)), p_{gp}(\fout(X))),
\]
}
where $\div$ is an \emph{arbitrary}, differentiable divergence measure, and $S$ is a number of samples. In practice, we perform gradient-based optimization with $S=1$ sample used for each step.


\subsection{Differentiable Priors and Activations}

The results presented in Sec.~\ref{sec:convergence} hold for distributions with finite variances and light tails. We use the basic zero-centered factorized Gaussians with learned variances, e.g., $p(w|\sigma^2_w)$ and $p(b|\sigma^2_b)$, as prior distributions on model weights and biases. To enable gradient-based optimization, such as gradient propagation through weight and bias samples, we reparameterize the distributions using the reparameterization trick~\citep{kingma2014auto}. Additionally, \emph{we propose to employ parametric and differentiable activations $\phi(\cdot|\eta)$ to enhance the network’s flexibility.} This previously overlooked idea allows us to {model complex functional priors within a single hidden layer of a BNN, even with simple prior distributions on weights and biases}.

Given the reparameterized distributions on weights and biases, as well as the parametric activation, $p_{nn}$ prior is fully characterized by the parameters $\lambda = \{\sigma^0_b, \sigma^0_w, \sigma^l_b, \sigma^l_w, \eta\}$. The optimization objective can be then expressed as:
{\small
\begin{equation}
    \lambda^* = \text{argmin}_{\lambda} \frac{1}{S} \sum_{X \sim p_X} \div\left(p_{nn}\left(\fout(X|\lambda)\right), p_{gp}\left(\fout(X)\right)\right), 
    \label{eq:opt}
\end{equation}
}
where $S$ denotes the number of input samples.
Eq.~\ref{eq:opt} is solved through gradient-based optimization w.r.t.~$\lambda$. 

Eq.~\ref{eq:opt} requires deciding on the divergence measure $\div$ and a model for the activation function $\phi$.
The problem of learning activations for neural networks involves designing functions that enable networks to capture complex and non-linear relationships effectively. In principle, any function $\phi: \mathcal{R} \rightarrow \mathcal{R}$ can serve as an activation, but the choice of $\phi$ significantly impacts both the network's expressiveness and its training dynamics. 

We explore several models for $\phi$, including Rational (Pade) activations~\citep{molina2019activation}, Piecewise Linear (PWL) activations\footnote{\url{https://pypi.org/project/torchpwl/}}, and activations implemented as a neural network with a single narrow (with only 5 neurons) hidden layer, using ReLU/SiLU own activations. These functions are computationally efficient and introduce desirable non-linear properties, striking a balance between efficiency and the capacity to model intricate patterns.

Our choice of a NN-based activation with a single hidden layer consisting of 5-10 neurons and using ReLU/SiLU activations was informed by empirical studies (see Fig.~\ref{fig:matching_priors} and Tables~\ref{tab:prior_matching_time1}~and~\ref{tab:prior_matching_time2} in Section~\ref{sec:wallclock}) comparing various learnable functions. These experiments, including ablations on the number of layers and width of the NN activations, show that this configuration offers a good balance of flexibility, fast convergence, and quality of prior fit. While deeper or wider NN activations can marginally improve fidelity in some cases, they also increase convergence times and can make optimization harder.

\subsection{Periodic Activations For Stationary~GPs}
\label{sec:periodic}
The optimization in Eq.~\ref{eq:opt} acts on range of inputs $X$, where the BNN's behavior increasingly resembles the GP as training progresses. However, there are no guarantees about the BNN's behavior outside this subset, particularly far from it. This limitation is exacerbated by \emph{local stationarity} in BNNs: although GPs with stationary kernels are \emph{globally} stationary, BNNs usually only exhibit stationarity within a limited input range. Outside this range, the covariance induced by the BNN may diverge from the desired GP behavior.
This localized training can cause uncertainty quantification issues. In regions far from the data, a BNN may either become overly confident or overly uncertain, depending on the priors and the mismatch between the BNN's architecture and the GP's true behavior. It is challenging to enforce global properties, like stationarity or smoothness, across the entire input space, as the BNN’s learned covariance structure is overly dependent on the inputs' range. 

We address the local stationarity challenge by \emph{introducing trainable activations designed to exhibit periodic behavior.}
This approach is motivated by the result of \cite{meronen2021periodic}, who showed that periodic activation functions induce stationarity in BNNs. 
Based on this theoretical result, we propose a practical solution.
Our proposed activations  
{\small
$$
\phi(x|\psi, A) = \sum_{i=1}^K A_i \cos(2\pi \psi_i x) +  \sum_{j=1}^K A_j \sin(2\pi \psi_j x)
$$
}
are inspired by Fourier analysis and can be trained to fit a desired functional prior. They rely on the variational parameters $\eta = \{\psi_i, A_i, \psi_j, A_j\}$ steering respectively frequencies and amplitudes.  
For experiments, we used $K=5$ components.


\subsection{Conditional Priors and Activations}
\label{sec:conditioning}

Researchers previously, when fitting BNN priors, were limited to upfront fixing hyperparameters (such as lengthscale) of target GPs or alternatively, they would employ workarounds like hierarchical kernels with hyperparameters sampled from hyperpriors (see, e.g.,~\cite{tran2022all}).
Instead, \emph{we propose to condition the priors on weights, biases, and activations in BNNs on the hyperparameters} of a GP kernel. By incorporating GP hyperparameters directly into BNNs, we \emph{bridge the equivalence gap between GPs and BNNs}, finally enabling BNNs to fully replicate the behavior of GPs. In particular, integrating hyperparameters directly within BNNs allows for their optimization, for example, using marginal likelihood (evidence). This facilitates effective model selection, which is a significant novelty compared to previous approaches.

Although our work focuses on conditioning on GP hyperparameters, the proposed conditioning framework is however more general and could be extended to priors dependent on other factors. For example, conditioning on input data could enhance model robustness against input range shifts by allowing the BNN to adapt its priors to the input range. This adaptation could alleviate issues with the locality of the matching BNN to a GP, as discussed in Section\ref{sec:periodic}. 

The conditioning we implemented using hypernetworks~\citep{HyperNetworks,chauhan2024brief}, as $[\sigma, \eta] := \text{hnet}(\gamma | \theta)$, where $\sigma$ and $\eta$ are the sets of parameters for the priors on weights and activations, respectively. Here, $\gamma$ represents the set of conditioning hyperparameters, which are transformed by the hypernetwork $\text{hnet}$. The hypernetwork has its own parameters $\theta$, which are now optimized in Eq.~\ref{eq:opt} as $\lambda = \{\theta\}$ instead of $\{\sigma, \eta\}$. For a fixed architecture of $\text{hnet}$, $\sigma$ and $\eta$ are now fully determined by $\gamma$ along with $\theta$.

We are the first to test hypernetworks for generating activation parameters and using them for conditioning with hyperparameters. This poses a technical challenge in network design. For example, no architectures other than those based on RBFs showed promising results. Ultimately, we employed an MLP with three hidden layers, each using RBF activations. On top of that, we applied separate linear layers to map to the appropriate outputs: one for each of the prior variances $\sigma$ and one for the parameters $\eta$ of the trainable activation.

\subsection{Loss}
\label{sec:loss}
The loss $D$ measures the divergence between two distributions over functions. While $p_{gp}$ can be sampled and evaluated (for fixed inputs GPs act like Gaussians), $p_{nn}$ is implicitly defined by a BNN, meaning it can be sampled from, but not evaluated directly. Due to the implicit nature of $p_{nn}$, $D$ must be specified for samples $\{\fout\}$ and approximated using Monte Carlo. 

We discovered that the standard losses fail for our task. For example, estimating the empirical entropy term ($\int p_{nn}(f) \log\left( p_{nn}(f) \right) df$) for KL presents a numerical challenge. Consequently, we followed \cite{tran2022all} and relied on the Wasserstein distance instead (see \cite{tran2022all} for details):
{\small
\begin{align}
& D = \biggl(\inf_{\varsigma \in \Gamma(p_{nn}, p_{gp})} \int_{\mathcal{F} \times \mathcal{F}} d({f}, {f'})^{p} \varsigma ({f}, {f'}) d{f}d{f'}\biggr)^{1/p} \nonumber \\
& = \sup_{|\Psi|_L\leq 1} E_{p_{nn}} [\Psi(f)] - E_{p_{gp}} [\Psi(f)]
\label{eq:wasser_opt}
\end{align}
}

However, unlike \cite{tran2022all}, we propose to use the 2-Wasserstein metric, which for Multivariate Gaussians (applicable in the case of GPs and wide BNNs -- at least approximately) has a \emph{closed-form solution} \citep{NIPS2017_7a006957}:
{\small
\begin{align*}
D = ||\mu_1-\mu_2||_2^2 + \text{Tr}\left(\Sigma_1+\Sigma_2-2\sqrt{\sqrt{\Sigma_1}\Sigma_2\sqrt{\Sigma_1}}\right),
\end{align*}
}

where $\mu_{1/2}$, $\Sigma_{1/2}$ are respectively expectations and covariance matrices estimated for $p_{nn}$ and $p_{gp}$ from samples $\{\fout(X)\}$ (we used $512$ or $1024$ reparameterized samples) evaluated for inputs $X$.
Not only we avoid the internal optimisation due to $\sup_{|\psi|}$, but additionally
$D$ can be efficiently computed based on results by \cite{buzuti2023frechet}.
For experiments, we report numerical values of $D$ normalized by number of elements in $X$.

\subsection{Output Structure}
The optimization objective in Eq.~\ref{eq:opt} is defined over functions $f$. The functions are passed through likelihoods to form model outputs $y$. For regression tasks, a Gaussian likelihood is typically used; for binary classification, Bernoulli; and for multiclass classification, a Categorical likelihood with multivariate $f$ transformed via softmax. Note that the \emph{prior transfer is independent of a likelihood}, meaning it does not rely on how the latent functions $f$ relate to the outputs $y$. Then, as explained in Sec.~\ref{sec:loss}, the optimization can be performed efficiently between Gaussians. If these assumptions were not satisfied—e.g., if the desired priors are not expressible via a GP -- we can revert to the nested optimization for $|\Psi|_L$ as implied by Eq.~\ref{eq:wasser_opt}. Overall, we explain our method in terms of GPs and for brevity, we denote the target functional priors by $p_{gp}$. However, it is important to note that the method is \emph{applicable to arbitrarily specified functional priors}.

For completeness, we follow to briefly discuss also the transfer of priors from GPs to BNNs for multivariate/multi-class-output settings; however, such extensions fall outside of the scope of this paper.
In particular, the multivariate models~\citep{rasmussen2005,alvarez2012kernels} can be approached using various architectural strategies, depending on the nature of the dependencies among the output dimensions. One approach is to construct independent BNNs for each output dimension, \emph{stacking them side-by-side}, where each BNN is pretrained separately with priors obtained from independent single-output GPs. This approach is appropriate when output dimensions are assumed to be independent, leading to an ensemble of independently trained BNNs. Alternatively, \emph{shared hidden layer} BNN can be used, where a common hidden representation is shared across all output dimensions, thereby capturing potential dependencies between outputs. The shared hidden layer structure reflects the idea of a multi-output or multi-task GP, where dependencies among outputs can be modeled explicitly using coregionalization techniques~\citep{goovaerts1997geostatistics,bonilla2008multi}. The shared hidden layer BNN can thus be endowed with priors derived from these multi-output GPs, ensuring that both spatial and output correlations are incorporated into the BNN model.
%


\subsection{Optimisation Challenges}
\label{sec:opt_challenges}

The problem of finding BNNs behaving like GPs (e.g. inverting covariance equation) in \emph{unidenfiable}. 
There exist multiple solutions assuring similar quality of the final match. Activations $\phi$ solving Eq.\eqref{eq:opt} are not unique. For example:
\begin{itemize}[noitemsep]
\item  The solutions are symmetric w.r.t activity values (y-axis), i.e., for $\phi'(f) = -\phi(f)$, values of covariance given by Eq.\eqref{eq:layer_cov} are not changed, simply because $(-1)^2 = 1$.
\item For $p(f_i^{0}(x))$ symmetric around $0$ (for example, Gaussians), activations with flipped arguments $\phi'(f) = \phi(-f)$ result in the same covariances. 
\item Scaling activations $\phi'(f) = \alpha \phi(f)$ leads to the same covariances as scaling variances of the output weights as $(\sigma^l_w)' = \alpha \cdot \sigma^l_w$ 
\end{itemize}

Given a sufficiently flexible model (like a neural network itself), one can learn to approximate any target activation function to an arbitrary degree of accuracy on a compact domain. This is in line with \emph{the universal approximation theorem}. However, in practice, there are multiple limitations and challenges. A model may require an impractically large number of parameters to approximate certain complex functions to a desired level of accuracy. More complex models may be harder to fit and require more training data. Some functions might require high numerical precision to be approximated effectively, and even if a model can fit a target activation function on a compact set, it might not generalize well outside the training domain. 
Overall, multiple factors may lead gradient-based optimization to fail for our task. However, such the \emph{poor outcomes will be reflected in low values of the final loss} (Eq.~\ref{eq:wasser_opt}). A practical remedy is to rerun optimization, once poorly performing outliers are noticed.

\section{Findings}

\begin{table*}[t]
\caption{\label{tab:uci_boston}Results for UCI regression task (Boston dataset) with prior transferred from a GP; comparison between the baseline (using a deep BNN; \citep{tran2022all}) and ours (single hidden layer BNN with varying widths). \emph{Periodic} activation (Sec.~\ref{sec:periodic}) and an activation realized by a NN 
with SiLU own activation (\emph{default}) were used.}
    \centering
    \resizebox{0.98\linewidth}{!}{
    \begin{tabular}{lccccccc}
        \toprule
        Method $\rightarrow$
        & \multicolumn{2}{c}{our ($width = 128$, \textit{Periodic act.})}
        & \multicolumn{2}{c}{our ($width = 128$)} 
        & \multicolumn{2}{c}{our ($width = 1000$)} 
        & \multicolumn{1}{c}{Baseline} 
        \\
        \cmidrule(lr){2-3}\cmidrule(lr){4-5}\cmidrule(lr){6-7}\cmidrule(lr){8-8}
        Metric $\downarrow$
        & \multicolumn{1}{c}{- regularization} 
        & \multicolumn{1}{c}{+ regularization}
        & \multicolumn{1}{c}{- regularization} 
        & \multicolumn{1}{c}{+ regularization}
        & \multicolumn{1}{c}{- regularization}
        & \multicolumn{1}{c}{+ regularization} 
        & \multicolumn{1}{c}{---}
        \\
        \midrule
        RMSE  & 2.9067\std{0.8257}  & 2.8967\std{0.8258}  & 2.8643\std{0.8386}  & \textbf{2.8348}\std{\textbf{0.8371}}  & 2.9189\std{0.8408}   & 3.0059\std{0.9068}  & 2.8402\std{0.8986} \\
        NLL & 2.5057\std{0.1870}  & 2.5072\std{0.1904}  & 2.4937\std{0.1798} & 2.4862\std{0.1763}  & 2.5122\std{0.1871} & 2.5971\std{0.1206}  &  \textbf{2.4778}\std{\textbf{0.1481}}\\
        \bottomrule
    \end{tabular}
    }
\end{table*}

\begin{figure}[t]
\centering
\begin{minipage}{0.325\textwidth}
    \centering
    {\small input space dimensionality = 1} \\
    \includegraphics[width=0.85\textwidth]{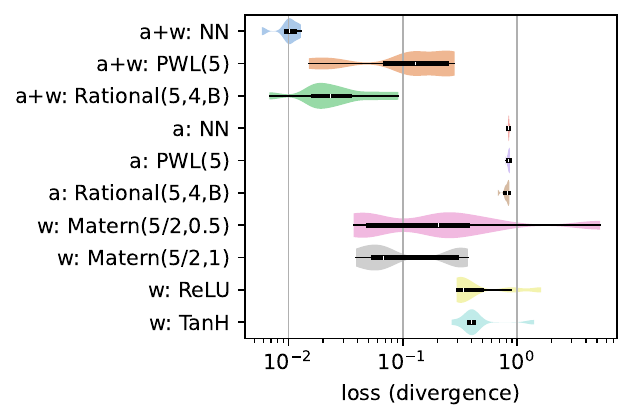} 
    \\
    \end{minipage}%
\hfill
\begin{minipage}{0.325\textwidth}
    \centering
    {\small input space dimensionality = 16} \\
    \includegraphics[width=0.85\textwidth]{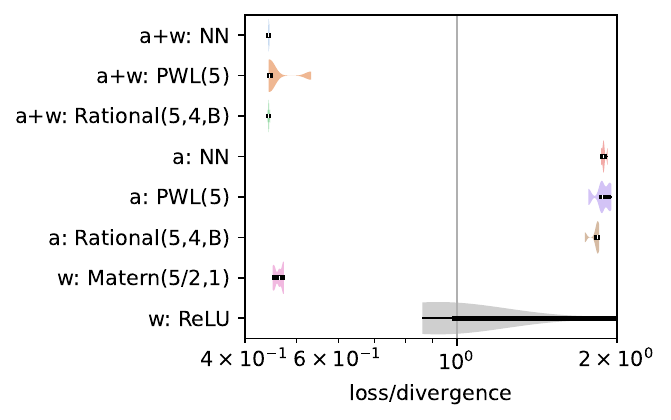} \\
    \end{minipage}
\caption{
Quality of matching BNNs to the prior of a GP with a Mat\'ern kernel ($\nu=5/2$, $\ell=1$) for 1D (top) and 16D inputs (bottom). We evaluate models with trained parameter priors (denoted by \textit{w}), activations (denoted by \textit{a}), and both (denoted by \textit{a+w}). Each label specifies whether a fixed activation (e.g., ReLU) or a specific activation model (e.g., Rational) was employed. Gaussian parameter priors  were used by default. If not trained, we set variances to $1.$, and for the hidden layer, we normalized the variance by its width. The label \textit{w: Mat\'ern} refers to a BNN with the closed-form (fixed) activation as derived by \cite{meronen2020stationary}.
}
\label{fig:matching_priors}
\end{figure}

\subsection{Does Learning Activations Improve Learning of Function-space Priors?}

To map GP function-space priors to a BNN, we can: 1) train its priors on parameters, 2) train both priors and activation, or 3) learn just the activation function. We empirically show that learning activations in addition to priors provide better solutions to the considered problem. 

Fig.~\ref{fig:matching_priors} illustrates the results of an extensive study in which we compare the quality of the GP prior fit for various models of parameter priors and activations.
The target was \(\text{GP}(0, \text{Mat\'ern}(\nu=5/2, \textit{l}=1))\).
For each configuration, we conducted several training iterations and measured the final (converged) loss multiple times. 
We observe that learning activations alone is not sufficient for good fits, but when combined with learning priors, it significantly improves the results.

\begin{figure}[t]
\centering
\includegraphics[width=0.47\textwidth]
{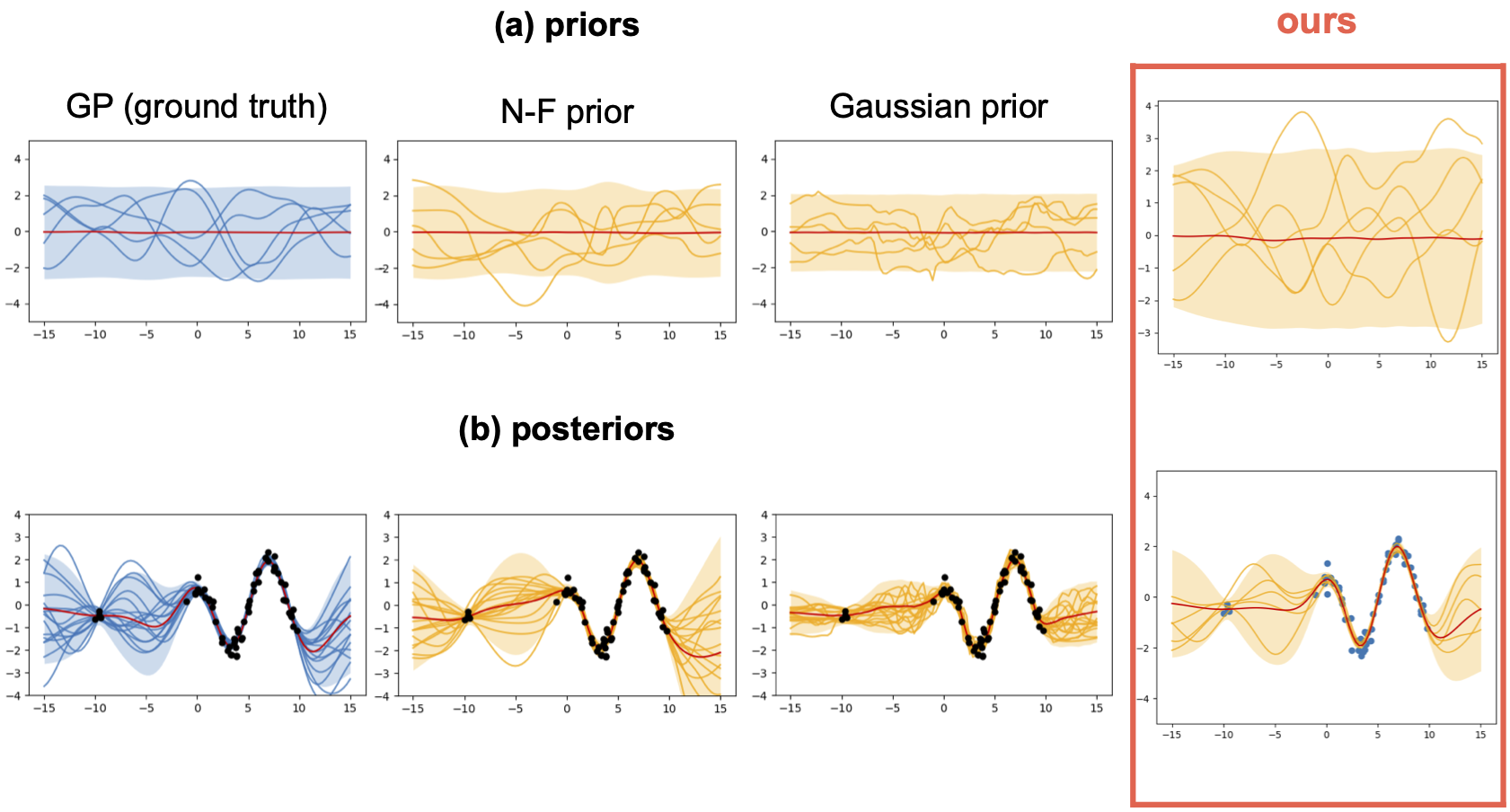} \\
\caption{
Prior \textbf{(a)} and posterior \textbf{(b)} predictive distributions for a BNN with trained parameters priors and activations (ours; 4th column), and for \cite{tran2022all} approach with different prior realizations (Gaussian (3 hidden layers; 3rd column) and NF (2 hidden layers; 2nd). The first column illustrates the ground truth (GP). Numerical results complementing the figures we provide in the Appendix.
}
\label{fig:1d_regression}
\end{figure}

Fig.~\ref{fig:matching_periodic_priors} in the Appendix presents the results of a similar experiment conducted for a GP with the Periodic kernel ($\ell=1$, $p=1$). The best performance is observed for the activation introduced in Sec.~\ref{sec:periodic}, however, increasing the number of layers or neurons can also improve the fit for the activation modeled by an MLP.

We supplement the figure with additional plots showing samples from runs where poor convergence was observed. As explained in Sec.~\ref{sec:opt_challenges}, due to unfortunate initialization or the insufficient modeling capacity of $\phi$, gradient-based optimization may fail to capture a GP functional prior. In such cases, the optimization gets stuck in a local optimum, and the observed final loss differs significantly from the values obtained in other scenarios.


\subsection{Do Expressive Function-space Priors Require Networks to be Deep?}
\label{sec:tran}

Function-space priors require expressive models, which can be achieved either  through a multi-layer neural network architecture or with a shallow BNN with more flexible (\textit{e.g.,} learnable) activations. \citep{tran2022all} explored the former, focusing on multi-layer networks. Instead of considering BNNs with wide layers -- where the theoretical results in Sec.~\ref{sec:preliminaries} apply -- they \emph{postulated} that a deep network can model a functional prior after tuning the network's parameter priors to match a target functional prior.

\citep{tran2022all} investigated several types of priors for BNN weights, including Gaussian, hierarchical, and Normalizing Flow (NF)-based priors. They argue that only the NF prior is flexible enough to capture complex distributions. While NFs~\citep{rezende2015variational} can model intricate priors across all weights, by assigning a distinct prior to each weight they violate the assumptions of the Central Limit Theorem (CLT), preventing the BNN from converging to a GP (\textit{see} Sec.~\ref{sec:preliminaries}). 
Although \citep{tran2022all} achieved their best results using this heuristic NF prior, among the considered priors, the theoretical guarantees hold only for the Gaussian and hierarchical priors, and only in the context of wide BNNs.

Beyond the lack of theoretical justification, using complex priors such as NFs or requiring deep networks introduces additional challenges in posterior inference. MCMC-based methods~\citep{chen2014stochastic,SMC} become computationally expensive, while approximate inference techniques~\citep{hoffman2013stochastic,ritter2018scalable} may lack expressiveness.

\begin{figure}[h]
    \centering
    \includegraphics[width=0.45\linewidth]{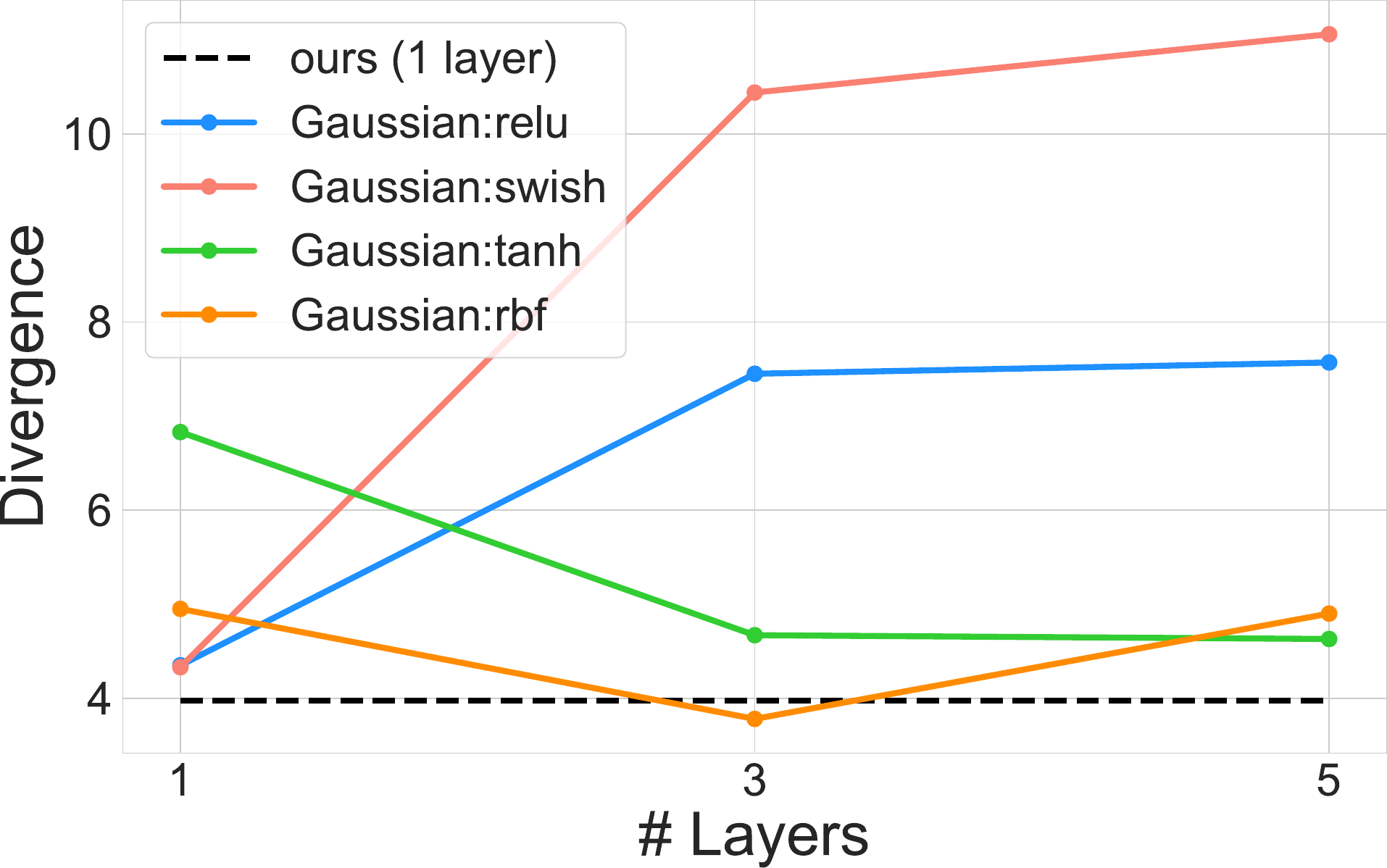}
    \includegraphics[width=0.45\linewidth]{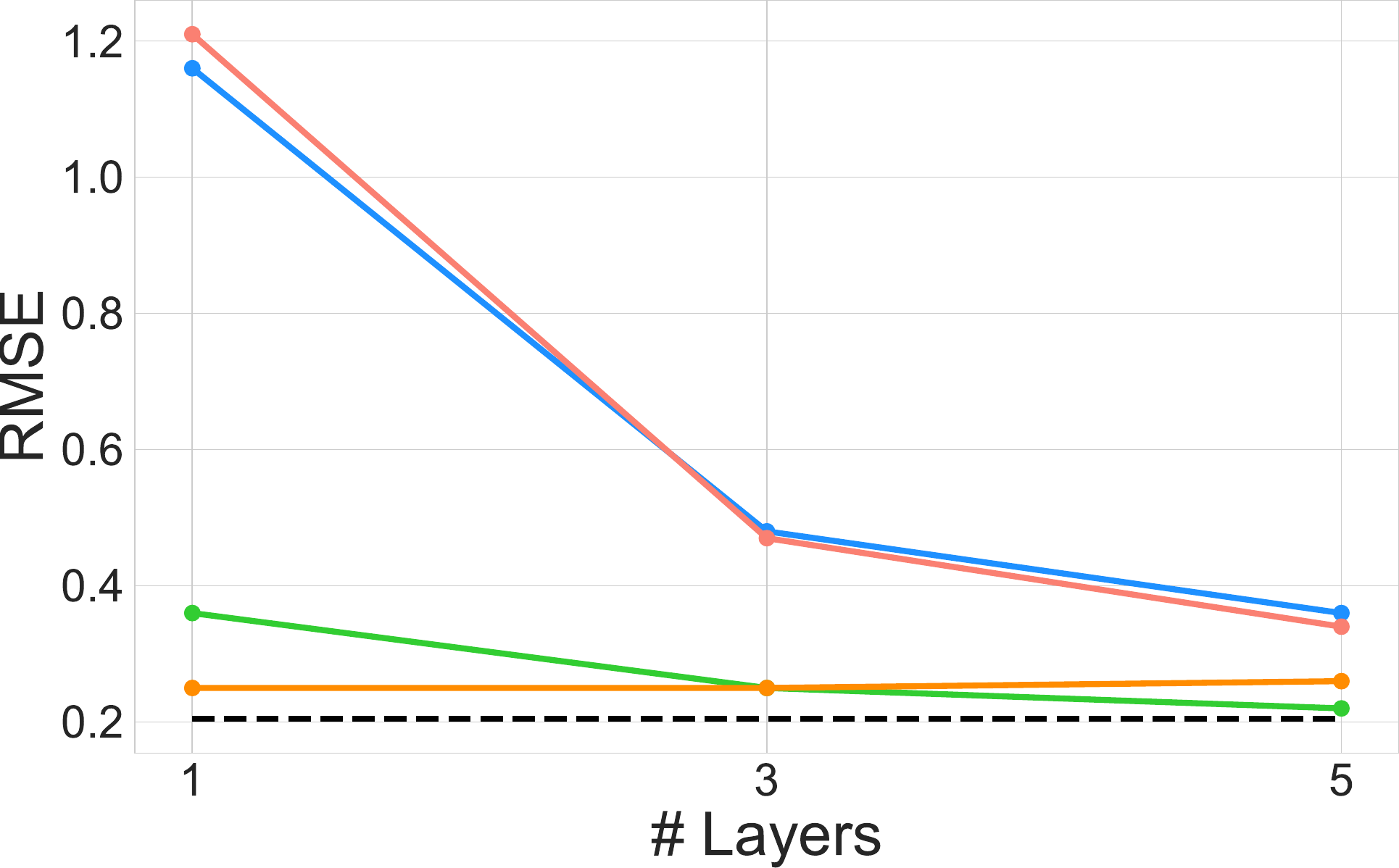} \\
    \includegraphics[width=0.45\linewidth]{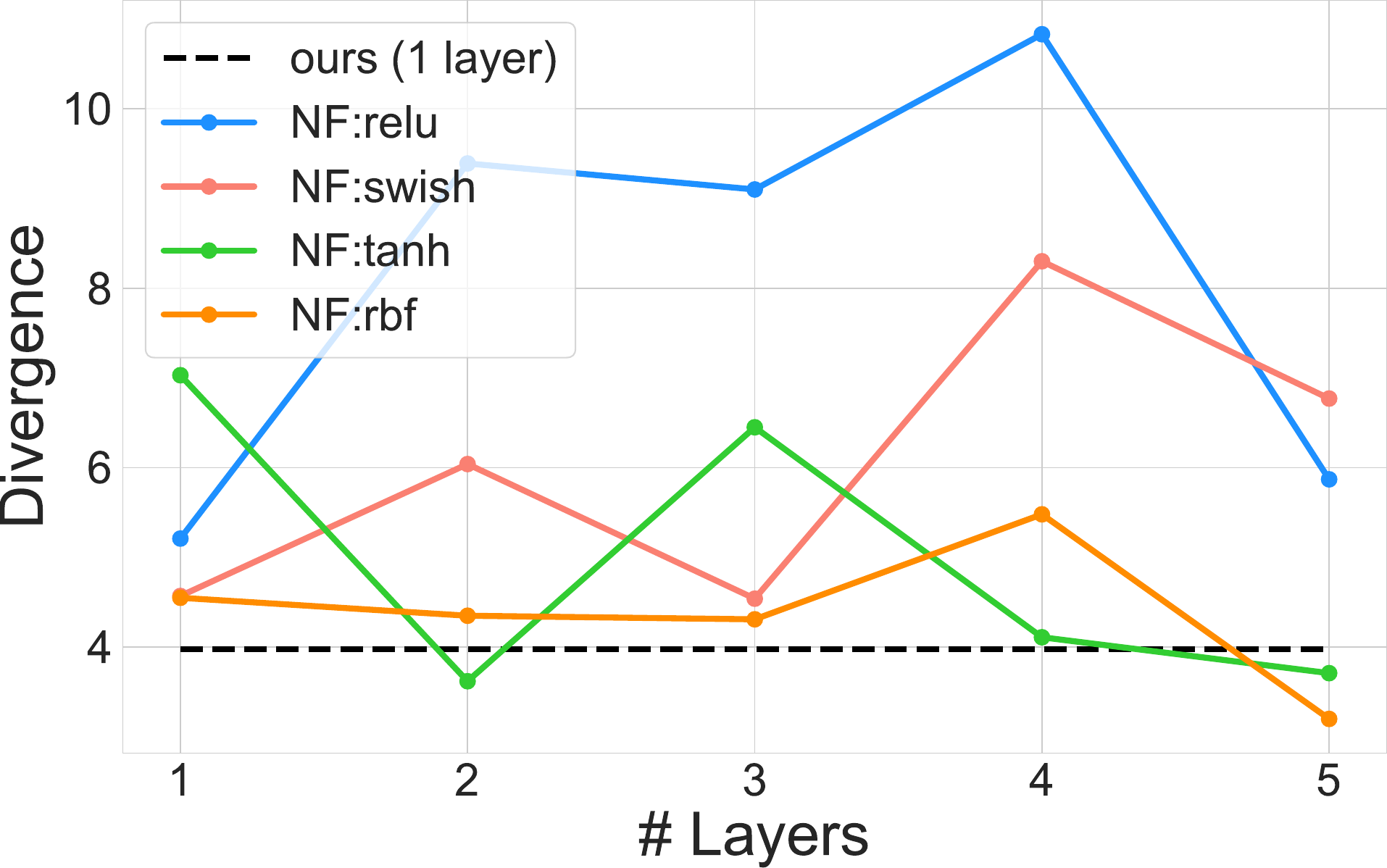}
    \includegraphics[width=0.45\linewidth]{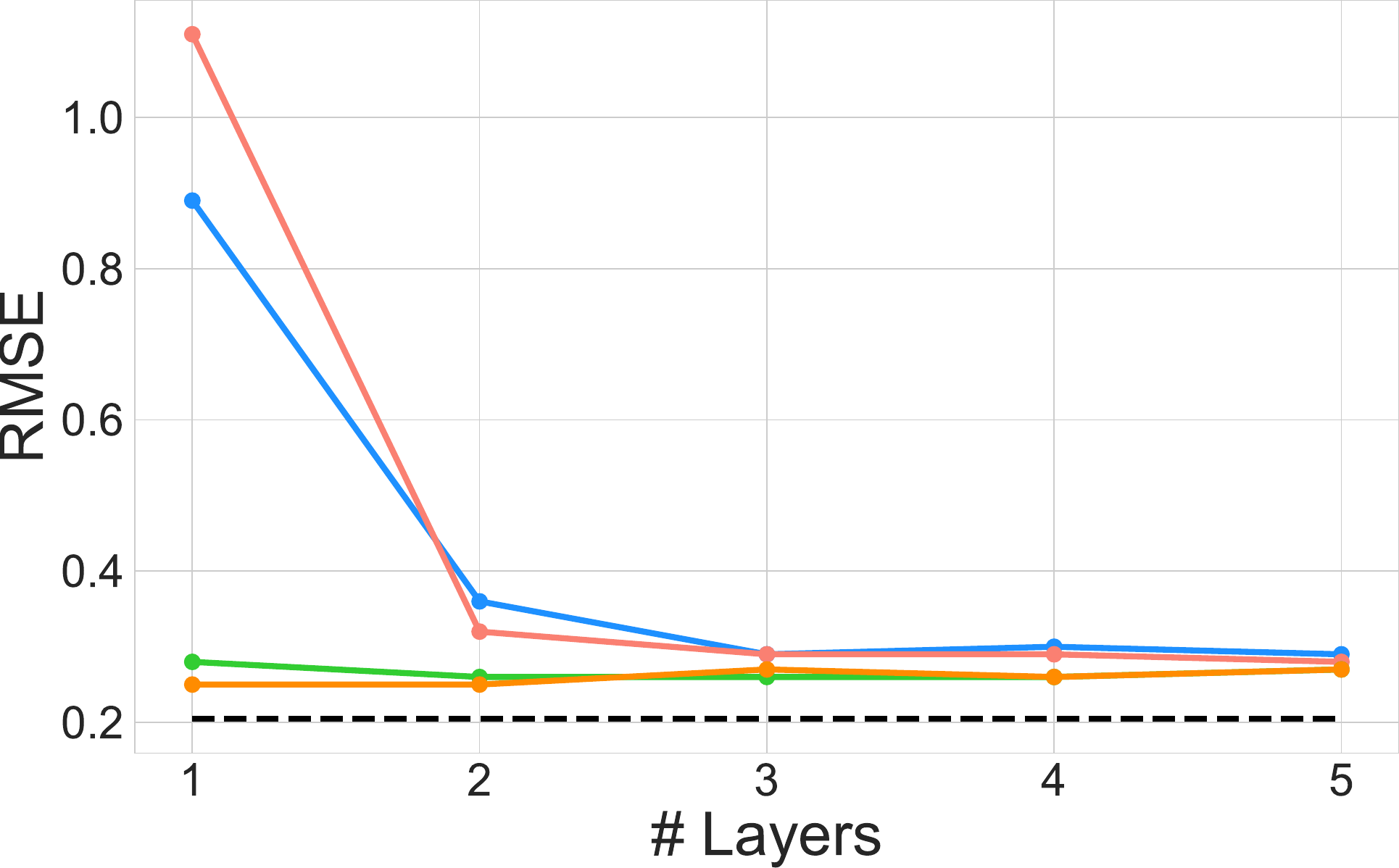}
    \caption{
    Comparison of posterior predictive quality across the methods from~\citep{tran2022all} for the problem from Fig.~\ref{fig:1d_regression}, considering different priors, activation functions, and numbers of hidden layers. We evaluate both \emph{out-of-distribution} (in-between region) performance (\textbf{left}, Wasserstein divergence) and \emph{in-distribution} performance (\textbf{right}, RMSE). For the Gaussian prior (\textbf{top}), the results align with the assumption by the authors -- adding more layers improves the GP approximation. Contrary, for the Normalizing Flow prior (\textbf{bottom}), the best performance is observed with one or two hidden layers, while deeper networks suffer from instability, highlighting failure of the heuristic.
    Our method (\textbf{black line}) consistently outperforms the baselines.
    }
    \label{fig:different_depth}
\end{figure}

In Fig.~\ref{fig:different_depth} we compare our approach against the behavior of their methods for varying numbers of layers and activations.
Specifically, we evaluate posterior in a 1D regression task (Fig.~\ref{fig:1d_regression}), both on training data and in out-of-distribution (OOD) regions. The results show that multiple hidden layers are required for the Gaussian prior. In contrast, for the NF priors, increasing depth leads to worse performance in OOD regions -- illustrating an example of the heuristic's failure. For both methods, we note high sensitivity to activation change. 
NFs perform best for one or two layers, but only with RBF activation. For Swish/ReLU they fail to capture the data (high RMSE/NLL). 
Please see also Sec.~\ref{sec:extensive}. 


For further evaluation, we compare our approach with~\cite{tran2022all} on a range of regression tasks, including both synthetic and real-world datasets (e.g., the Boston dataset). The results are shown in Fig.\ref{fig:1d_regression} and Tab.\ref{tab:uci_boston}, respectively. 
Additional discussion and results for several other regression tasks are provided in the Appendix.
We closely follow the settings picked by~\cite{tran2022all} and still observe that our method performs comparably or better than the baseline, while requiring only a single-hidden-layer BNN. 

Additionally, we  checked the findings from ~\citep{wu2024functional} that moment matching helps biasing optimisation towards better solutions. For the UCI regression task, we included an additional moment matching regularization term: {\small $\mathcal{R}(p_{nn}, p_{gp}) = $ ${\left(E_{p_{nn}}[var(f)]-E_{p_{gp}}[var(f)]\right)}^2 +$ ${\left(E_{p_{nn}}[kurtosis(f)]-E_{p_{gp}}[kurtosis(f)]\right)}^2 +$ ${\left(E_{p_{nn}}[skeweness(f)]-E_{p_{gp}}[skeweness(f)]\right)}^2$ }. However, we have \emph{not} observed any significant improvements.


\subsection{Can Learned Activations Match Performance of Closed-form Ones?}
\label{sec:closed_form}

Deriving suitable neural activations to match BNNs to GPs is a formidable challenge. For example, \cite{meronen2020stationary} derived recently an analytical activation function for the popular Matérn kernel. However, their solution relies on the assumption that the priors on the BNN weights follow binary white noise. Such the prior hinders existing posterior learning algorithms, making the proposed solution impractical. Nevertheless, \cite{meronen2020stationary} shows that their solution can work with more convenient Gaussian priors, albeit as a suboptimal approximation. In this work, we propose a general gradient-based alternative that can also be applied to the GPs with the Matérn kernel and we demonstrate that our black-box approach can match or even surpass the performance of the aforementioned approximation.

\begin{figure}[t]
\centering
\includegraphics[width=\linewidth]{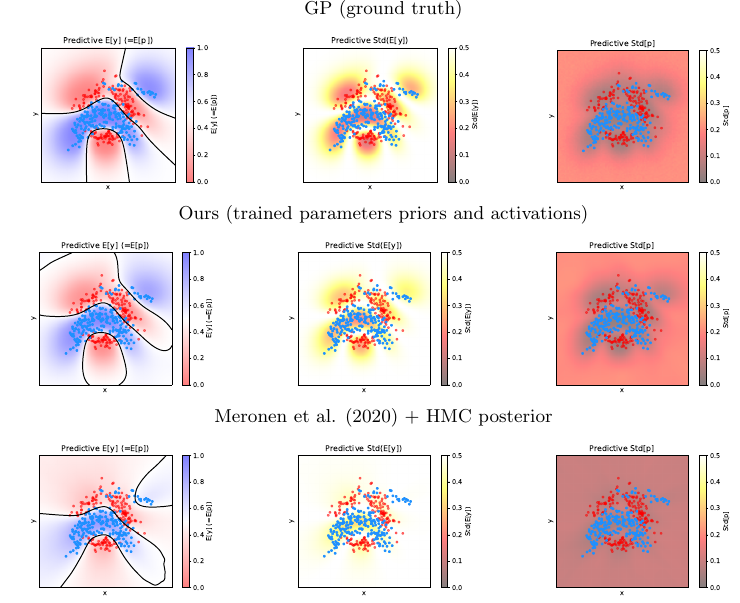}
\caption{
Posterior predictive distributions for a BNN with trained parameter priors and activations (ours; 2nd row) and for a BNN with analytically derived activations (3rd row). The first column illustrates class probabilities, the second column shows the total variance in class predictions, and the last column depicts the epistemic uncertainty component of the total uncertainty. Here, we show posteriors obtained using HMC, while in Sec.~\ref{sec:meronen2020stationary_appendix} of the Appendix, we provide an extended version of the figure that includes results obtained using the original code from \cite{meronen2020stationary} (which uses MC-Dropout instead of MCMC), along with additional numerical results.
}
\label{fig:e2e_twomoons_posteriors}
\end{figure}

The empirical evaluation we performed for the 2D data classification problem following the setting of~\cite{meronen2020stationary}. For our method, we used a BNN with Gaussian priors with trained variances, where the activation function was modeled by an NN with a single hidden layer consisting of 5 neurons using SiLU activation. For the baseline, we trained just variances and used the fixed activation. Posterior distributions were generally obtained using a HMC sampler. 

The results presented in Fig.~\ref{fig:e2e_twomoons_posteriors} (and extended results provided in Sec.~\ref{sec:meronen2020stationary_appendix} in the Appendix) demonstrate that our method enhances the match between the posteriors of a BNN and the desired target GP. Note that \cite{meronen2020stationary} originally used MC Dropout to obtain the posteriors, achieving much worse results than those obtained with HMC. 
For fairness in comparison, we tested both MC Dropout and HMC.
In terms of performance, our model not only \emph{captures class probabilities accurately but also adeptly handles the total variance in class predictions and the epistemic uncertainty component}, which are crucial for robust decision-making under uncertainty.


\subsection{Can Stationarity be Induced with Learned Periodic Activations?}

\begin{figure}[t]
\centering
\hspace{-0.25cm}
\scalebox{1.0}{
\begin{minipage}{1.0\linewidth} 
    \centering
    \scriptsize{Activation = NN($1 \times 5$, SiLU)}\\
    \includegraphics[width=1.02\linewidth]{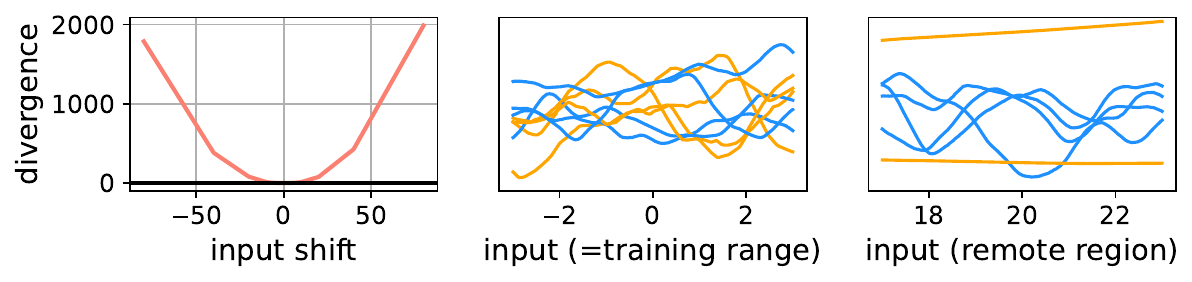} 
    \\
    \scriptsize{Activation = Sec.~\ref{sec:periodic}}\\
    \includegraphics[width=1.02\linewidth]{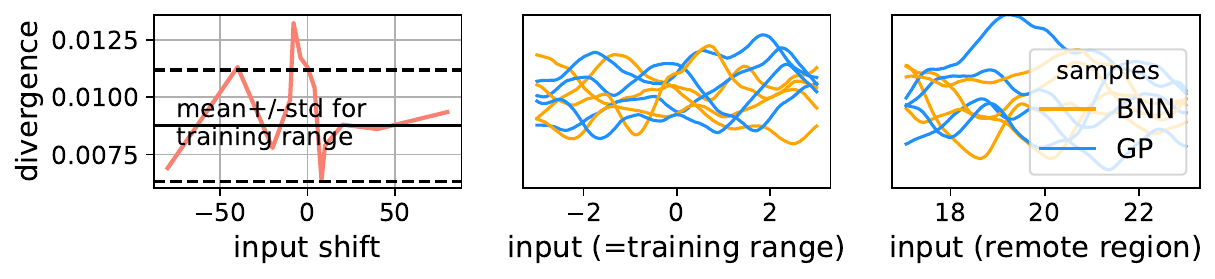} 
\end{minipage}
}
\caption{
Sensitivity of priors transferred from a GP to a BNN with respect to shifts in the input range for learned activations: (top) activation realized by a NN with a single hidden layer containing 5 neurons and SiLU own activation; (bottom) periodic activation introduced in Sec.~\ref{sec:periodic}, which is robust to input range shifts. Priors were trained on random inputs $X \in [-3, 3]$. The first column shows the mismatch between the desired target GP and the trained BNN prior for shifted input ranges, measured by the 2-Wasserstein loss (lower is better). The remaining columns present samples from the priors for two ranges: one matching the training range and one far from it.
}
\label{fig:experiments_shift_sensitivity_analysis}
\end{figure}

Trained priors for BNNs are inherently local, effectively mimicking GP behavior only within the range of training inputs $X$. As illustrated in Fig.~\ref{fig:experiments_shift_sensitivity_analysis} (top), outside this range, the BNN's learned prior may diverge from the desired behavior, and properties such as stationarity or smoothness may not be preserved beyond the observed input domain. We however can not only detect this issue by noting high values of the divergence $D$, but also mitigate it by using activations with appropriate structural biases. Specifically, the periodic activation introduced in Sec.~\ref{sec:periodic} proves to be robust to input domain translations, as demonstrated in Fig.~\ref{fig:experiments_shift_sensitivity_analysis}~(bottom).

\begin{figure}[t]
\centering
\centering
\includegraphics[width=0.325\textwidth,trim={0 4.39cm 0 0},clip]{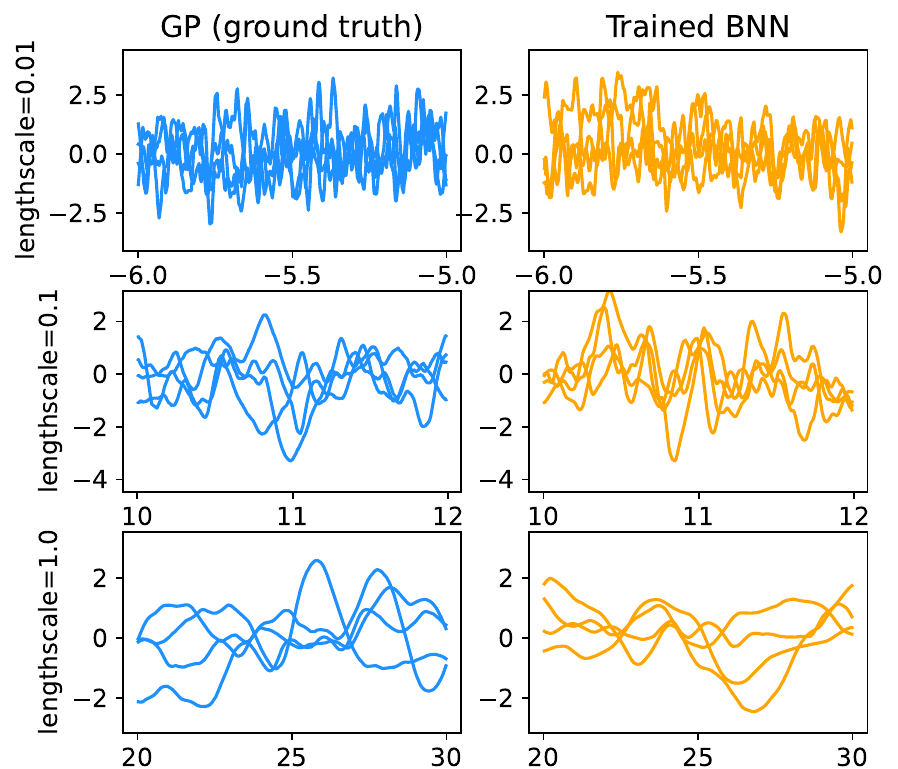} 
\caption{
Comparison of samples from a GP (left) and a trained BNN (right) using priors and activations (periodic activation as in Sec.~\ref{sec:periodic}) conditioned on the GP's hyperparameter (e.g., lengthscale). The BNN was trained on random inputs $X \in [-3, 3]$. Regardless of input range and the hyperparameter's value, samples from both models appear \emph{indistinguishable}, i.e., the BNN imitates the GP perfectly.
}
\label{fig:experiments_conditioning_on_lengthscale_preview}
\end{figure}

\subsection{Is Model Selection for BNNs with Transferred Priors Possible?}
\label{sec:conditioning_experiment}

A BNN with a functional prior learned from a GP essentially inherits the properties dictated by the GP, including those determined by specific values of the GP's hyperparameters. While the hyperparameters of the source GP can be tuned and optimized, the BNN does not inherently possess this capability, requiring researchers to either pre-optimize the hyperparameters or look for suitable workarounds~\citep{tran2022all}. Conditioning the priors on weights, biases, and activation, as explained in Sec.~\ref{sec:conditioning}, addresses this limitation, ultimately closing the BNN-GP equivalence gap.
Fig.~\ref{fig:experiments_conditioning_on_lengthscale_preview} shows results for a prior trained to adapt to a varying lengthscale of a Mat\'ern kernel. The prior closely follows the behavior of the ground truth model.

To validate the usefulness of this method, we performed marginal likelihood maximization on the data from Fig.~\ref{fig:1d_regression}, using gradient-based optimization for both the GP with the Mat\'ern kernel and our BNN. The optimal lengthscale found for the BNN ($\ell = 1.65$) closely matches the one found for the GP ($\ell = 1.84$).

\subsection{Does It Scale?}
\label{sec:houseelectric}

Learning posteriors for Gaussian Processes (GPs) is challenging, as exact inference scales cubically with the number of data points~\cite{rasmussen2005}. To address this, scalable approximations such as Stochastic Variational Gaussian Processes (SVGP)~\citep{pmlr-v38-hensman15} have been proposed for practical applications. SVGP employs stochastic variational gradients to optimize a fixed set of inducing points, typically (and also here) $1024$.

\begin{table*}[t]
\caption{
Mean test-set NLL and RMSE for the HouseElectric dataset. The optimal prior lengthscale $\ell^*$ and additive Gaussian noise were \emph{learned} by maximizing $\mathcal{L}$ with $\beta=1$. Results for $\ell = 10\ell^*$ and $\ell = 0.1\ell^*$ highlight the importance of prior hyperparameter selection. Additionally, using a model of activation function with more neurons improves performance. The best results are achieved by combining these modifications with $\beta=0.4$.
}
    \begin{center}
    \resizebox{0.98\linewidth}{!}{
    \begin{tabular}{lc|ccccc|c|c|c}
        \toprule
        & \multicolumn{4}{c}{Ours with $\beta=1$} & \multicolumn{2}{c}{Ours with $\beta=0.4$} & \multicolumn{1}{c}{$\pm$} & \multicolumn{1}{c}{SVGP}  & \multicolumn{1}{c}{Exact GP}  \\
        \cmidrule(lr){2-5} \cmidrule(lr){6-7} 
        & \emph{learned} & $\ell=0.1\times\ell^*$ & $\ell=10\times\ell^*$ & $\ell=\ell^*$ \& improved activation & $\ell=\ell^*$ & $\ell=\ell^*$ \& improved activation & std. & (baseline) & (BBMM)  \\
        \midrule
        $\downarrow$ Test RMSE & $0.0549$ & $0.0550$	& $0.0560$	& $0.0550$	& $0.0537$	& $\textbf{0.0535}$	& $\leq0.0007$  & $0.0566$ & $0.055$ \\
        $\downarrow$ Test NLL & $-1.4925$	& $-1.4850$	& $-1.4550$	& $-1.4980$	& $-1.5107$	& $\textbf{-1.5155}$	 &  $\leq0.0064$ & $-1.4600$ & $-0.152^\dagger$ \\
        \bottomrule
    \end{tabular}
    }  
    \end{center}    
    \scriptsize{
    ${}^\dagger$ Due to lack of available implementation, we failed to reproduce results from~\cite{wang2019exact} and provide here values copied directly from the paper.
    }
\label{tab:house_electric}
\end{table*}

On the other hand, there exist a number of posterior inference methods for BNNs. For the experiments presented in the previous sections, we utilized HMC. However, Variational Inference (VI) with Normalizing Flows~\citep{rezende2015variational} offers a scalable alternative, albeit with challenges related to both the accuracy of posterior approximation and  selection of hyperparameters. However, the former limitation can be mitigated by employing RealNVPs~\citep{dinh2016density} with appropriate capacity for approximating posteriors. \cite{agrawal2024disentangling} showed that RealNVPs can exhibit sufficient fidelity to match posteriors for models with complex priors.

For VI learning is performed using gradients of
{\small
\begin{align*}
    \mathcal{L} &= \mathbb{E}_{q(w | \zeta)} \left[ \log p(\mathcal{D} | w) \right] - \beta \cdot \text{KL}\left[q(w | \zeta) \, \| \, p(w)\right] \\
    &\approx \frac{1}{S} \sum_{w \sim q(w | \zeta)} \left( \frac{|\D|}{|\mathcal{B}|} \log p(\mathcal{B} | w) + \beta \cdot \left( p(w) - q(w | \zeta) \right) \right)
    \end{align*}
}
where we estimate the Evidence Lower Bound (ELBO) objective for Stochastic Variational Inference (SVI)~\citep{hoffman2013stochastic} using minibatches $\mathcal{B} \subset \mathcal{D}$ with $|\mathcal{B}| = 10{,}240$. The expectation is approximated with $S = 128$ reparameterized Monte Carlo samples, where $q(w | \zeta)$ is the posterior approximation modeled by a RealNVP.

Tab.~\ref{tab:house_electric} summarizes the test set performance on the UCI Household Power Consumption (HouseElectric) regression dataset. This dataset contains approximately 2M rows, split into training (7/9) and test (2/9) subsets, and is the largest dataset used to evaluate GPs~\citep{wang2019exact}. 

The regression model for the data has two hyperparameters: the prior \emph{lengthscale} $\ell$ of the Matérn ($\nu=3/2$) kernel and the additive \emph{observation noise} scale in the Gaussian likelihood. We selected these hyperparameters by maximizing the ELBO $\mathcal{L}$, which for $\beta=1$ provides a proper lower bound for the evidence. This was possible thanks to employing the conditional priors introduced in Sec.~\ref{sec:conditioning}. 

Variational Inference (VI) often performs better with $\beta < 1$~\citep{beta-vae}, which also holds in our case - setting $\beta=0.4$ (a typical value) improved both RMSE and NLL. However, as indicated by discrepancies with the Exact GP results, this improvement is likely due to the reduced influence of the learned GP prior, which may be suboptimal for this particular dataset.

To analyze the importance of prior selection, we also computed posteriors for lengthscales reduced by a factor of ten ($\ell = 0.1 \times \ell^*$) and increased tenfold ($\ell = 10 \times \ell^*$). In both cases, we observed a drop in NLL, highlighting the sensitivity of performance to choice of the prior.

We further tested a learned prior with slightly \emph{improved activation} model. We used a neural network with two hidden layers of width 10, instead of the default single layer with five neurons. This resulted in a $0.15\%$ improvement in the 2-Wasserstein loss. Although this represents a minor enhancement in prior fidelity, it still led to noticeable improvements in posterior quality.

\begin{figure}[t]
    \centering
    \includegraphics[width=0.45\linewidth]{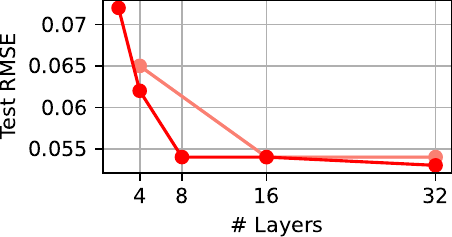}    
    \includegraphics[width=0.45\linewidth]{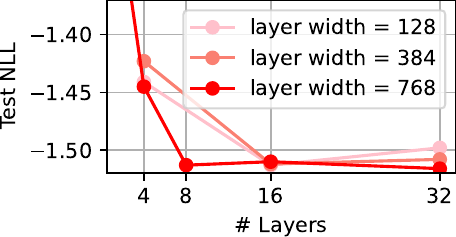}
    \caption{
Impact of the architecture of $q(w|\zeta)$ parameterized by \textit{RealNVP}~\citep{dinh2016density}: (1) although adding more layers improves performance, the gains become marginal for $>16$ layers; (2) wider flows require fewer layers to achieve performance saturation.
    }
    \label{fig:house_electric_rnvp}
\end{figure}

Finally, we investigated the capacity of the RealNVP used for posterior approximation. Fig.~\ref{fig:house_electric_rnvp} compares test set performance for varying the number of flow layers and the width of each layer. We observed that increasing the number of layers improves performance, but the gains become marginal beyond 16 layers, and that wider flows require fewer layers to achieve performance saturation.

\section{Conclusion}

In this paper, we addressed the problem of transferring functional priors for wide Bayesian Neural Networks to replicate desired a priori properties of Gaussian Processes. Previous approaches typically focused on learning distributions over weights and biases, often requiring deep BNNs for sufficient flexibility. We proposed an alternative approach by also learning activations, providing greater adaptability for shallow models and eradicating the need for task-specific architectural designs. To the best of our knowledge, we are the first to explore learning activations in this context. Moreover, to further enhance adaptability of these transferred priors, we came up with the idea of conditioning them with hypernetworks, which opens a new interesting future research direction.
To demonstrate the flexibility and effectiveness of the proposed methods, we conducted a comprehensive experimental study validating our ideas.

\clearpage

\begin{contributions}
{\small
\textbf{Marcin Sendera} actively participated in shaping the project across all its stages, contributing both during discussions and through hands-on development. He explored and assessed multiple variants of the Wasserstein loss, as well as a range of learnable activation functions proposed in the literature, including piecewise linear (PWL), Padé, and others. He proposed to use the closed-form solution for the Wasserstein metric. He was primarily responsible for implementing and executing the experiments reported in Figures~\ref{fig:1d_regression} and \ref{fig:different_depth}, and in Table \ref{tab:uci_boston}, ensuring the reproducibility and robustness of the results.
He contributed to writing of the manuscript, particularly in sections related to empirical results.

\textbf{Amin Sorkhei} participated in discussions at the initial stage of the project. He implemented and conducted the initial experiments involving the mapping of Gaussian Process priors to Bayesian Neural Networks using the Kullback-Leibler divergence and a learnable activation parameterized by a neural network.

\textbf{Tomasz Kuśmierczyk} conceived the project and supervised its development across all stages, including the core theoretical contributions and experimental design. He introduced the idea of learning activations and proposed the use of Wasserstein-based loss and regularizations. He proposed to use neural networks to parameterize learnable activations. He developed and implemented the periodic (Section~\ref{sec:periodic}) and conditioned activations (Section~\ref{sec:conditioning}).
He was responsible for implementing and executing the experiments presented in Figures~\ref{fig:matching_priors}, 
\ref{fig:e2e_twomoons_posteriors}, 
\ref{fig:experiments_shift_sensitivity_analysis}, 
\ref{fig:experiments_conditioning_on_lengthscale_preview}, 
\ref{fig:house_electric_rnvp}, and Table~\ref{tab:house_electric}. 
He was responsible for the majority of the manuscript writing as well as for shaping its final form.
}
\end{contributions}

\begin{acknowledgements}
{\small
We thank Nikolay Malkin and Jacek Tabor for helpful discussions at early
stages of this project and revising the manuscript.
\\ \\
This research is part of the project No. \textbf{2022/45/P/ST6/02969} co-funded by the National
Science Centre and the European Union Framework Programme for Research and
Innovation Horizon 2020 under the Marie Skłodowska-Curie grant agreement No.
945339. For the purpose of Open Access, the author has applied a CC-BY public copyright
licence to any Author Accepted Manuscript (AAM) version arising from this submission. 
\\
\includegraphics[width=1cm]{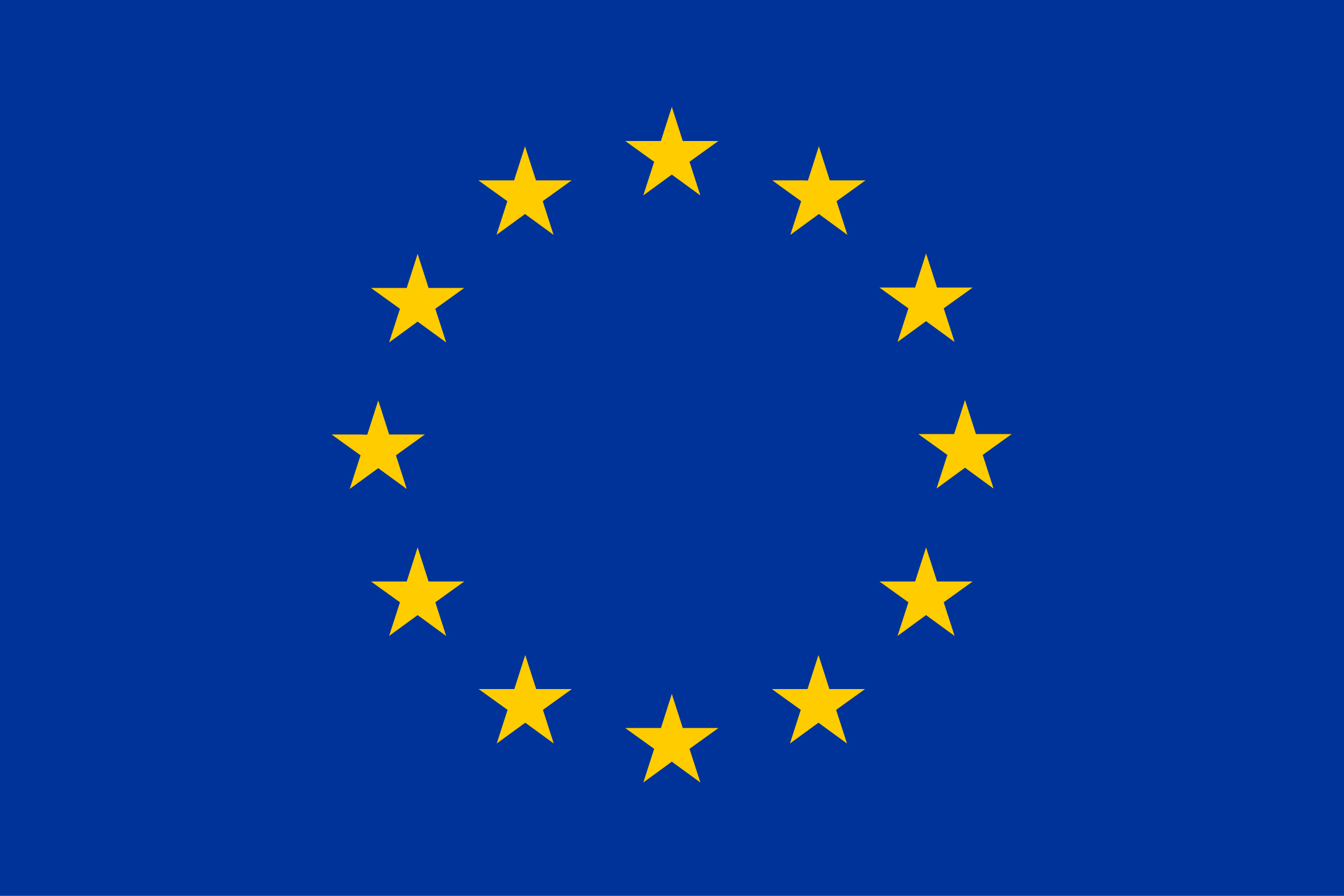} \includegraphics[width=1.9cm]{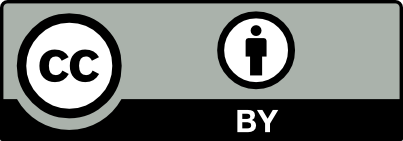}
\\ \\
This research was in part funded by National Science Centre, Poland, \textbf{2022/45/N/ST6/03374}.
\\ \\
We gratefully acknowledge Polish high-performance computing infrastructure PLGrid (HPC Center: ACK Cyfronet AGH) for providing computer facilities and support within computational grant no. \textbf{PLG/2023/016302}.
}
\end{acknowledgements}

\bibliography{bibliography}

\newpage

\onecolumn

\title{Revisiting the Equivalence of Bayesian Neural Networks and Gaussian Processes: On the Importance of Learning Activations\\(Supplementary Material)}
\maketitle

\appendix
We supplement the main text here with a description of the related work, a discussion of computational costs, and additional details of the experimental setup, followed by additional numerical results.
Our code we made publicly available~\footnote{\href{https://github.com/gmum/bnn-functional-priors}{https://github.com/gmum/bnn-functional-priors}}.

\section{Related Work}
\label{sec:related_work}

The relationship between (B)NNs and GPs has been a subject of significant research interest. Much of this work has been motivated by the desire to better understand NNs through their connection to GPs. A seminal result by \cite{neal96a, williams1996computing}, later extended to deep NNs by \cite{lee2017deep, matthews2018gaussian}, shows that infinitely wide layers in NNs exhibit behavior akin to GPs. 
While we explore a similar setting, our focus is the inverse: implementing function-space priors, specifically GP-like, within BNNs.

This problem was previously addressed by \cite{meronen2020stationary}, who proposed an activation function for BNNs that corresponds to the popular Mat\'ern kernel. Due to the complexity of deriving analytical solutions, other works, such as \cite{flam2017mapping,flam2018characterizing,tran2022all}, have explored gradient-based optimization to learn priors on weights and biases. However, their approach requires deep networks with complex distributions and additionally, presents a challenge in learning posterior. In contrast, we achieve high-fidelity functional priors in shallow BNNs by matching \emph{both} priors on parameters and activations.

Finding accurate posteriors for deep and complex models such as BNNs is notoriously challenging due to their high-dimensional parameter spaces and complex likelihood surfaces. However, for single-hidden-layer wide BNNs, it has been recently shown that posterior sampling via Markov Chain Monte Carlo can be performed efficiently~\citep{pmlr-v162-hron22a}. Moreover, research has demonstrated that the exact posterior of a wide BNN weakly converges to the posterior corresponding to the GP that matches the BNN's prior~\citep{hron2020exact}. This allows BNNs to inherit GP-like properties while preserving their advantages. 

An alternative line of work touching efficient learning for BNNs in function space has explored variational inference. 
\cite{sun2018functional} introduced functional variational BNNs (fBNNs) that maximize an Evidence Lower Bound defined over functions. However, \cite{ma2020understanding} highlighted issues with using the Kullback-Leibler divergence in function space, leading to ill-defined objectives. To overcome this, methods like Gaussian Wasserstein inference \citep{xu2023generalized} and Functional Wasserstein Bridge Inference \citep{wu2024functional} leverage the Wasserstein distance to define well-behaved variational objectives. These approaches try to incorporate functional priors into variational  objective for deep networks. In contrast, in our method transfer of priors to shallow BNNs remains independent and fully separated from an inference method. 

Finally, other works worth mentioning include \cite{meronen2021periodic}, who explored periodic activation functions in BNNs to connect network weight priors with translation-invariant GP priors. 
\cite{pearce2020expressive} derived BNN architectures to mirror GP kernel combinations, showcasing how BNNs can produce periodic kernels.
\cite{karaletsos2020hierarchical} introduced a hierarchical model using GP for weights to encode correlated weight structures and input-dependent weight priors, aimed at regularizing the function space. \cite{matsubara2021ridgelet} proposed using ridgelet transforms to approximate GP function-space distributions with BNN weight-space distributions, providing a non-asymptotic analysis with finite sample-size error bounds. Finally, \cite{tsuchida2019richer} extended the convergence of NN function distributions to GPs under broader conditions, including partially exchangeable priors.
A more detailed discussion of related topics can be found for example, in Sections 2.3 and 4.2 of \citep{FortuinPriors}.

\section{Computational and Architectural Advantages}
\label{sec:wallclock}

Our method involves an initial prior fitting step, which precedes the posterior inference step.
The prior fitting step takes approximately one hour for the HouseElectric dataset (Section~\ref{sec:houseelectric}); however, one can alternatively use a prior from a library of pre-trained priors, thereby amortizing the cost of this step.

Our experiments show that substituting fixed activations (ReLU/TanH, $\sim 0.5$s/iteration) with learnable NN activations (single hidden layer, 5 neurons, $\sim 2.3$s/iteration) or periodic activations ($\sim 2.7$s/iteration) increases the time per iteration by approximately a factor of 4 on our 16-core CPU setup. 
On the other hand,
learning parameter priors alongside activations has only a marginal effect on computation time compared to fixing priors and only learning activations. 

\begin{figure}[t!]
\centering
\begin{minipage}[t]{0.375\textwidth}
\centering
\includegraphics[width=\linewidth]{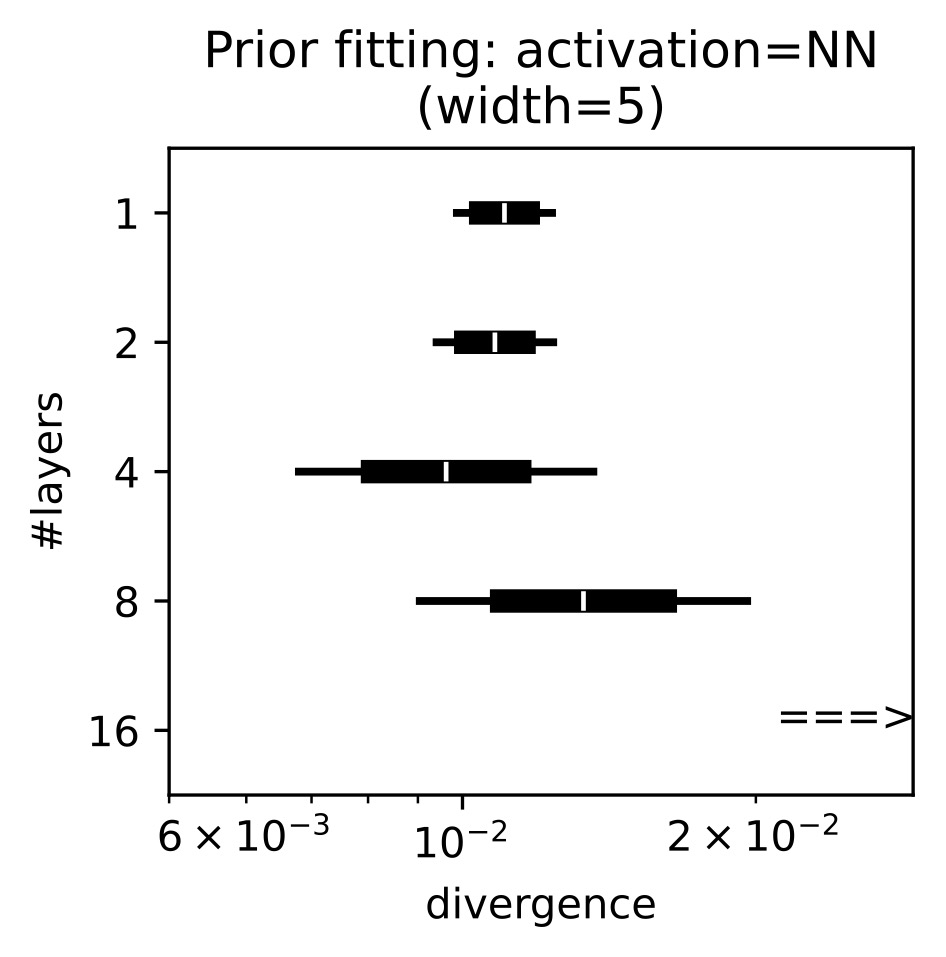}
\caption{Prior matching quality: fixed width.}
\label{fig:prior_matching_time1}
\end{minipage}
\hfill
\begin{minipage}[t]{0.375\textwidth}
\centering
\includegraphics[width=\linewidth]{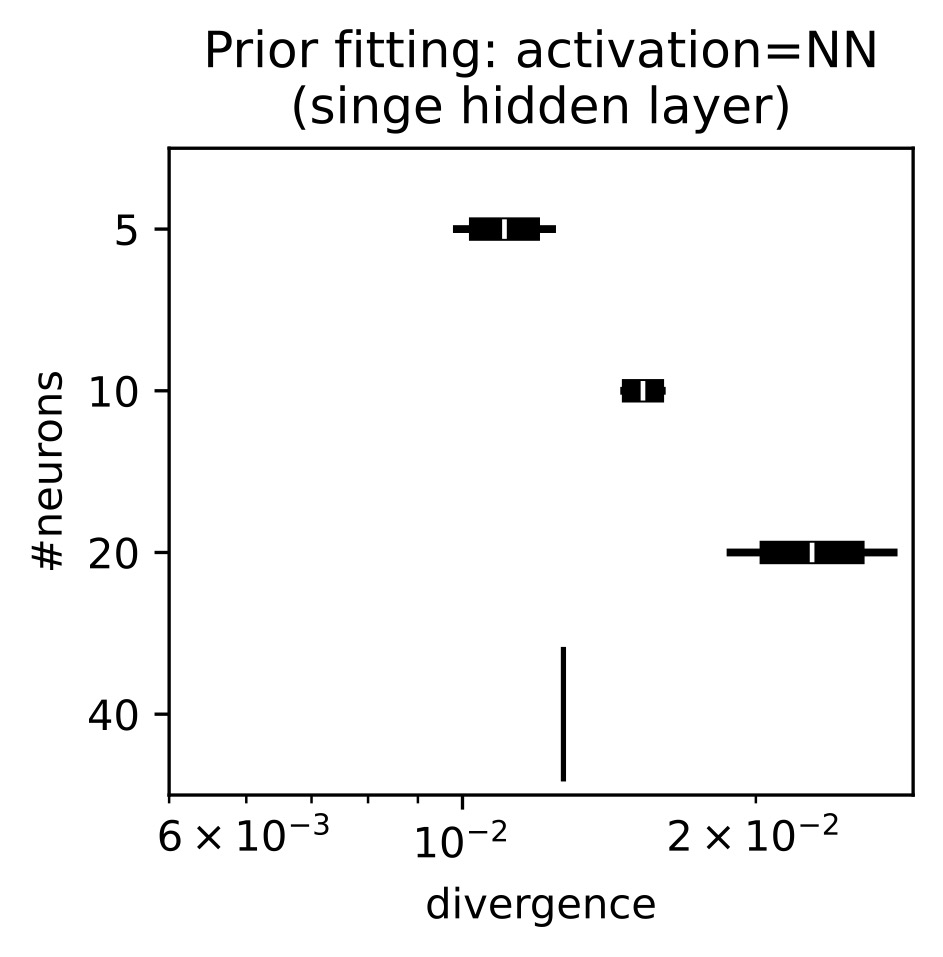}
\caption{Prior matching quality: fixed depth.}
\label{fig:prior_matching_time2}
\end{minipage}
\end{figure}

Additionally, we conducted a study measuring both the quality of prior fit and the runtime for a prior modeled by a neural network with varying numbers of layers and widths. The respective results can be found in Figures~\ref{fig:prior_matching_time1},~\ref{fig:prior_matching_time2}
and 
Tables~\ref{tab:prior_matching_time1},~\ref{tab:prior_matching_time2}. 
These results suggest that increasing the number of layers up to a certain depth (e.g., 4 layers in this case) may marginally improve the fidelity of the solutions, whereas increasing the network width tends to make convergence harder. In either case, using a larger neural network consistently leads to longer convergence times.

\begin{minipage}[t]{0.45\textwidth}
\centering
\begin{table}[H]
\centering
\caption{Prior matching time\\(activation=NN, width=5)}
\begin{tabular}{|c|c|}
\hline
\#layers & iteration time [s] \\
\hline
1  & 2.34 \\
2  & 3.10 \\
4  & 4.57 \\
8  & 7.38 \\
16 & 15.03 \\
\hline
\end{tabular}
\label{tab:prior_matching_time1}
\end{table}
\end{minipage}
\hfill
\begin{minipage}[t]{0.45\textwidth}
\centering
\begin{table}[H]
\caption{Prior matching time\\(activation=NN, single hidden layer)}
\begin{tabular}{|c|c|}
\hline
\#neurons & iteration time [s] \\
\hline
5  & 2.34 \\
10 & 2.59 \\
20 & 3.95 \\
40 & 7.59 \\
\hline
\end{tabular}
\label{tab:prior_matching_time2}
\end{table}
\end{minipage}

Once the prior is established (or a pre-trained one is used), posterior inference can be performed. 
Our approach offers here computational advantages, particularly for large datasets where exact GP inference is prohibitive.
For example, RealNVP-based variational inference (Section~\ref{sec:houseelectric}) converges in about 2 hours on a single GPU for HouseElectric, while exact GP inference can take over three days on a single GPU, or 1.5 hours when parallelized on 8 GPUs using methods such as BBMM~\citep{wang2019exact}.
Furthermore, during posterior inference using HMC for the data in Fig.~\ref{fig:1d_regression}, we observed no significant differences in step time ($\sim 1$s) between fixed and learnable activations, though this may vary with implementation and hardware. 

Finally, beyond computational performance, BNNs provide greater modularity and extensibility. Single-layer BNNs can serve as components in larger neural architectures, for example, as Bayesian last layers (with remaining layers trained point-wise). This enables leveraging architectural innovations from deep learning in ways that GPs cannot easily accommodate. 
Our method could replace existing approaches in models like SNGP \citep{10.5555/3495724.3496353}, potentially offering improved prior control (as random Fourier feature GPs are limited to specific stationary kernels), more off-the-shelf end-to-end training options (e.g., SVGD), and potentially enhanced performance, as suggested by our comparisons against SVGP in Section~\ref{sec:houseelectric}.

\section{Learned GP Kernels}

In this paper, we show that our approach allows for inducing a GP-like behavior onto single-layer BNNs without being restricted to a family of GP kernels.
In particular, we induced the following GP kernel behavior, while performing the mentioned experiments:

\begin{itemize}
    \item Mat\'ern (3/2) - house electric regression;
    \item Mat\'ern (5/2) - inducing stationarity, ablation on learning activations and priors, 2d classification;
    \item RBF - 1d regression;
    \item RBF + ARD - UCI regression;
    \item Periodic - ablation on learning activations and priors.
\end{itemize}

\section{Experimental Details}
\label{sec:experiments}

\subsection{Can Learned Activations Achieve More Faithful Function-space Priors? -- Additional Figures}

\begin{figure}[h]
\centering
\begin{minipage}{0.5\linewidth}
\includegraphics[width=1.0\textwidth]{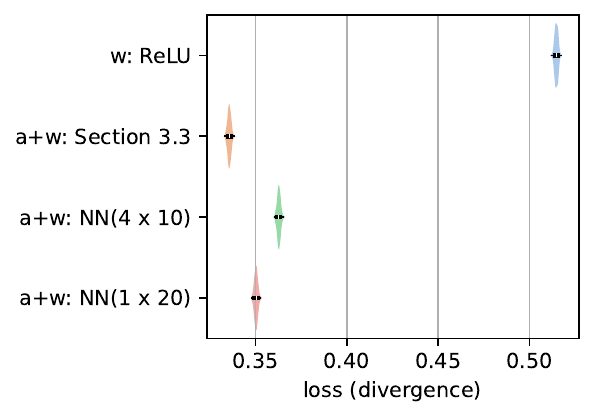} 
\end{minipage}%
\hfill
\begin{minipage}{0.5\linewidth} 
\centering
    \begin{minipage}{0.5\textwidth}
    \centering
    GP samples 
    \end{minipage}%
    \hfill
    \begin{minipage}{0.5\textwidth}
    \centering
    BNN samples
    \end{minipage}%
    \\
    \includegraphics[width=0.45\textwidth]{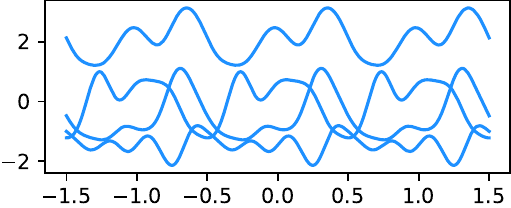} 
    \includegraphics[width=0.45\textwidth]{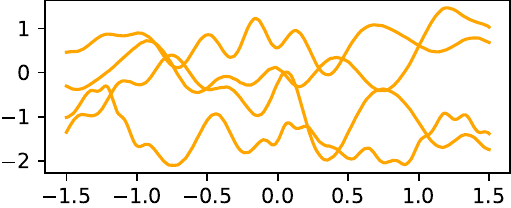} \\
    Various failure modes:
    \\
    \includegraphics[width=0.45\textwidth]{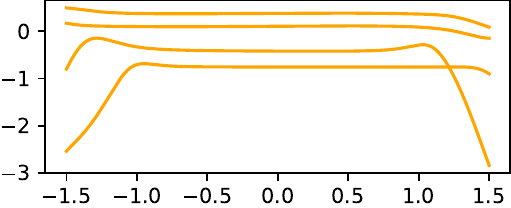}
    \includegraphics[width=0.45\textwidth]{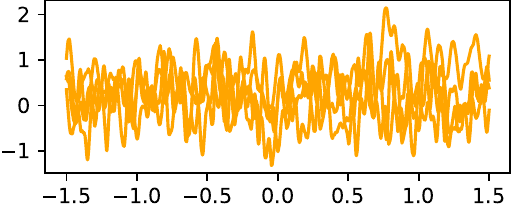} \\
    \includegraphics[width=0.45\textwidth]{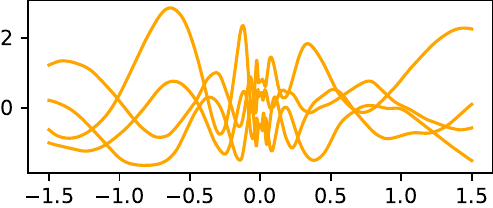}
    \includegraphics[width=0.45\textwidth]{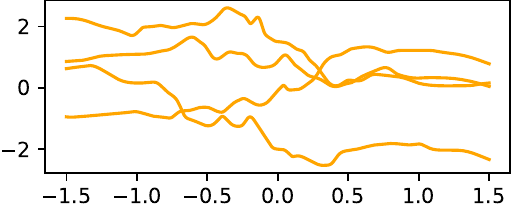} \\
\end{minipage}
\caption{
Quality of matching BNNs to the prior of a GP with the Periodic kernel ($\ell=1$, $p=1$). (Left): Evaluation of models with trained parameter priors (denoted by \textit{w}) and with trained both weights priors and activations (denoted by \textit{a+w}). (Right): Samples from 
the ground truth GP, a well-fitted BNN and
BNN outliers with poor final loss. As explained in Sec.~\ref{sec:opt_challenges}, or because of the insufficient modeling capacity of $\phi$, gradient-based optimization may fail to capture a GP functional prior.
}
\label{fig:matching_periodic_priors}
\end{figure}

\clearpage
\subsection{1-dimensional Regression - Comparison with \cite{tran2022all}}

In order to illustrate the abilities of our approach on a standard regression task and for a fair comparison with another method optimizing the Wasserstein distance, we followed the exact experimental setting presented in \cite{tran2022all}. We used a GP with an RBF kernel (\textit{length scale} $\ell$=0.6, \textit{amplitude}=1.0) and a Gaussian likelihood (\textit{noise variance}=0.1) as the ground truth.

For a baseline, we utilized the approach by \cite{tran2022all}, consisting of a BNN with 3 layers and 50 neurons each, using TanH activation function. We considered all three prior realizations presented in that work: a simple Gaussian prior, a hierarchical prior, and a prior given by a Normalizing Flow.
On the other hand, our model configuration consisted of a BNN with Gaussian priors centered at $\mathbf{0}$ and trained variances, where the activation function was modeled by a neural network with a single hidden layer with 5 neurons and SiLU activation. Posterior distributions were obtained using an HMC sampler.

We found that our approach allows for achieving the same or better results than the baseline by \cite{tran2022all}, both in visual and numerical comparisons. We obtained better results in terms of distributional metrics as presented in Tab. \ref{tab:e2e_regression_priors} and Tab.~\ref{tab:e2e_regression_posteriors} and getting better (visually) posterior distributions (see Fig.~\ref{fig:1d_regression}) without utilizing any 
computationally heavy prior realization (like, \textit{e.g.,} Normalizing Flow).

\begin{table}[h!]
\caption{
Comparison of trained (function-space) priors in a 1D regression task: \citep{tran2022all} vs. our method. Whereas \cite{tran2022all} considered three different priors on parameters (including Gaussian, Hierarchical, and Normalizing Flow) for deep BNNs, our implementation consists of a single-hidden-layer BNN with standard Gaussian priors with learned variances and trained activation. The methods were compared using a set of distributional metrics against the ground truth provided by a GP. We compared prior distributions of functions over a range of $X$ covering regions with and without data (to account also for overconfidence far from data). The lower, the better.
}
\centering
\resizebox{1\linewidth}{!}{
\begin{tabular}{l | ccccccccccc}
 Setting & \rotatebox{90}{1-Wasserstein} & \rotatebox{90}{2-Wasserstein} & \rotatebox{90}{Linear-MMD} & \rotatebox{90}{Poly-MMD $\times 10^3$} & \rotatebox{90}{RBF-MMD} & \rotatebox{90}{Mean-MSE} & \rotatebox{90}{Mean-L2} & \rotatebox{90}{Mean-L1} & \rotatebox{90}{Median-MSE} & \rotatebox{90}{Median-L2} & \rotatebox{90}{Median-L1}
\\ \hline \\ 
\textbf{\textit{Priors:}} &   &  &  &  &  &  &  &  &  &  & \\
 Gaussian prior &  7.52 & 7.77 & -6.08 & 1.158 & 0.036 & 0.003 & 0.058 & 0.045 & 0.004 & 0.063 & 0.052\\
 Hierarchical prior  & 8.14 & 8.37 & 0.22 & 1.421 & 0.027 & 0.004 & 0.064 & 0.059 & 0.008 & 0.090 & 0.074\\
 Normalizing Flow prior  & 7.61 & 8.09 & -8.192 & 0.393 & 0.019 & 0.004 & 0.061 & 0.051 & 0.004 & 0.066 & 0.052\\
 \textbf{ours}  & \textbf{6.86} & \textbf{7.06} & \textbf{-9.40}  & \textbf{-0.016} & \textbf{0.008} & \textbf{0.001} & \textbf{0.032} & \textbf{0.027} & \textbf{0.002} & \textbf{0.048} & \textbf{0.041} \\
\end{tabular}
\label{tab:e2e_regression_priors}
}
\end{table}

\begin{table}[h!]
\caption{
Comparison of trained (function-space) posteriors in a 1D regression task: \citep{tran2022all} vs. our method. Whereas \cite{tran2022all} considered three different priors on parameters (including Gaussian, Hierarchical, and Normalizing Flow) for deep BNNs, our implementation consists of a single-hidden-layer BNN with standard Gaussian priors with learned variances and trained activation. The methods were compared on RMSE and NLL of test data.
}
\centering
\begin{tabular}{l | cc}
 Setting & RMSE & NLL \\ \hline \\ 
\textbf{\textit{Posteriors:}} &   &  \\
 Gaussian prior &  0.2559 & \textbf{0.2456} \\
 Hierarchical prior  & 0.2461 & 0.2613 \\
 Normalizing Flow prior  & 0.2534 & 0.2550 \\
 \textbf{ours}  & \textbf{0.2045} & 0.3044  \\
\end{tabular}
\label{tab:e2e_regression_posteriors}
\end{table}

\subsection{UCI Regression - Comparison with \cite{tran2022all}}

In addition, we compare against \cite{tran2022all} also on UCI regression tasks, which have more input dimensions $d$ -- between $8$ and $12$, while having 1-dimensional output. We use 10-split of training datasets as in typical in this direction of research.

For the baseline, we followed the exact experimental setting presented in \cite{tran2022all}. We used a GP with an RBF kernel (\textit{length scale} $\ell = \sqrt{2.0 * input\_dim}$ and \textit{amplitude}=1.0, $ARD$) and a Gaussian likelihood as the ground truth. We set the noise variance value and architectures for the specific datasets as in the mentioned baseline paper.

On the other hand, our model configuration consisted of a BNN with Gaussian priors centered at $\mathbf{0}$ and trained variances, where the activation function was modeled by a neural network with a single hidden layer with 5 neurons and SiLU activation. Posterior distributions were obtained using an HMC sampler.

We found that our approach allows for achieving the same or better results than the baseline, in terms of numerical values (\textit{RMSE} and \textit{NLL}) as presented in Tab.~\ref{tab:uci_extended}. 

\def\gW{\mathcal{W}}
\begin{table*}[h!]
\caption{\label{tab:uci_extended}Extended results for a set of UCI regression tasks (various input dimensionality $d$) with prior transferred from a GP; comparison between the baseline (using a deep BNN; \citep{tran2022all}) and ours (single hidden layer BNN with 128 neurons; w/o regularization) and an activation realized by a NN 
with SiLU own activation were used.}%
    \centering
    \resizebox{0.98\linewidth}{!}{
    \begin{tabular}{lcccccccccc}
        \toprule
        Dataset $\rightarrow$
        & \multicolumn{2}{c}{Boston ($d = 12$)}
        & \multicolumn{2}{c}{Concrete ($d = 8$)} 
        & \multicolumn{2}{c}{Energy ($d = 8$)} 
        & \multicolumn{2}{c}{Protein ($d = 9$)}
        & \multicolumn{2}{c}{Wine ($d = 11$)}
        \\
        \cmidrule(lr){2-3}\cmidrule(lr){4-5}\cmidrule(lr){6-7}\cmidrule(lr){8-9}\cmidrule(lr){10-11}
        Method $\downarrow$ Metric $\rightarrow$
        & \multicolumn{1}{c}{\textit{RMSE}} 
        & \multicolumn{1}{c}{\textit{NLL}}
        & \multicolumn{1}{c}{\textit{RMSE}} 
        & \multicolumn{1}{c}{\textit{NLL}}
        & \multicolumn{1}{c}{\textit{RMSE}} 
        & \multicolumn{1}{c}{\textit{NLL}} 
        & \multicolumn{1}{c}{\textit{RMSE}} 
        & \multicolumn{1}{c}{\textit{NLL}}
        & \multicolumn{1}{c}{\textit{RMSE}} 
        & \multicolumn{1}{c}{\textit{NLL}}
        \\
        \midrule
         \textit{baseline} & 2.8402\std{0.8986} & 2.4778\std{0.1481} & 4.4628\std{0.7511} & 2.9728\std{0.0876} & 0.3431\std{0.0613} & 0.3482\std{0.1607} & 0.5038\std{0.0093} & 0.7447\std{0.0142} & 0.4736\std{0.0420} & 0.8725\std{0.0285}  \\
         \textit{ours} & 2.8643\std{0.8386} & 2.4937\std{0.1798} & 4.9143\std{0.7528} & 3.0201\std{0.1011} & 0.3692\std{0.0566} & 0.4064\std{0.1347} & 0.4957\std{0.0058} & 0.7251\std{0.0094} & 0.4833\std{0.0398} & 0.8673\std{0.0255} \\
        \bottomrule
    \end{tabular}
    }
\end{table*}

\subsection{Can Learned Activations Match Performance of Closed-form Ones? -- Comparison~with \cite{meronen2020stationary}}
\label{sec:meronen2020stationary_appendix}

For the comparison against \cite{meronen2020stationary}, we closely followed their experimental setting and used a GP with a Mat\'ern kernel ($\nu=5/2$, $\ell$=1) as the ground truth. For a baseline, we utilized the approach by \cite{meronen2020stationary}, where the authors derived analytical activations to match the Matérn kernel. Our model configuration consists of a BNN with Gaussian priors and trained variances, where the activation function was modeled by a neural network with a single hidden layer consisting of 5 neurons using SiLU activation. Posterior distributions were generally obtained using an HMC sampler, except in the case of \cite{meronen2020stationary}, where the original implementation employed MC Dropout to approximate the posterior. However, for completeness and fairness in comparison, we also generated results using an HMC-derived posterior for the model by \cite{meronen2020stationary}. Additional tests (see Tab.~\ref{tab:e2e_twomoons_posteriors}) were conducted on BNNs with fixed parameter priors (\emph{Default}=Gaussian priors with a variance of 1, and \emph{Normal}=Gaussian priors normalized by the hidden layer width) and with activation functions including ReLU and TanH.

\begin{figure*}[h!]
\centering
\hspace{-2.5cm}
\includegraphics[width=1.0\textwidth]{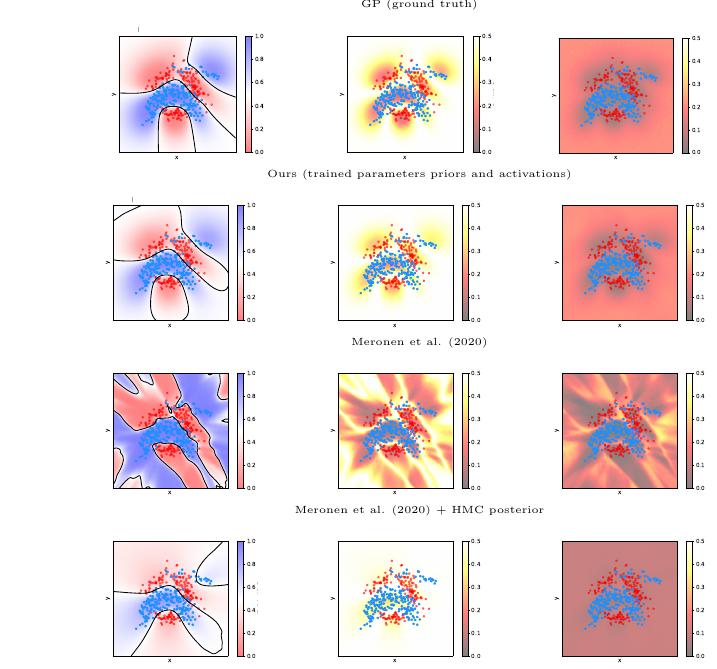}
\caption{
Posterior predictive distributions for a BNN with trained parameters priors and activations (ours; 2nd row), and for a BNN with analytically derived activation (3rd and 4th row). The first column illustrates class probabilities, the second column shows the total variance in class predictions, and the last column depicts the epistemic uncertainty component of the total uncertainty.
}
\label{fig:e2e_twomoons_posteriors_appendix}
\end{figure*}

\begin{table}[h!]
\caption{
Similarity of the posterior predictive distributions of BNNs with a single wide hidden layer to the posterior of a GP, considered as the ground truth. 
Calculations were performed for a 2D test grid $X$ ($100 \times 100$) spanning the area visible in Fig.~\ref{fig:e2e_twomoons_posteriors_appendix}, which includes regions both with and without training data to account for overconfidence far from the data. The posteriors were obtained using the HMC sampler in all cases, except for \citep{meronen2020stationary}, where the original code was used (however, results for this model with an HMC-derived posterior are also included below). Lower values indicate better performance. For our method, we present results using weights and priors pretrained for 1D inputs, as well as those trained on functional priors for 1D/2D inputs matched to the task on $X$.
}
\centering
\resizebox{1\linewidth}{!}{
\begin{tabular}{l | ccccccccccc}
 Weights and activation & \rotatebox{90}{1-Wasserstein} & \rotatebox{90}{2-Wasserstein} & \rotatebox{90}{Linear-MMD} & \rotatebox{90}{Mean-L1} & \rotatebox{90}{Mean-L2} & \rotatebox{90}{Mean-MSE} & \rotatebox{90}{Median-L1} & \rotatebox{90}{Median-L2} & \rotatebox{90}{Median-MSE} & \rotatebox{90}{Poly-MMD$\times 10^3$} & \rotatebox{90}{RBF-MMD}
\\ \hline \\ 
Default with ReLU & 39 & 39 & 667 & 0.23 & 0.26 & 0.067 & 0.3 & 0.33 & 0.11 & 4518 & 0.16\\
Default with TanH & 38 & 38 & 595 & 0.22 & 0.24 & 0.058 & 0.3 & 0.32 & 0.11 & 4181 & 0.15\\
Normal with ReLU & 31 & 32 & 603 & 0.21 & 0.25 & 0.061 & 0.22 & 0.26 & 0.069 & 3093 & 0.47\\
Normal with TanH & 25 & 25 & 314 & 0.14 & 0.18 & 0.031 & 0.15 & 0.18 & 0.034 & 1721 & 0.55\\

\\ \hline \\ 
\citep{meronen2020stationary}  & 36 & 36 & 777 & 0.24 & 0.28 & 0.078 & 0.28 & 0.32 & 0.1 & 7311 & 0.24\\
 Normal + HMC & 21 & 21 & 72 & 0.07 & 0.09 & 0.007 & 0.071 & 0.094 & 0.009 & 435 & 0.24\\
 Default + HMC & 42 & 42 & 284 & 0.14 & 0.16 & 0.026 & 0.32 & 0.34 & 0.12 & 1867 & 0.11\\

\\ \hline \\ 
\textbf{Ours: pretrained 1D} & 25 & 25 & 73 & 0.06 & 0.09 & 0.008 & 0.076 & 0.1 & 0.011 & 771 & 0.07\\
\textbf{Ours: trained 1D} & 23 & 23 & 6 & 0.03 & 0.03 & 0.001 & 0.029 & 0.035 & 0.001 & 45 & 0.03\\
\textbf{Ours: trained 2D} & 23 & 23 & 13 & 0.03 & 0.03 & 0.001 & 0.028 & 0.035 & 0.001 & 74 & 0.02\\
\end{tabular}
\label{tab:e2e_twomoons_posteriors}
}
\end{table}

\clearpage
\subsection{Periodic and Conditional Activations}

BNNs with periodic activations were trained on functions sampled from a GP with a Mat\'ern kernel with $\nu = \frac{5}{2}$ and lengthscales $\ell$ of values ranging from $0.01$ to $10.0$. The inputs were randomly sampled from the interval $[-3\ell, 3\ell]$. Each input batch comprised 8 sets $X$, each containing 512 values $x$. For each set, 128 functions were sampled. Adam optimizer, with a learning rate of $0.01$, was employed over 4000 iterations.

For the conditioning of priors and activations, a MLP hypernetwork was utilized. This hypernetwork consisted of 3 hidden layers with respective widths of 128, 32, and 8, and RBF activations. On top of the hidden layers, separate dense heads were deployed for each output—a 4-dimensional head for producing variances and two 10-dimensional heads for generating $\{\psi\}$ and $\{A\}$. Each head projected the output of the last hidden layer to the requested dimensions.

\section{Extensive Comparison Against~\cite{tran2022all}}
\label{sec:extensive}

As is well known, a single-hidden-layer BNN with infinite width approximates a Gaussian Process (GP). We leverage this fundamental result in the context of optimal transport (minimizing the Wasserstein metric) to induce a GP with a desired kernel behavior onto a BNN.

A related approach, formulated as the dual problem of optimal transport, was explored in~\citep{tran2022all}. However, their method utilizes the complex Normalizing Flow priors as default, and as such, does not incorporate any mathematical property that guarantees the resulting model is a GP. From this perspective, the approach in~\citep{tran2022all} can be seen as a heuristic that yields favorable performance -- measured by RMSE and NLL -- within the training data range, but without any formal assurance that the model ultimately corresponds to a GP. Their method relies on a multiple-layer BNN with narrow layers, where the number of layers is chosen empirically based on the specific problem. Although employing a Gaussian prior may encourage GP-like behavior, this property does not necessarily hold when using Normalizing Flow priors.

In the following experiments, we empirically demonstrate that this approach generally fails to converge to a GP or satisfy the theoretical requirements due to the use of narrow layers and Normalizing Flow priors.

\subsection{Usage of Wider BNNs}

First, we investigated the effect of increasing the width of a single-hidden-layer BNN from 50 neurons -- following the typical choice in~\citep{tran2022all} -- to 1024 neurons. Specifically, we compared two types of priors: Gaussian priors and Normalizing Flow priors. Additionally, we examined four different activation functions commonly used in BNNs and evaluated these settings on a 1D regression task.

Numerical results for RMSE, NLL, and the 1-Wasserstein ($\mathcal{W}_1^1$) and 2-Wasserstein ($\mathcal{W}_2^2$) metrics are presented in Tab.\ref{tab:1d_regression_gaussian_prior} and Tab.\ref{tab:1d_regression_nf_prior} for Gaussian and Normalizing Flow priors, respectively. We observe that increasing the layer width generally improves the Wasserstein metrics while maintaining similar or slightly worse RMSE and NLL when using Gaussian priors. This improvement is likely due to better adherence to theoretical assumptions (\textit{see} Tab.~\ref{tab:1d_regression_gaussian_prior}). However, even with a wider layer and a fixed activation function, achieving true GP behavior remains elusive.

\begin{figure}[h!]
\centering
\centering
\includegraphics[width=0.85\textwidth]{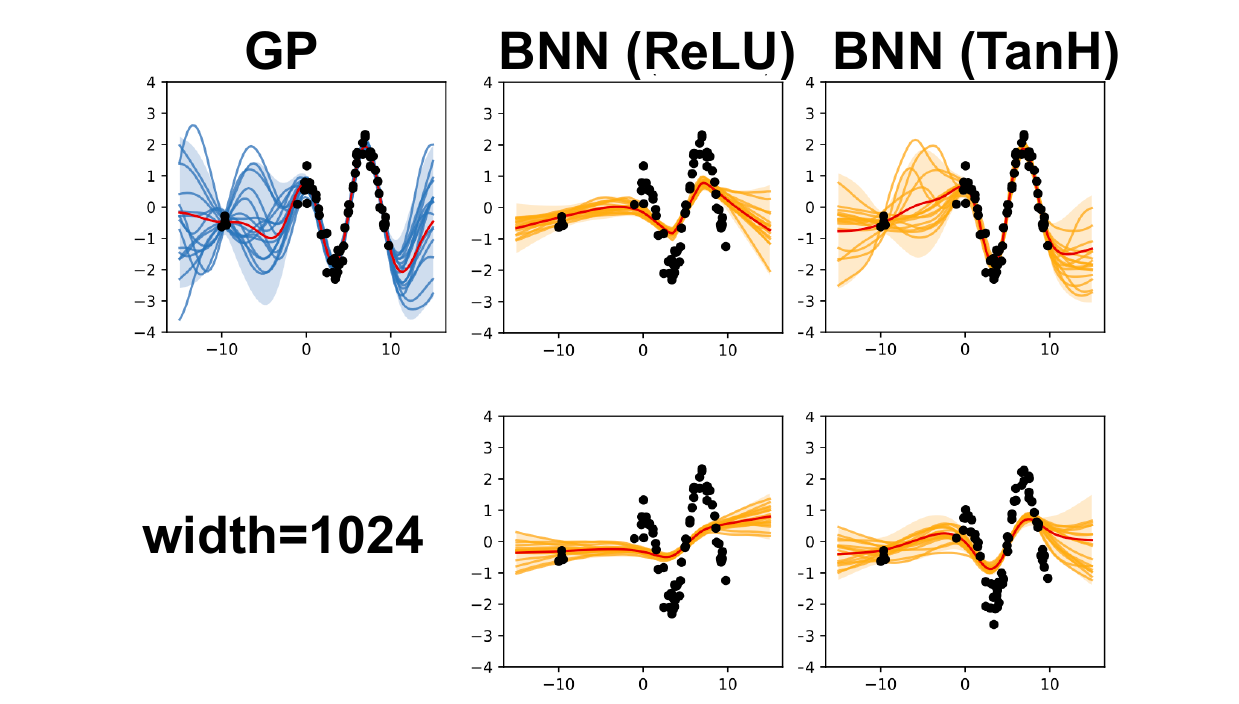} 
\caption{
Comparison of Normalizing Flow prior with a one-layer BNN having a varying number of neurons and ReLU (\textbf{middle column}) or TanH (\textbf{3rd rcolumn}) activation function. We provide a BNN with a narrow hidden layer ($50$ neurons) as the baseline and compare it against a BNN with a wider hidden layer ($1024$ neurons). We observe that even wider hidden layer does not provide a better GP approximation due to the lack of theoretical guarantees when using non-Gaussian priors.
}
\label{fig:posteriors_norm_flow_width}
\end{figure}

In contrast, using wider layers with Normalizing Flow priors significantly degrades performance across all considered metrics (\textit{see} Tab.\ref{tab:1d_regression_nf_prior}). We hypothesize that this is due to the lack of theoretical foundations for such priors. Furthermore, we visualize the posterior distributions of narrow and wide BNN layers for two activation functions: \textit{ReLU}, which performs poorly even with a Gaussian prior, and \textit{TanH} (\textit{see} Fig.\ref{fig:posteriors_norm_flow_width}). These results indicate that Normalizing Flow priors -- despite being uniquely assigned to each neuron -- fail to capture the increased capacity of wider networks, regardless of the activation function used. This serves as a clear example of where the heuristic proposed by~\citep{tran2022all} collapses.

\begin{table*}[h!]
\caption{Comparison of \cite{tran2022all} using a narrow and a wide hidden layer in a BNN with a set Gaussian prior. We observe that the wider hidden layer usually helps in achieving better Wasserstein metrics and similar or slightly worse RMSE and NLL when using Gaussian priors.}%
    \centering
    \begin{tabular}{lcccccccc}
        \toprule
        $width$ $\rightarrow$
        & \multicolumn{4}{c}{$50$}
        & \multicolumn{4}{c}{$1024$} 
        \\
        \cmidrule(lr){2-5}\cmidrule(lr){6-9}
        $activation$ $\downarrow$ $metric$ $\rightarrow$
        & \multicolumn{1}{c}{\textit{RMSE}} 
        & \multicolumn{1}{c}{\textit{NLL}}
        & \multicolumn{1}{c}{$\mathcal{W}_1^1$} 
        & \multicolumn{1}{c}{$\mathcal{W}_2^2$}
        & \multicolumn{1}{c}{\textit{RMSE}} 
        & \multicolumn{1}{c}{\textit{NLL}}
        & \multicolumn{1}{c}{$\mathcal{W}_1^1$} 
        & \multicolumn{1}{c}{$\mathcal{W}_2^2$}
        \\
        \midrule
        \textit{TanH}  & $0.36$  & $0.42$  & $7.57$  & $7.83$  &  $0.53$  & $0.86$  & $8.04$ & $8.26$\\
        \textit{RBF}  &  $0.25$ & $0.24$  & $7.27$  & $7.61$  &  $0.24$  & $0.25$  & $6.96$ &  $7.16$\\
        \textit{ReLU}  &  $1.16$ & $4.10$  & $13.67$  & $13.82$  & $1.19$  & $4.35$  & $12.45$ & $12.62$\\
        \textit{Swish}  &  $1.21$ & $4.52$  & $13.18$  & $13.34$  & $1.22$  & $4.59$  & $12.61$ & $12.78$\\
        \bottomrule
    \end{tabular}
    \label{tab:1d_regression_gaussian_prior}
\end{table*}

\begin{table*}[h!]
\caption{Comparison of \cite{tran2022all} using a narrow and a wide hidden layer in a BNN with a set Normalizing Flow prior. We notice that using wider layers with Normalizing Flow priors significantly lowers the result on each of the considered metrics.}%
    \centering
    \begin{tabular}{lcccccccc}
        \toprule
        $width$ $\rightarrow$
        & \multicolumn{4}{c}{$50$}
        & \multicolumn{4}{c}{$1024$} 
        \\
        \cmidrule(lr){2-5}\cmidrule(lr){6-9}
        $activation$ $\downarrow$ $metric$ $\rightarrow$
        & \multicolumn{1}{c}{\textit{RMSE}} 
        & \multicolumn{1}{c}{\textit{NLL}}
        & \multicolumn{1}{c}{$\mathcal{W}_1^1$} 
        & \multicolumn{1}{c}{$\mathcal{W}_2^2$}
        & \multicolumn{1}{c}{\textit{RMSE}} 
        & \multicolumn{1}{c}{\textit{NLL}}
        & \multicolumn{1}{c}{$\mathcal{W}_1^1$} 
        & \multicolumn{1}{c}{$\mathcal{W}_2^2$}
        \\
        \midrule
        \textit{TanH}  & $0.28$  & $0.27$  & $6.22$  & $6.53$  &  $0.89$  & $2.32$  & $10.41$ & $10.52$\\
        \textit{RBF}  & $0.25$  & $0.24$  & $5.48$  & $5.75$  & $0.35$ & $0.41$ & $6.00$ & $6.20$\\
        \textit{ReLU}  & $0.89$  & $2.34$  & $9.53$  & $9.67$  & $1.10$ & $3.67$ & $12.29$ & $12.44$\\
        \textit{Swish}  &  $1.11$ & $3.64$  & $13.67$  & $13.82$  & $1.18$  & $4.19$  & $13.71$ & $13.86$\\
        \bottomrule
    \end{tabular}
    \label{tab:1d_regression_nf_prior}
\end{table*}

\clearpage

\subsection{Influence of Number of Hidden Layers in BNNs}

Next, we examine whether deeper BNNs within the~\citep{tran2022all} framework lead to improved results.

\begin{figure}[h!]
\centering
\centering
\includegraphics[width=0.495\textwidth]{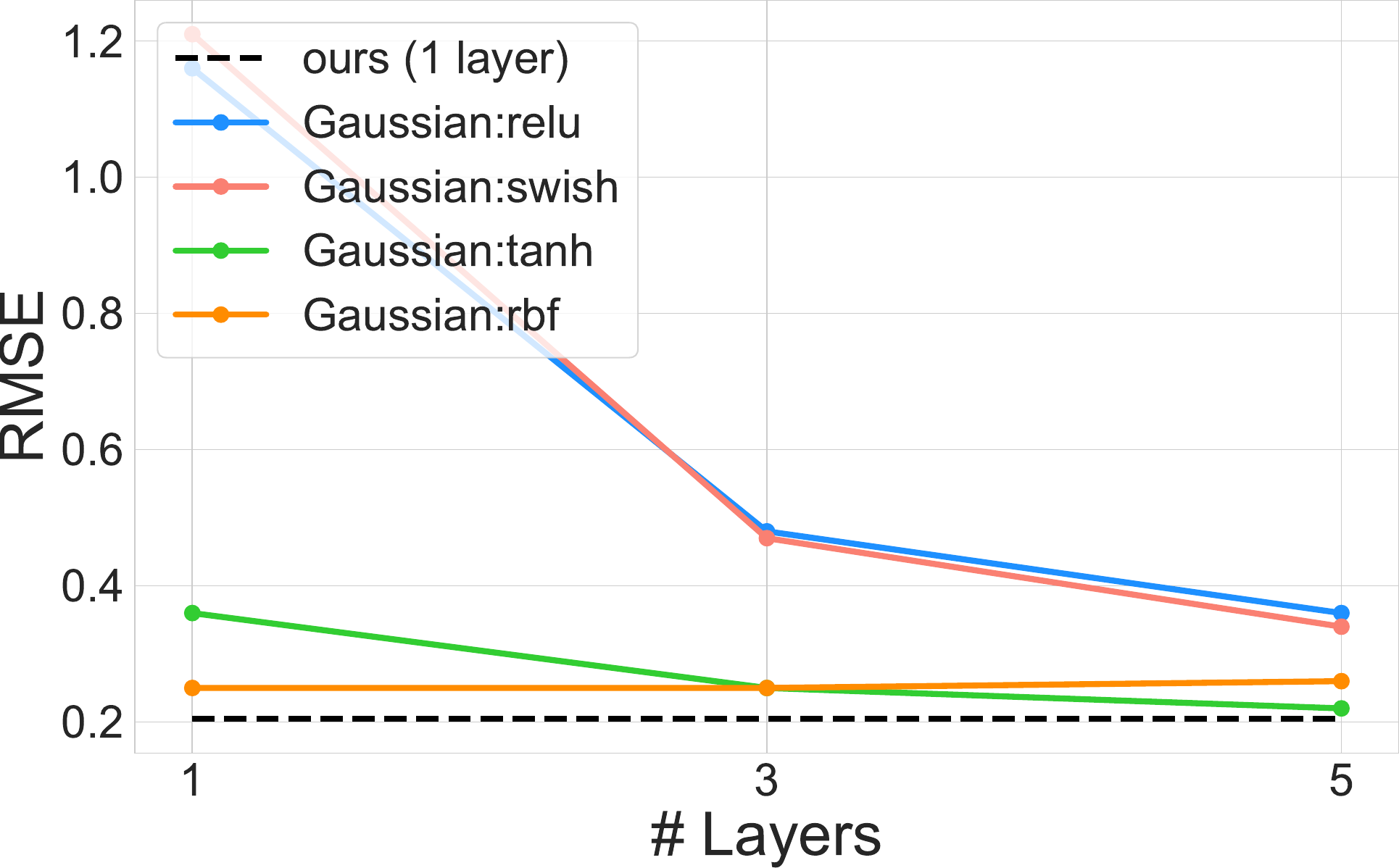}
\includegraphics[width=0.495\textwidth]{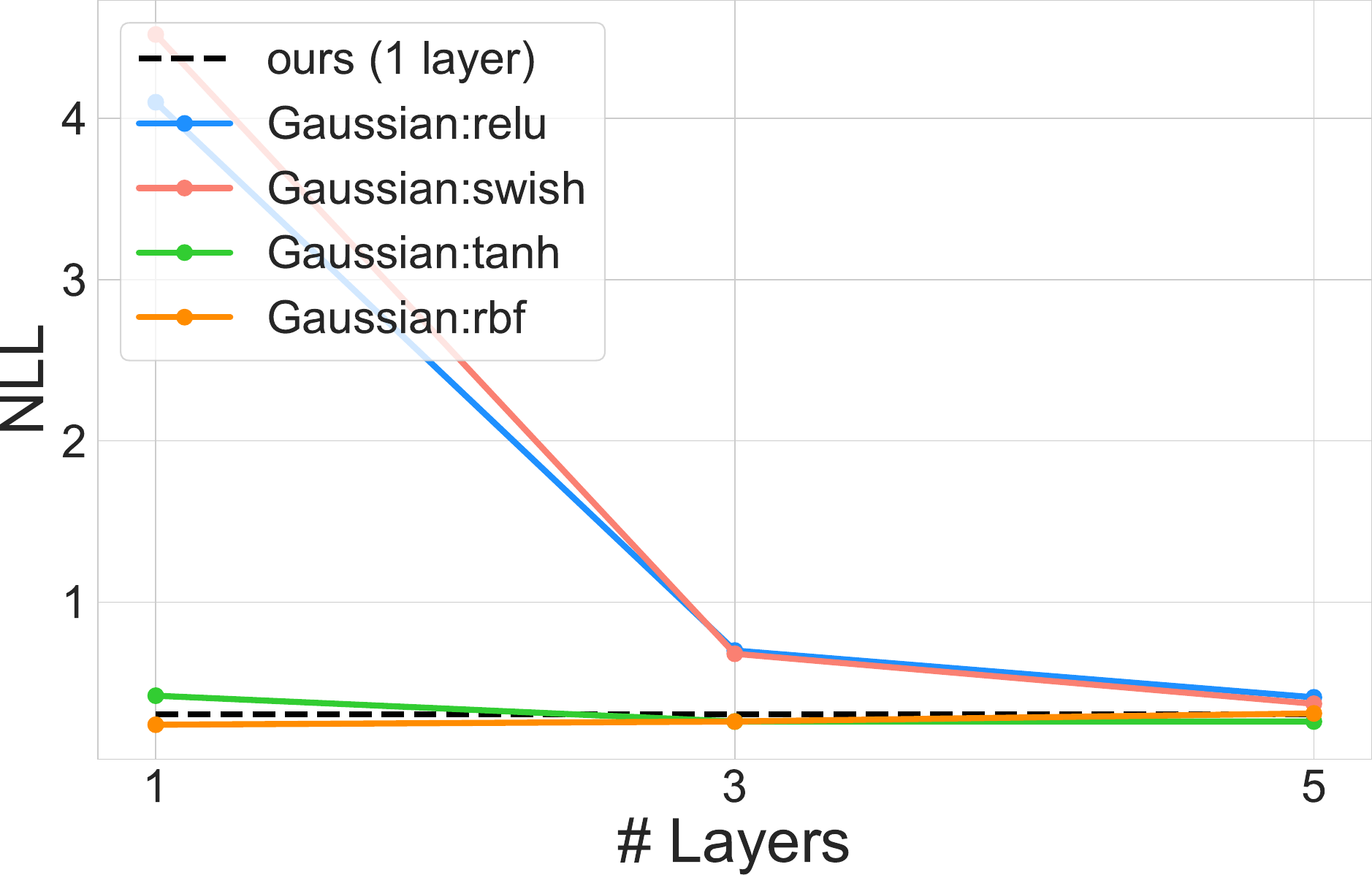}\\
\includegraphics[width=0.495\textwidth]{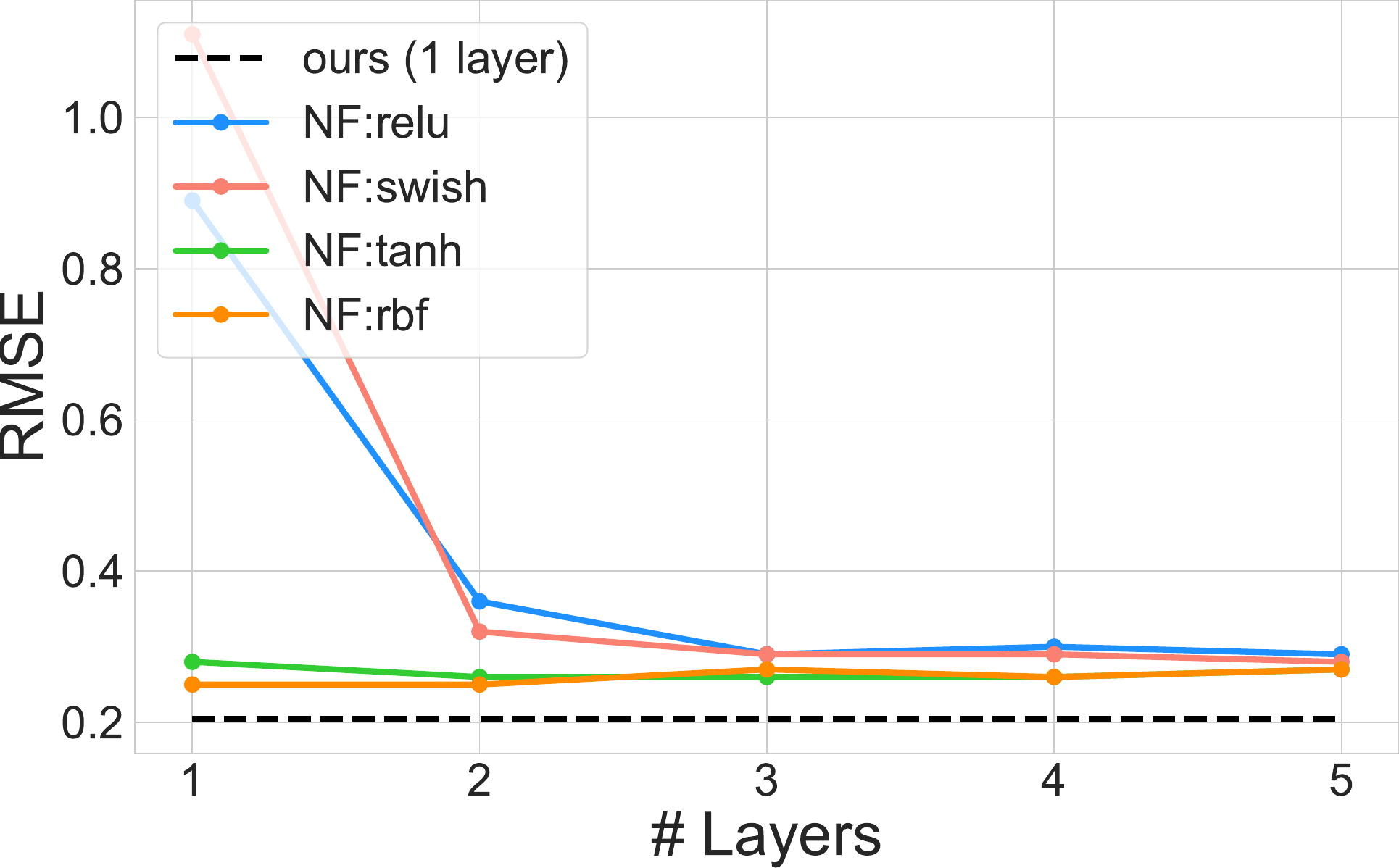}
\includegraphics[width=0.495\textwidth]{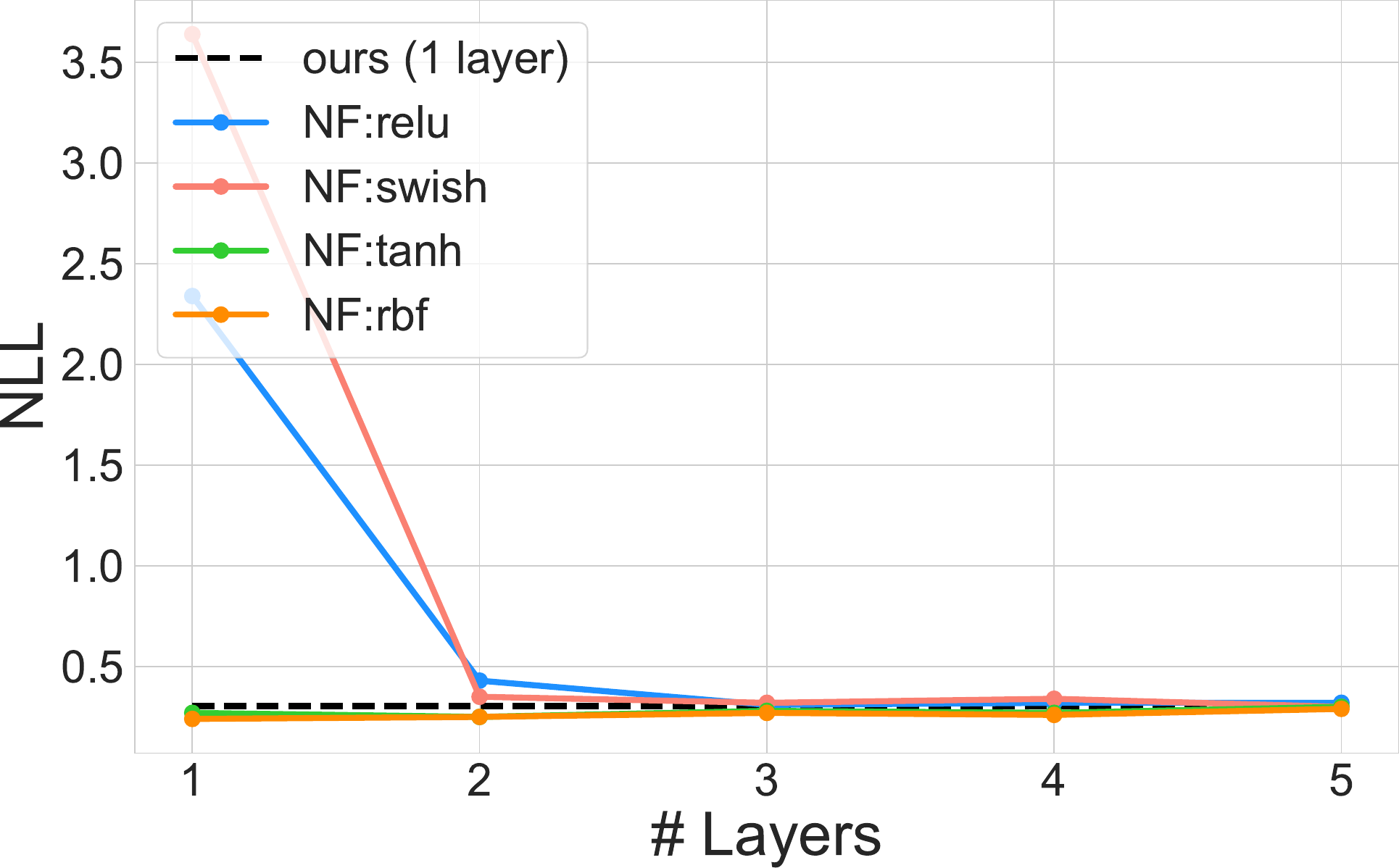}
\caption{
Comparison of different priors, activation functions and BNNs depth in \cite{tran2022all} method in terms of RMSE (\textbf{left column}) and NLL \textbf{right column} on test data. We observe that this method with Gaussian prior (\textbf{top row}) usually get better results with a larger number of hidden layers. Contrary, using Normalizing Flow prior get better results only for some (\textit{worse}) activations. The largest increase of quality might be observed for the worst activation functions. We find that our approach (\textbf{dashed black line}) has better, or similar results in terms of both NLL and RMSE.
}
\label{fig:deep_bnns_rmse_nll}
\end{figure}

Specifically, we compare BNNs with multiple hidden layers (\textit{i.e.,} $\{2, 3, 4, 5\}$) against single-hidden-layer baselines, evaluating four distinct activation functions -- ReLU, TanH, Swish, and RBF -- on RMSE and NLL, as shown in Fig.~\ref{fig:deep_bnns_rmse_nll}.

\begin{figure}[h!]
\centering
\centering
\includegraphics[width=0.85\textwidth]{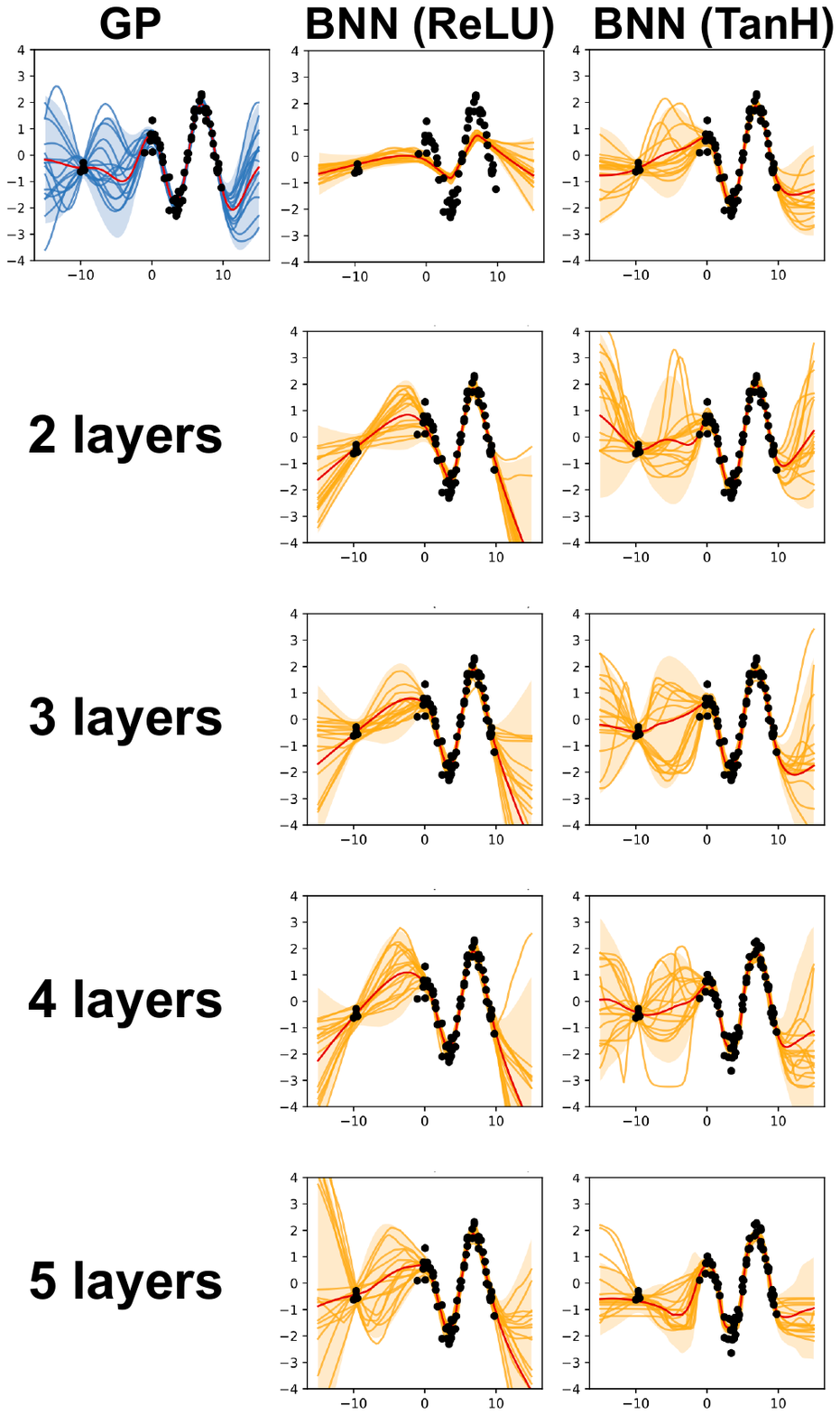} 
\caption{
Comparison of Normalizing Flow prior with a BNN having a varying number of hidden layers and ReLU (\textbf{middle column}) or TanH (\textbf{3rd column}) activation function. We select a single hidden layer ($50$ neurons) BNN as a baseline (\textbf{top row}), which we compare against a multiple hidden layers BNNs ($2, 3, 4,$ or $5$ layers). We notice that using a larger number of hidden layers destroys the posteriors for a BNN with TanH activation function, but for some limit helps when using poor ReLU activation.
}
\label{fig:posteriors_norm_flow_depth}
\end{figure}

Additionally, in Fig.~\ref{fig:posteriors_norm_flow_depth}, we visualize the posterior distributions for two activation functions, ReLU and TanH, under the Normalizing Flow prior to provide further qualitative insights.

\clearpage

\subsection{Out of Distribution Evaluation}

Finally, we evaluate the~\citep{tran2022all} method with different priors and varying numbers of hidden layers on an out-of-distribution (OOD) task. Since the true data distribution in OOD regions is unknown, we rely on optimal transport metrics—1-Wasserstein ($\mathcal{W}_1^1$) and 2-Wasserstein ($\mathcal{W}_2^2$). Additionally, we compare all these configurations against our single-hidden-layer method.

\begin{figure}[h!]
\centering
\centering
\includegraphics[width=0.85\textwidth]{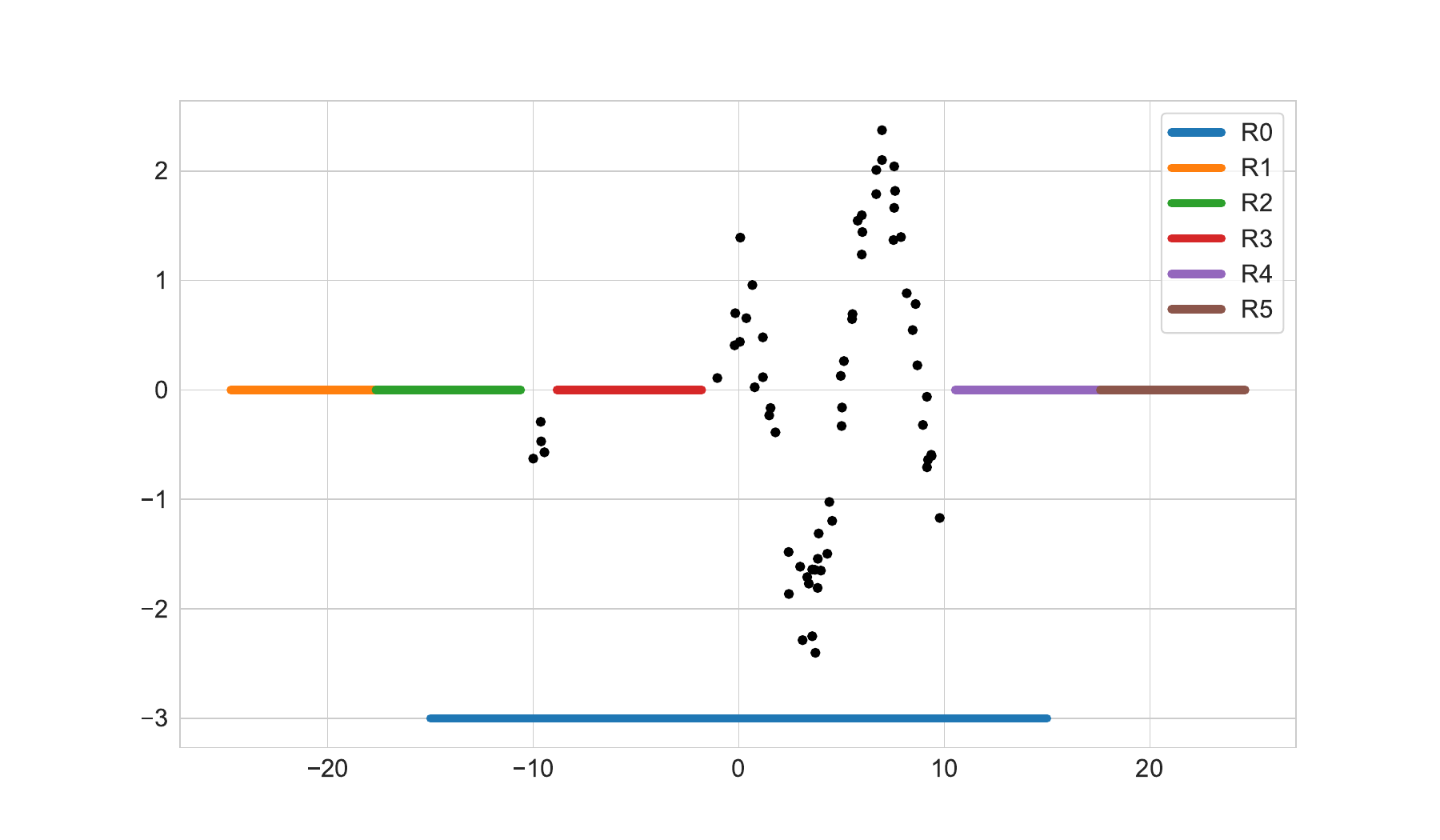} 
\caption{
We compare our method with \cite{tran2022all} on the out of distribution data. For this aim, we provide a wide experimental setting comparing both methods on a few different ranges of data ($x$). We present these ranges as different colour lines in this plot. Firstly, we consider the original test range (\textbf{R0}), then we focus on the data between the given blobs of points (\textbf{R3}). Finally, we consider the out of distribution ranges -- ones being closer to the train data (\textbf{R2}, \textbf{R4}, and \textbf{R2} $\bigcup$ \textbf{R4}) or further (\textbf{R1}, \textbf{R5}, and \textbf{R1} $\bigcup$ \textbf{R5}). The densities of evaluation points meets the density of the original data on each range despite \textbf{R0}. 
}
\label{fig:evaluation_ranges}
\end{figure}

To ensure a comprehensive comparison across various OOD scenarios, we define multiple distinct regions of $x$, where different methods may exhibit different behaviors. These regions are illustrated in Fig.~\ref{fig:evaluation_ranges}.

\begin{figure}[h!]
\centering
\centering
\includegraphics[width=0.33\textwidth]{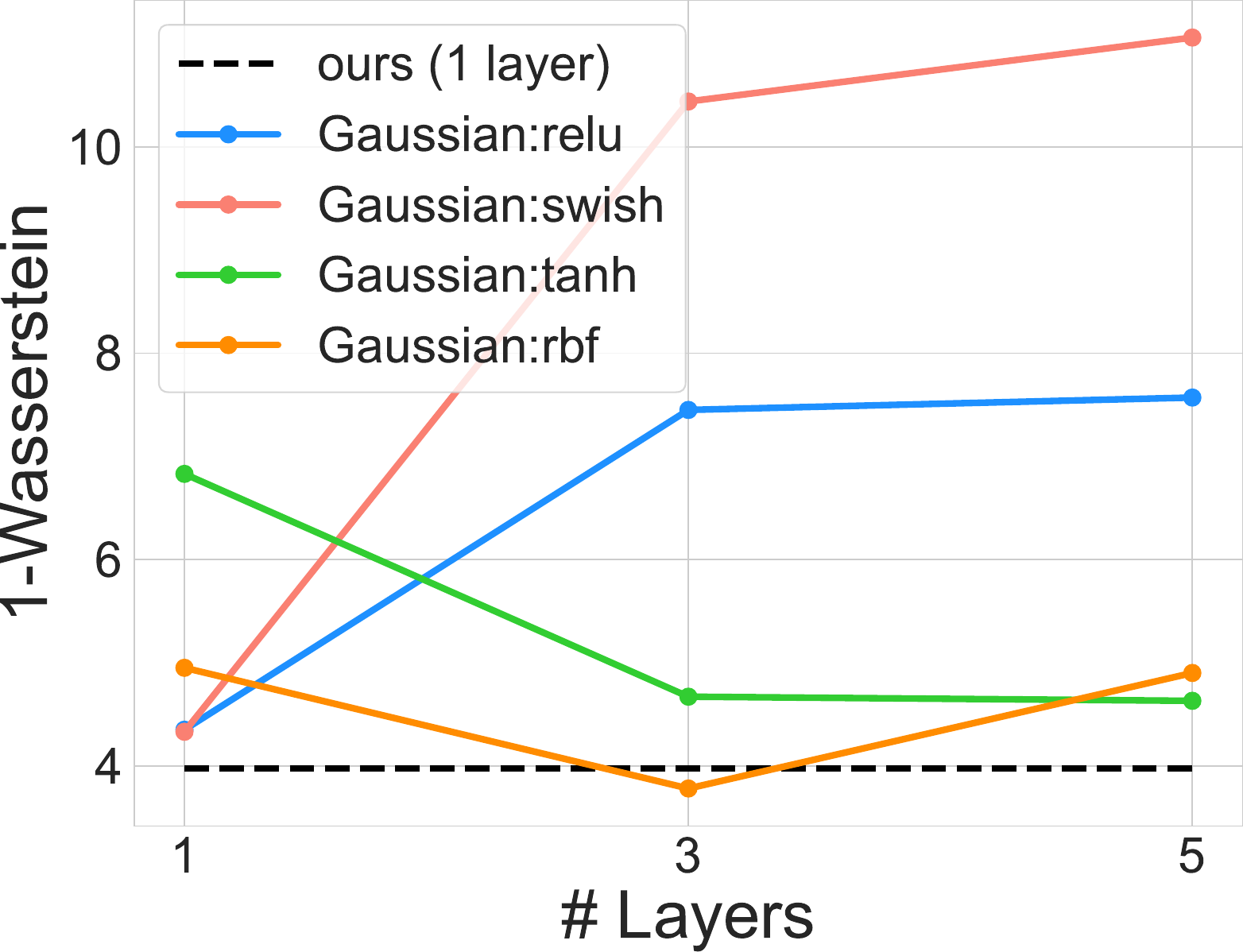} 
\includegraphics[width=0.33\textwidth]{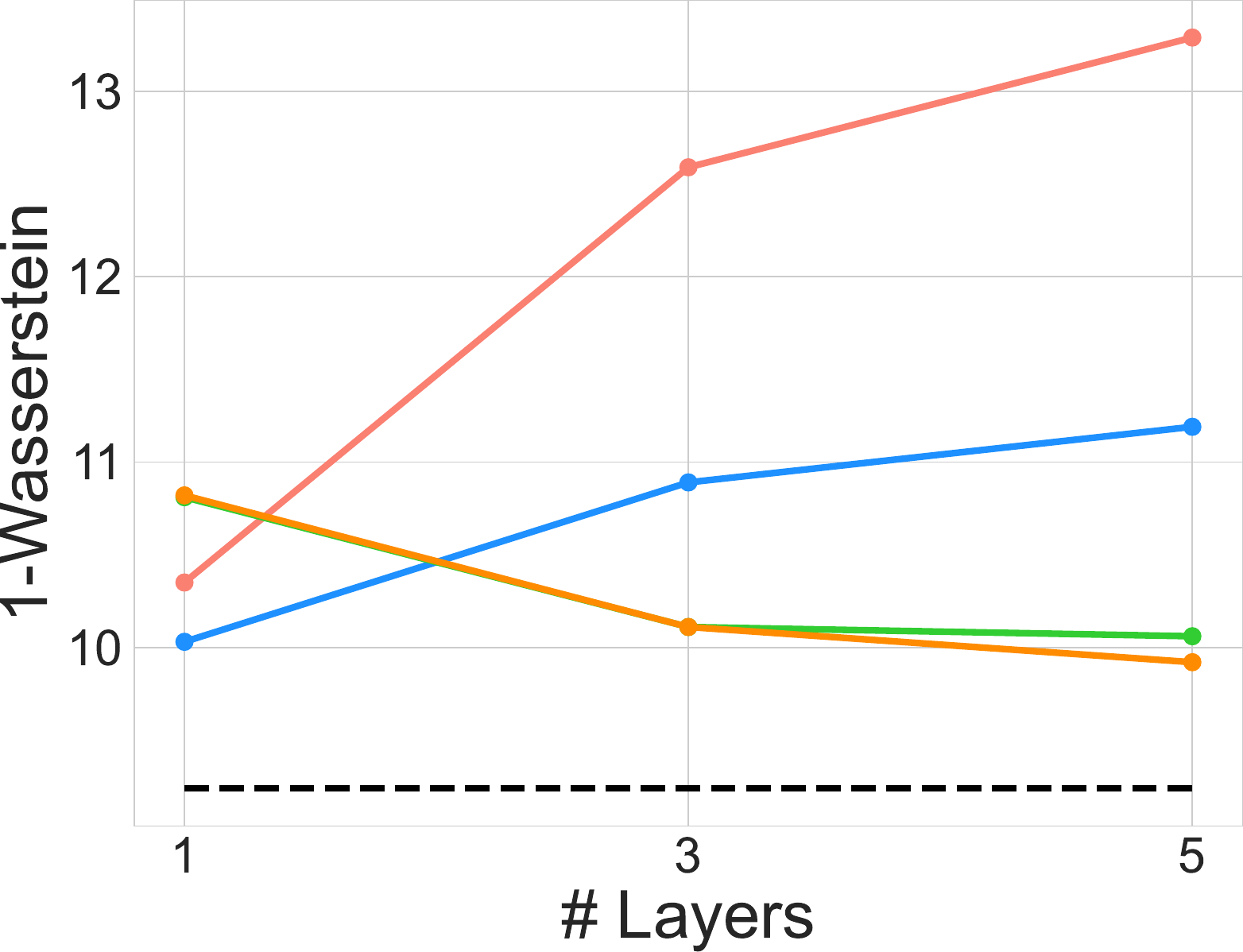} 
\includegraphics[width=0.33\textwidth]{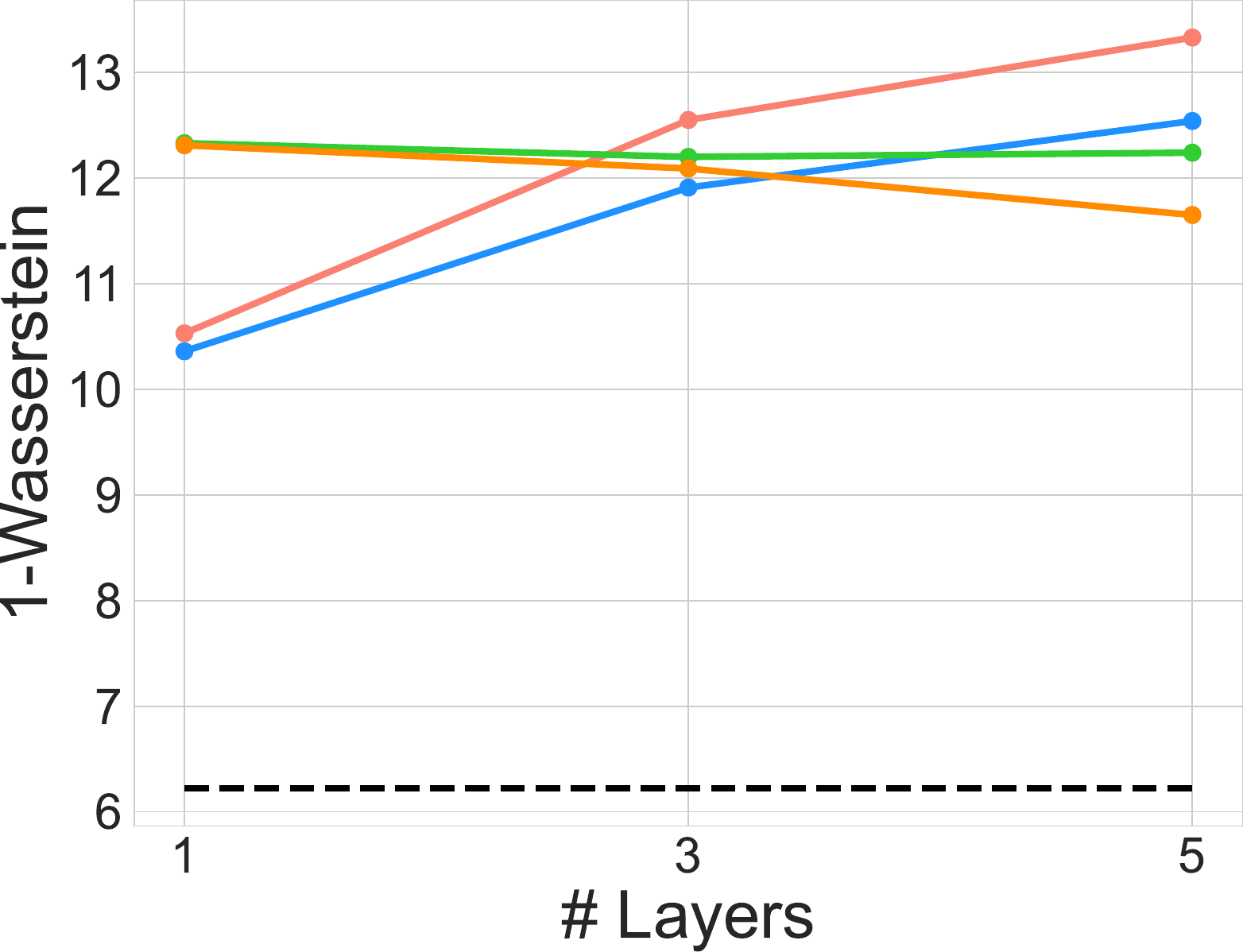} \\
\includegraphics[width=0.33\textwidth]{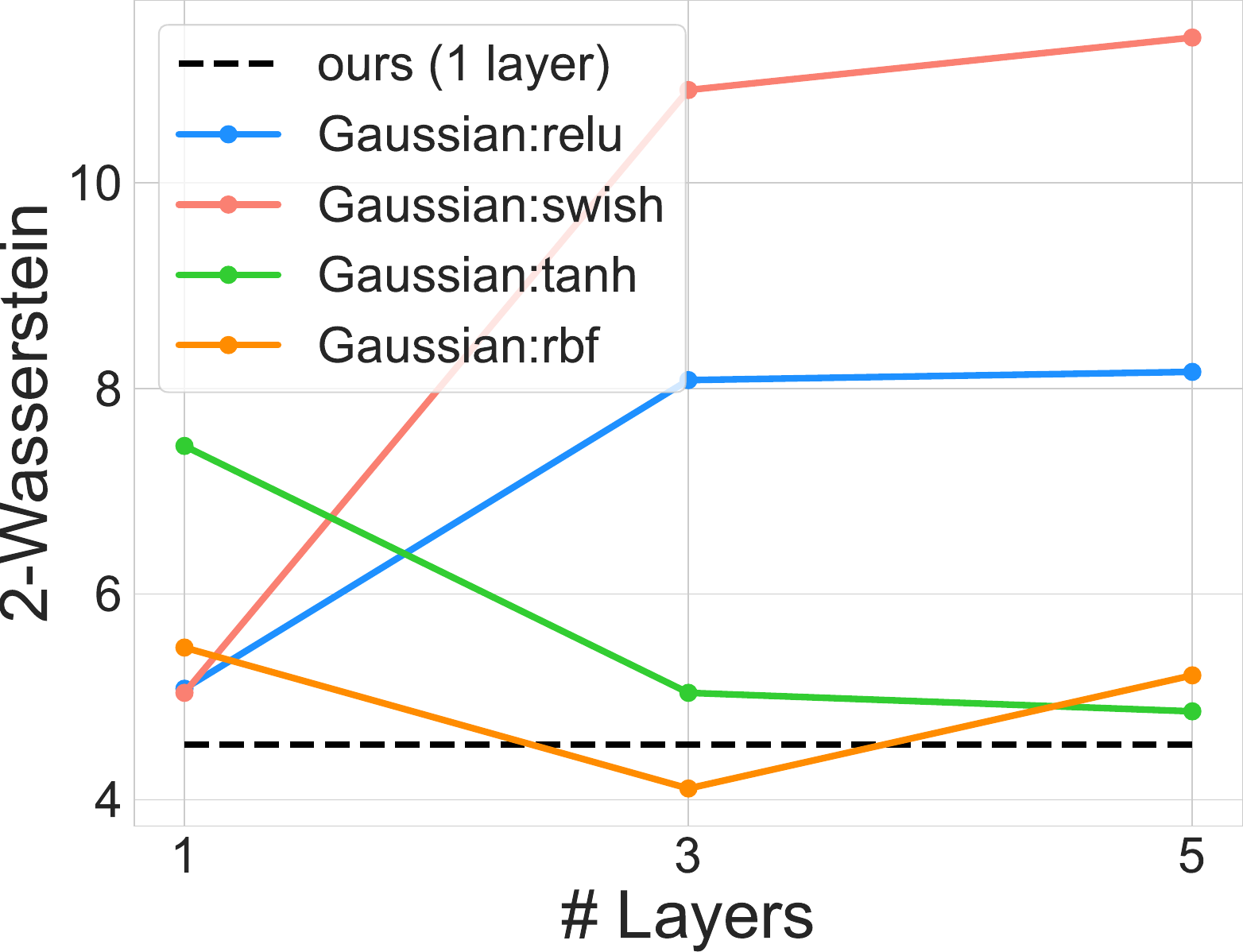} 
\includegraphics[width=0.33\textwidth]{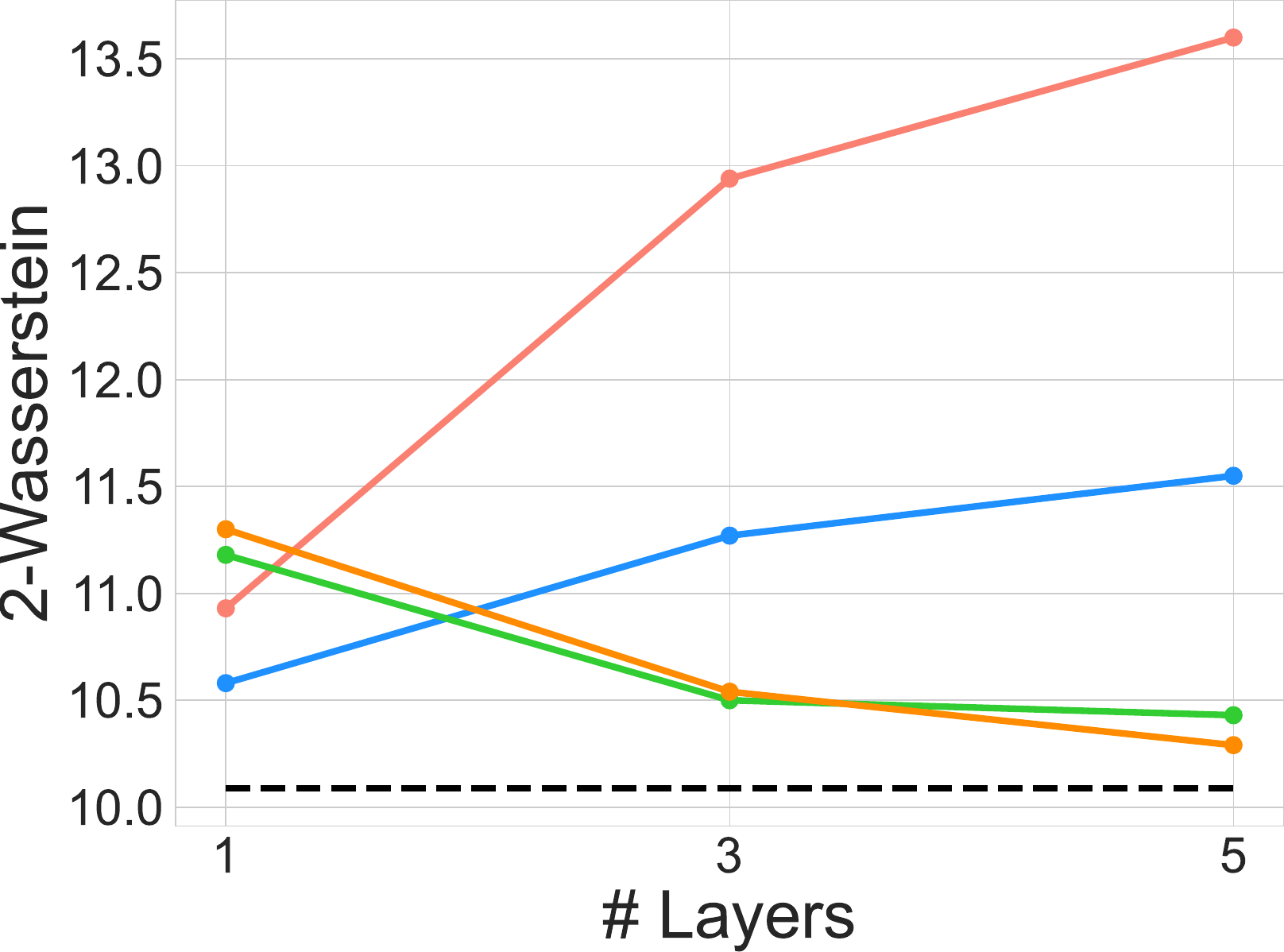} 
\includegraphics[width=0.33\textwidth]{fig/new_pdf_figures_appendix/near_blobs_gaussian_prior_w2_no_legend.pdf} \\
\caption{
Comparison of \cite{tran2022all} method with a Gaussian prior and different number of hidden layers. In this plot, we present $\mathcal{W}_1^1$ (\textbf{top row}) and $\mathcal{W}_2^2$ (\textbf{bottom row}) for the regions \textbf{R3}, \textbf{R4} $\bigcup$ \textbf{R2}, and \textbf{R1} $\bigcup$ \textbf{R5} from left to right side. The appropriate regions might be found in Fig.~\ref{fig:evaluation_ranges}.}
\label{fig:ood_eval_gaussian}
\end{figure}

\begin{figure}[h!]
\centering
\centering
\includegraphics[width=0.245\textwidth]{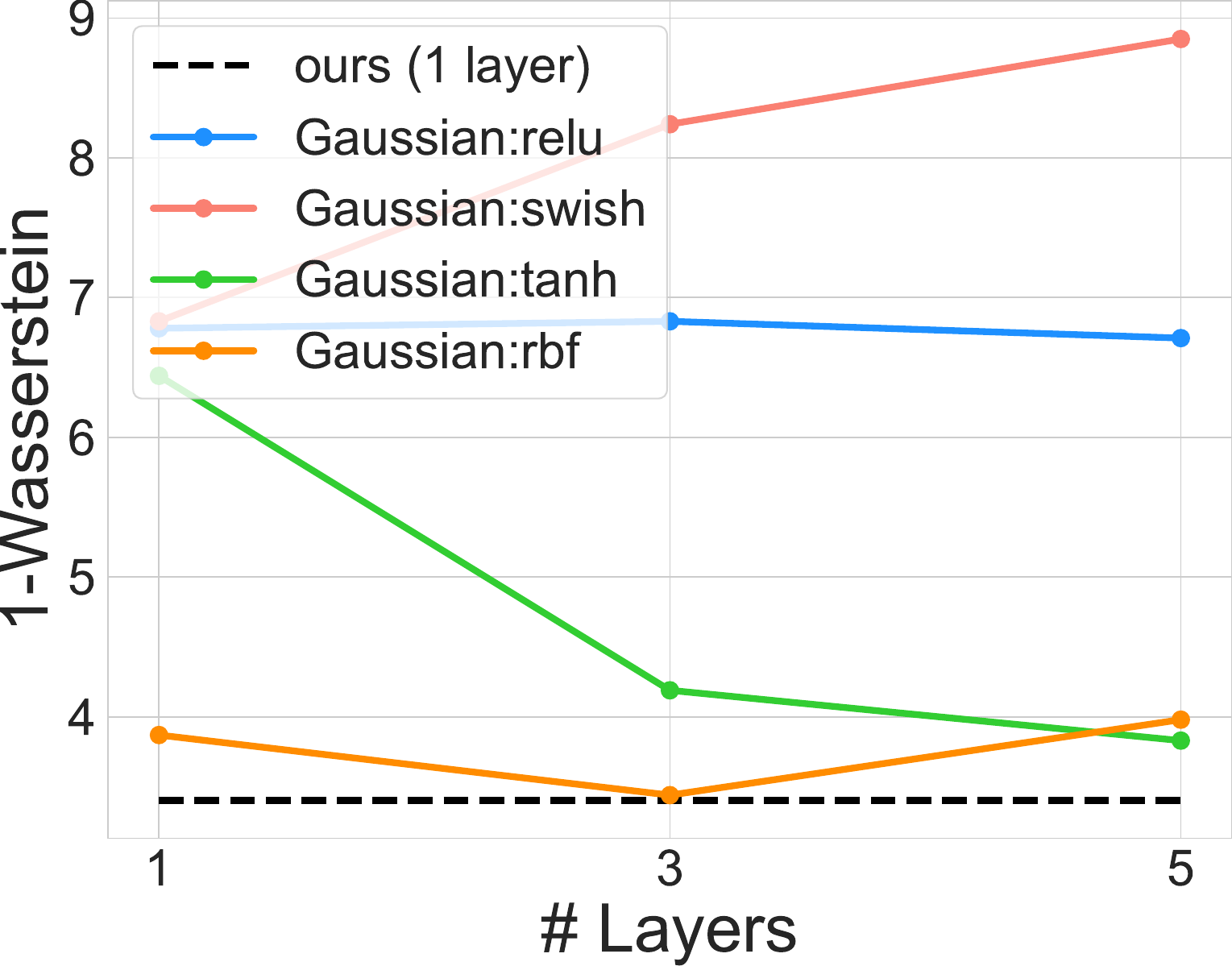} 
\includegraphics[width=0.245\textwidth]{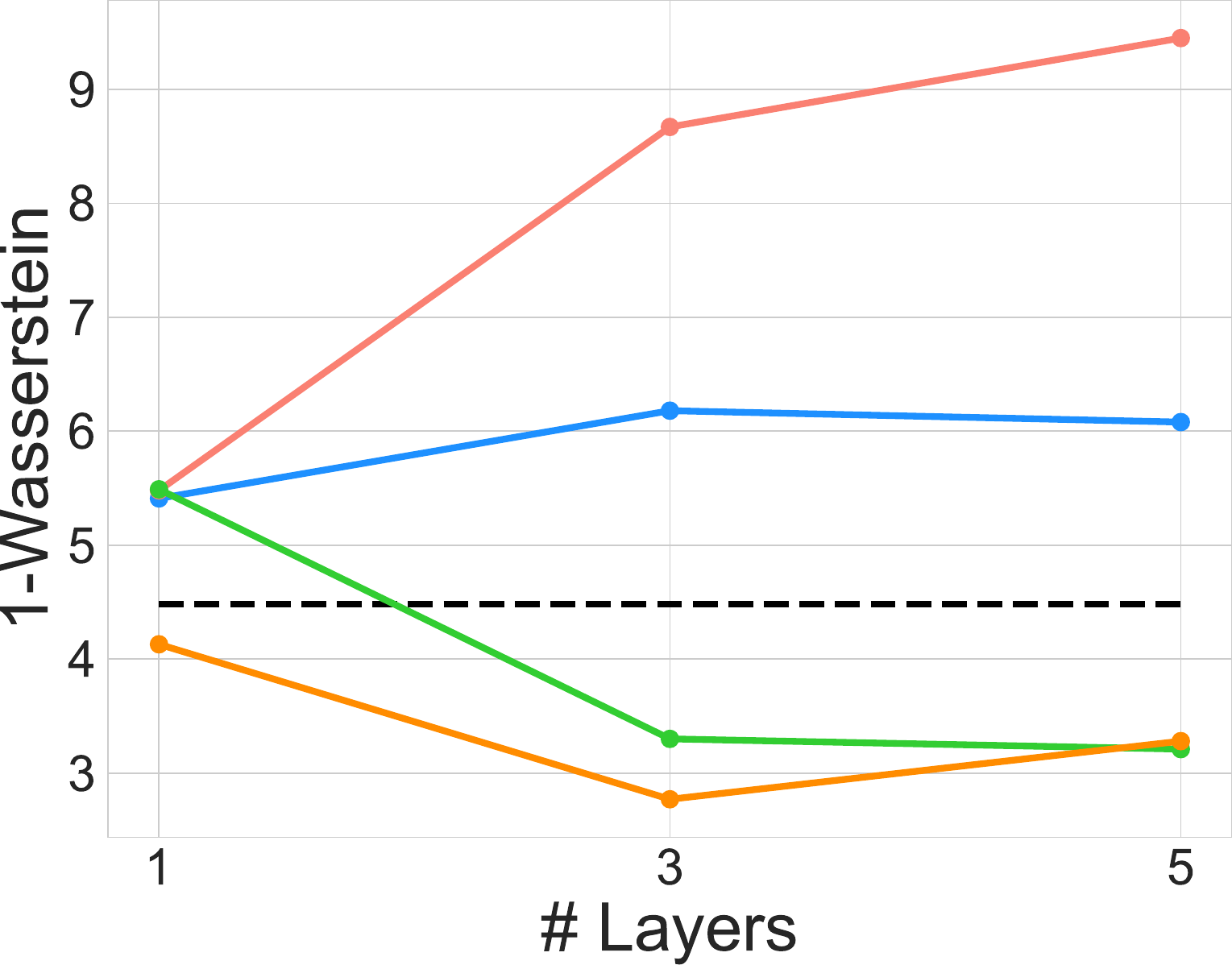} 
\includegraphics[width=0.245\textwidth]{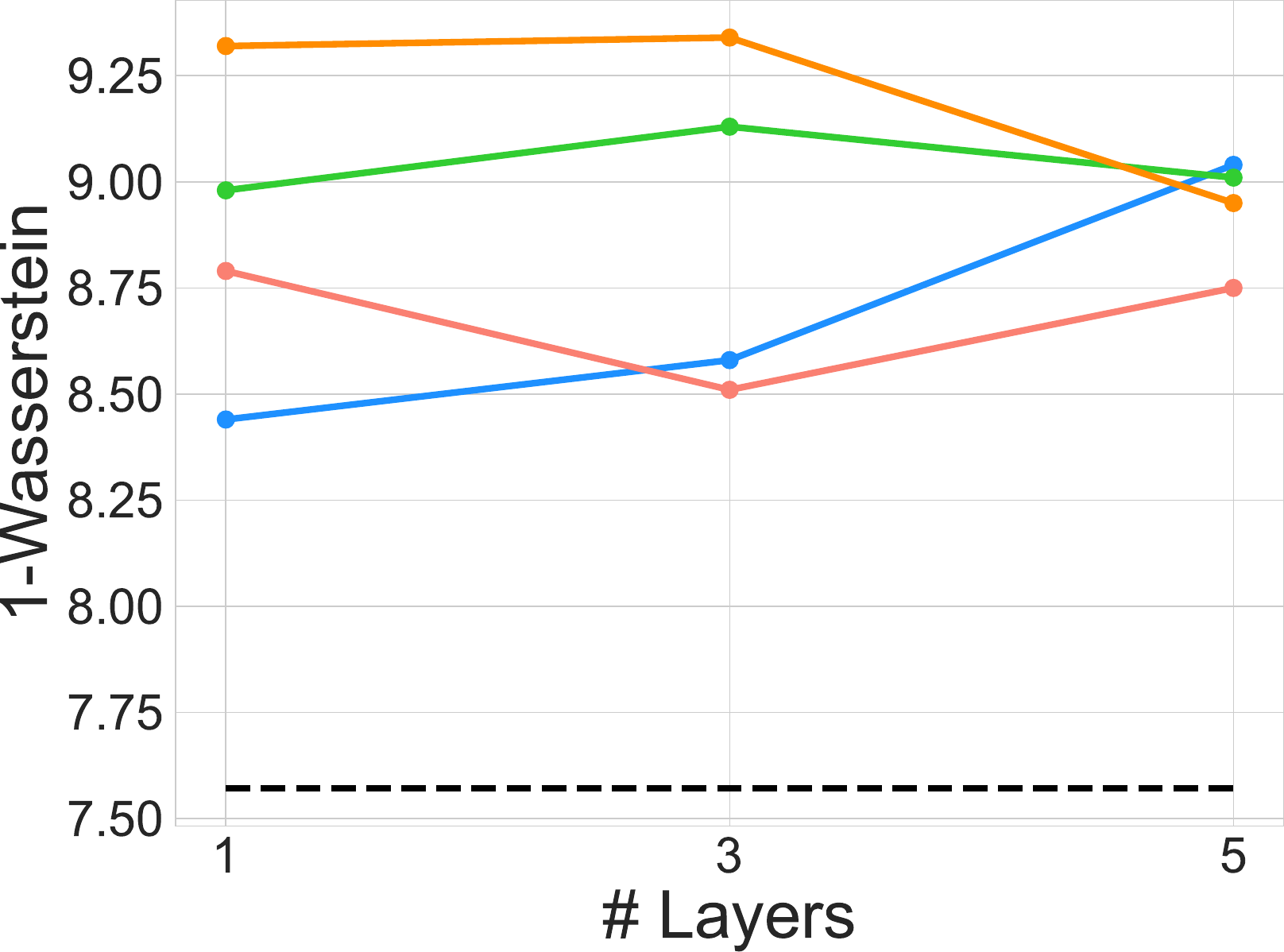}
\includegraphics[width=0.245\textwidth]{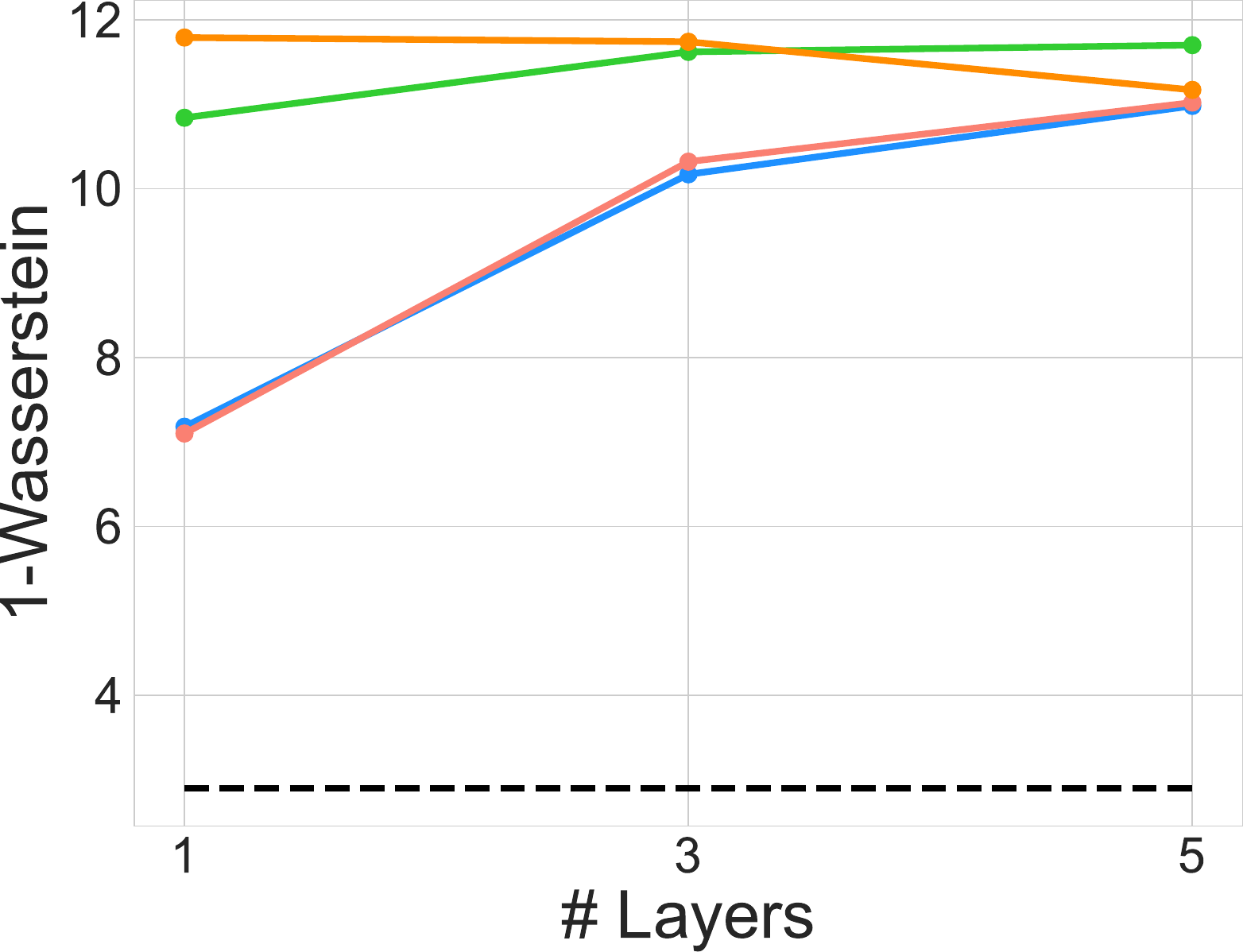}\\
\includegraphics[width=0.245\textwidth]{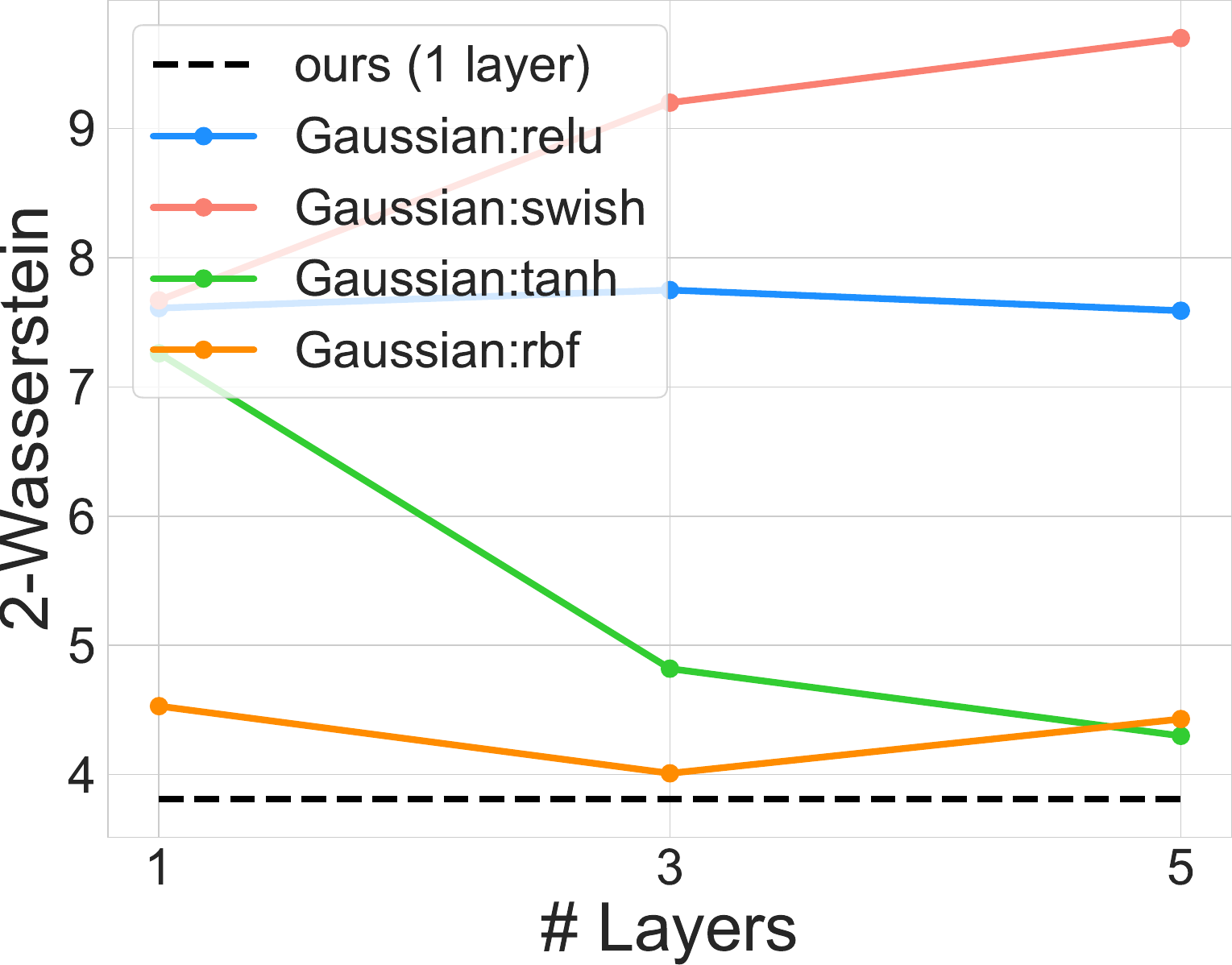} 
\includegraphics[width=0.245\textwidth]{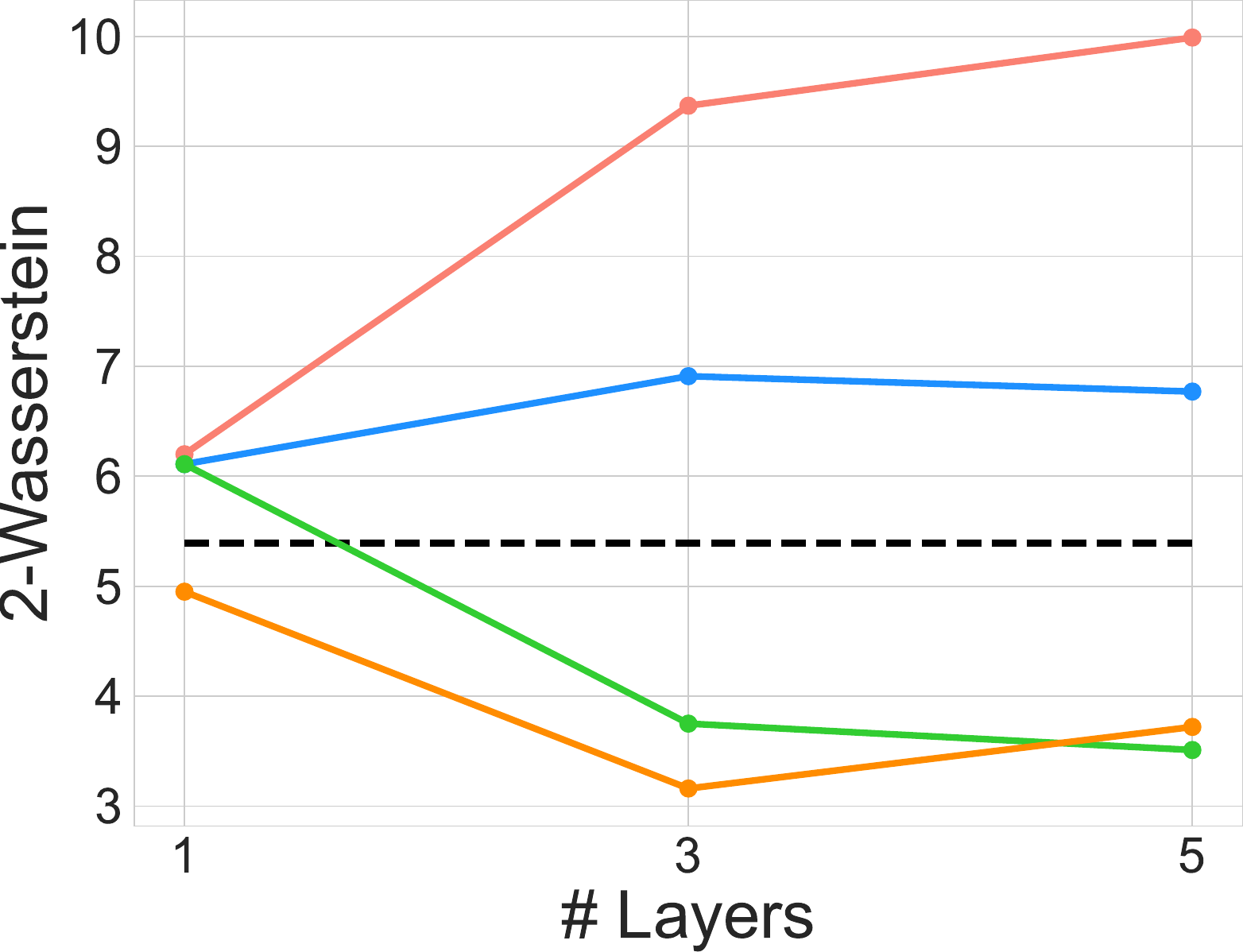} 
\includegraphics[width=0.245\textwidth]{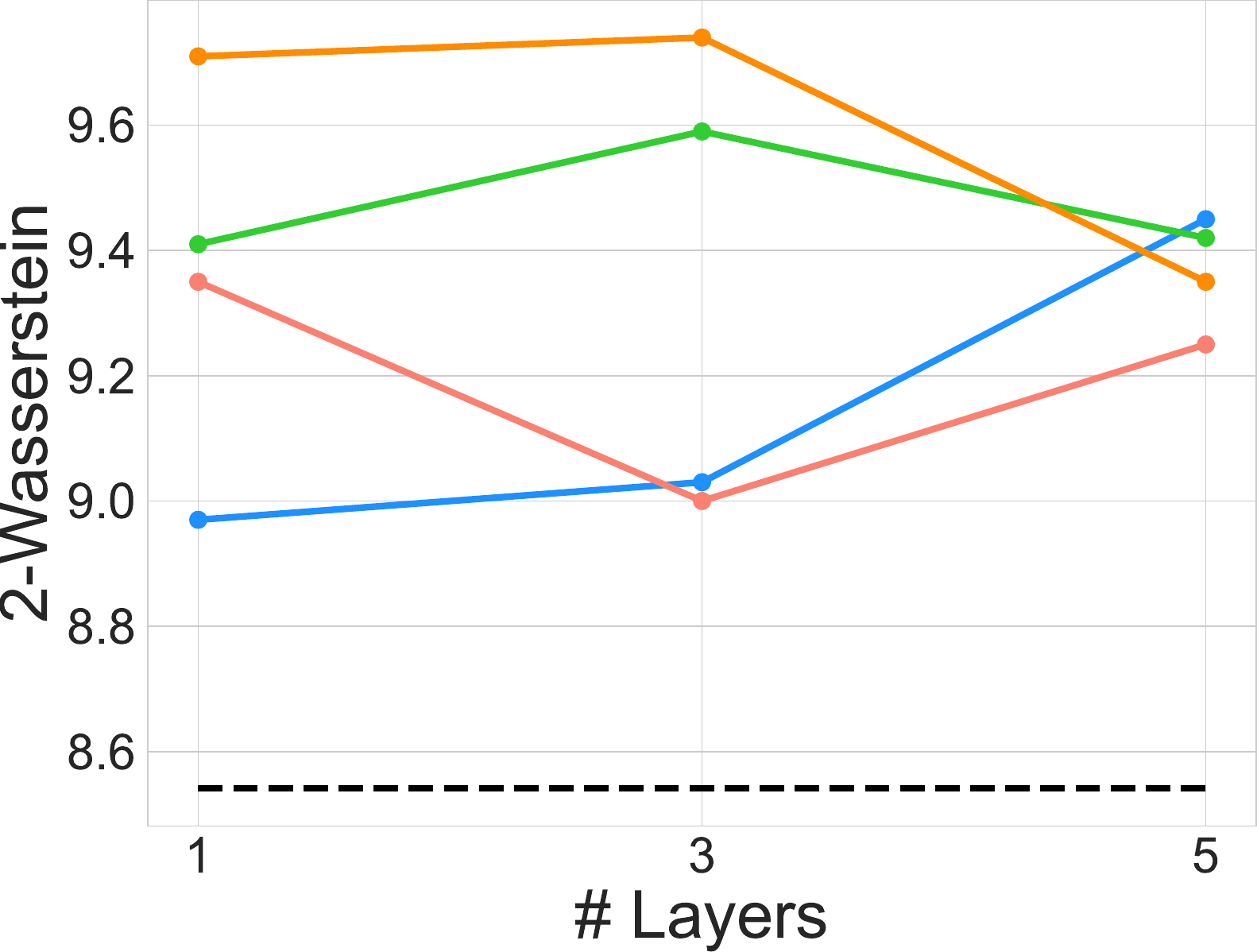} 
\includegraphics[width=0.245\textwidth]{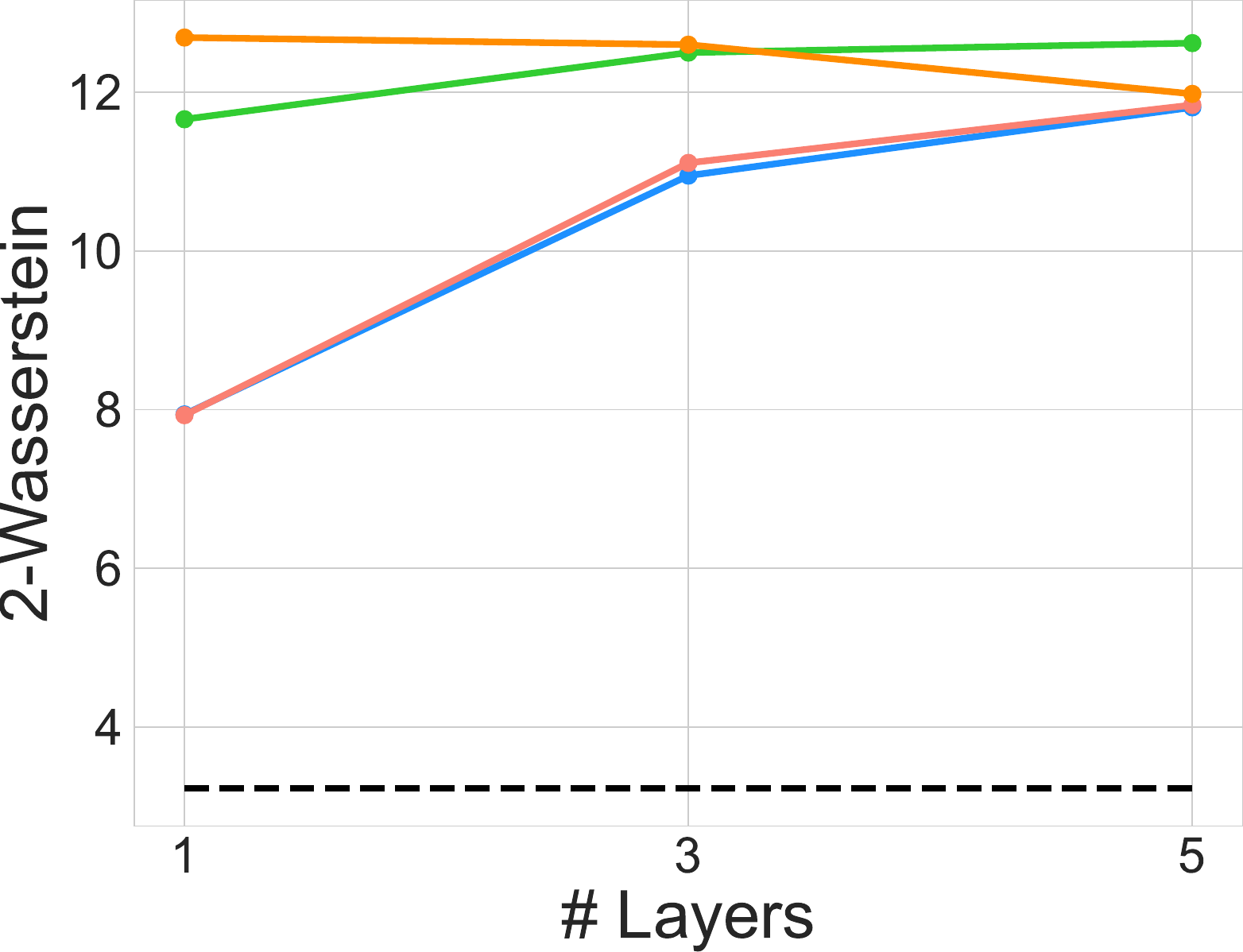}\\
\caption{
Comparison of \cite{tran2022all} method with a Gaussian prior and different number of hidden layers. In this plot, we present $\mathcal{W}_1^1$ (\textbf{top row}) and $\mathcal{W}_2^2$ (\textbf{bottom row}) for the regions \textbf{R1}, \textbf{R2}, \textbf{R4}, and \textbf{R5} from left to right side. The appropriate regions might be found in Fig.~\ref{fig:evaluation_ranges}.}
\label{fig:ood_eval_gaussian_2}
\end{figure}

\begin{figure}[h!]
\centering
\centering
\includegraphics[width=0.33\textwidth]{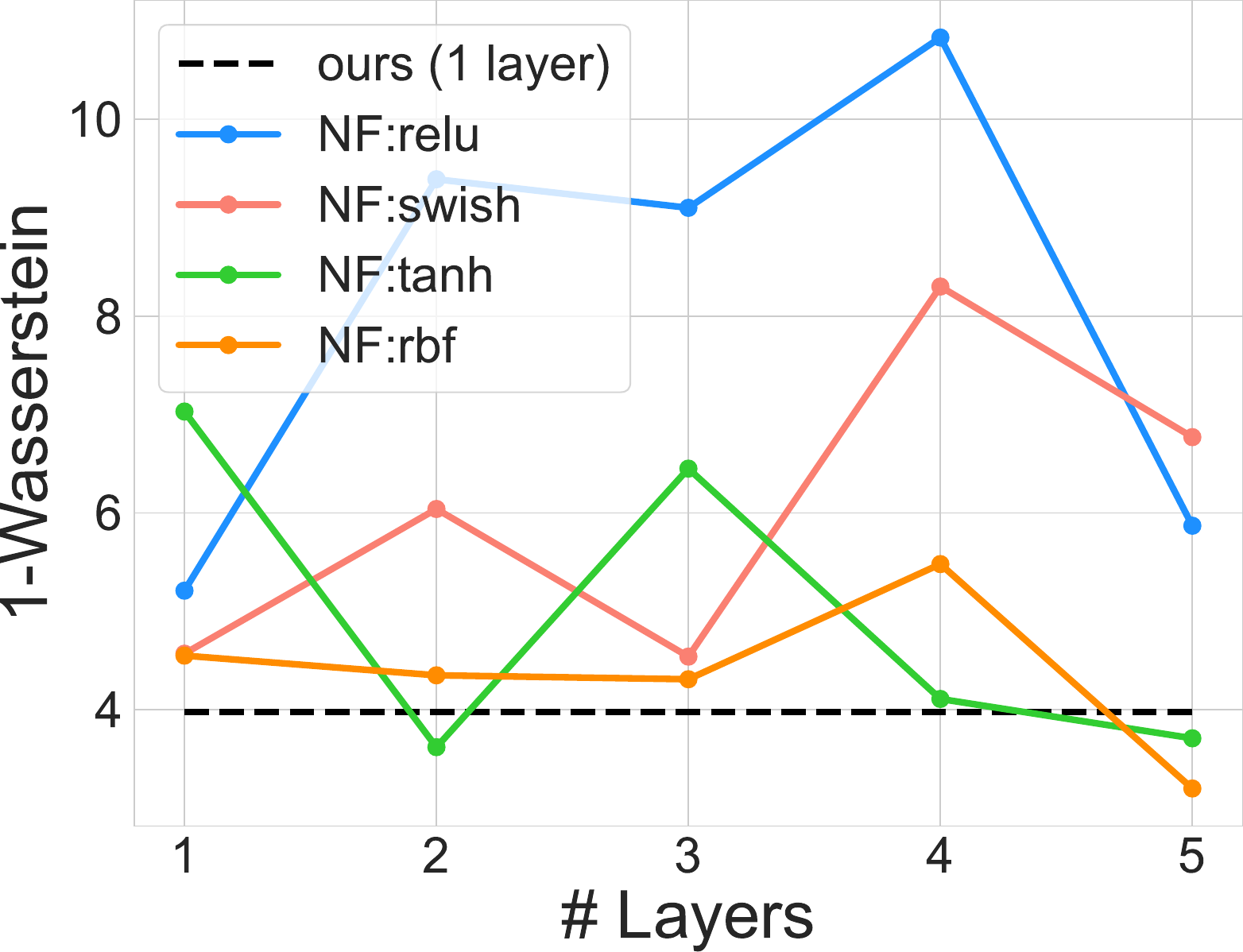} 
\includegraphics[width=0.33\textwidth]{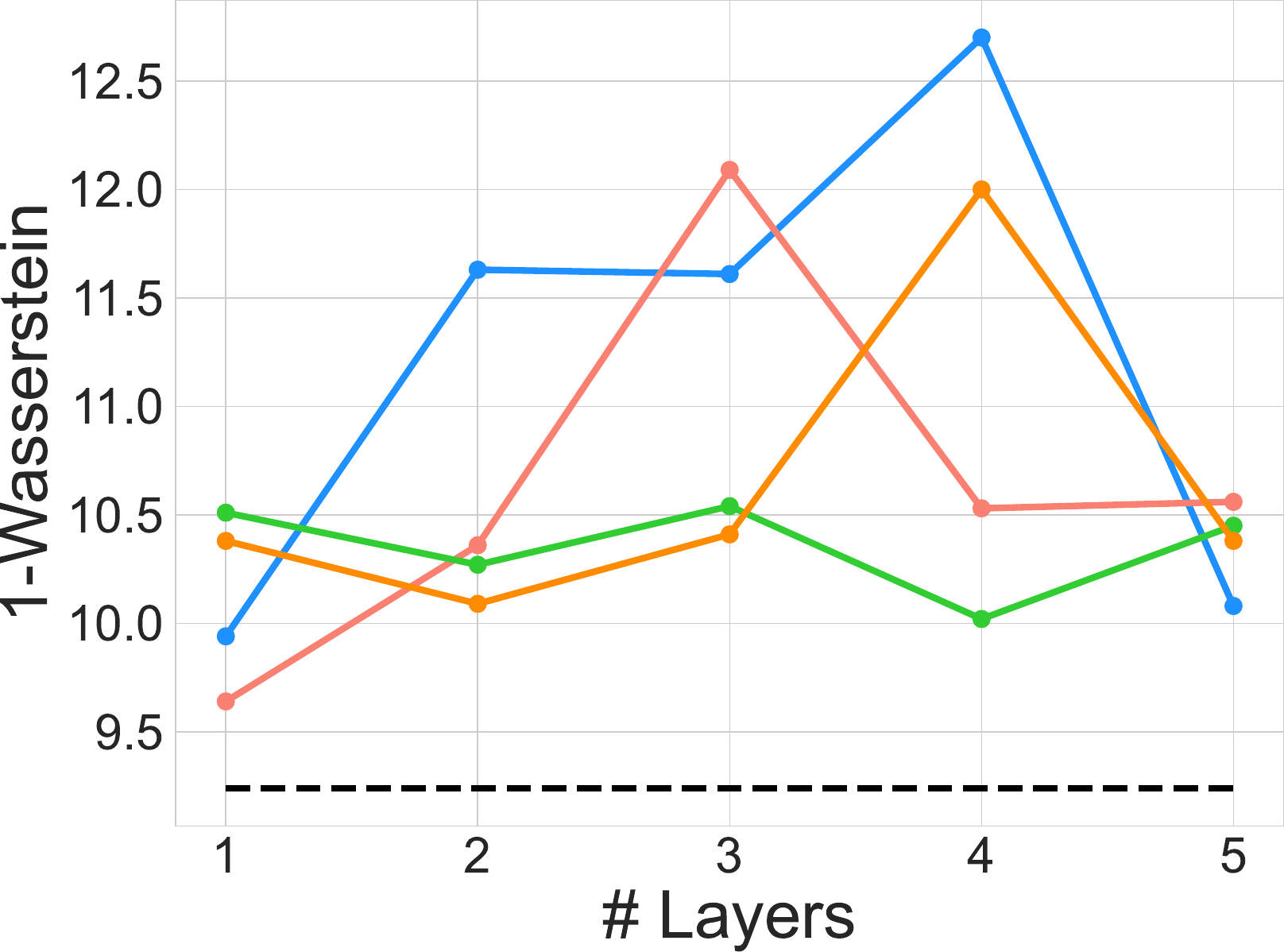} 
\includegraphics[width=0.33\textwidth]{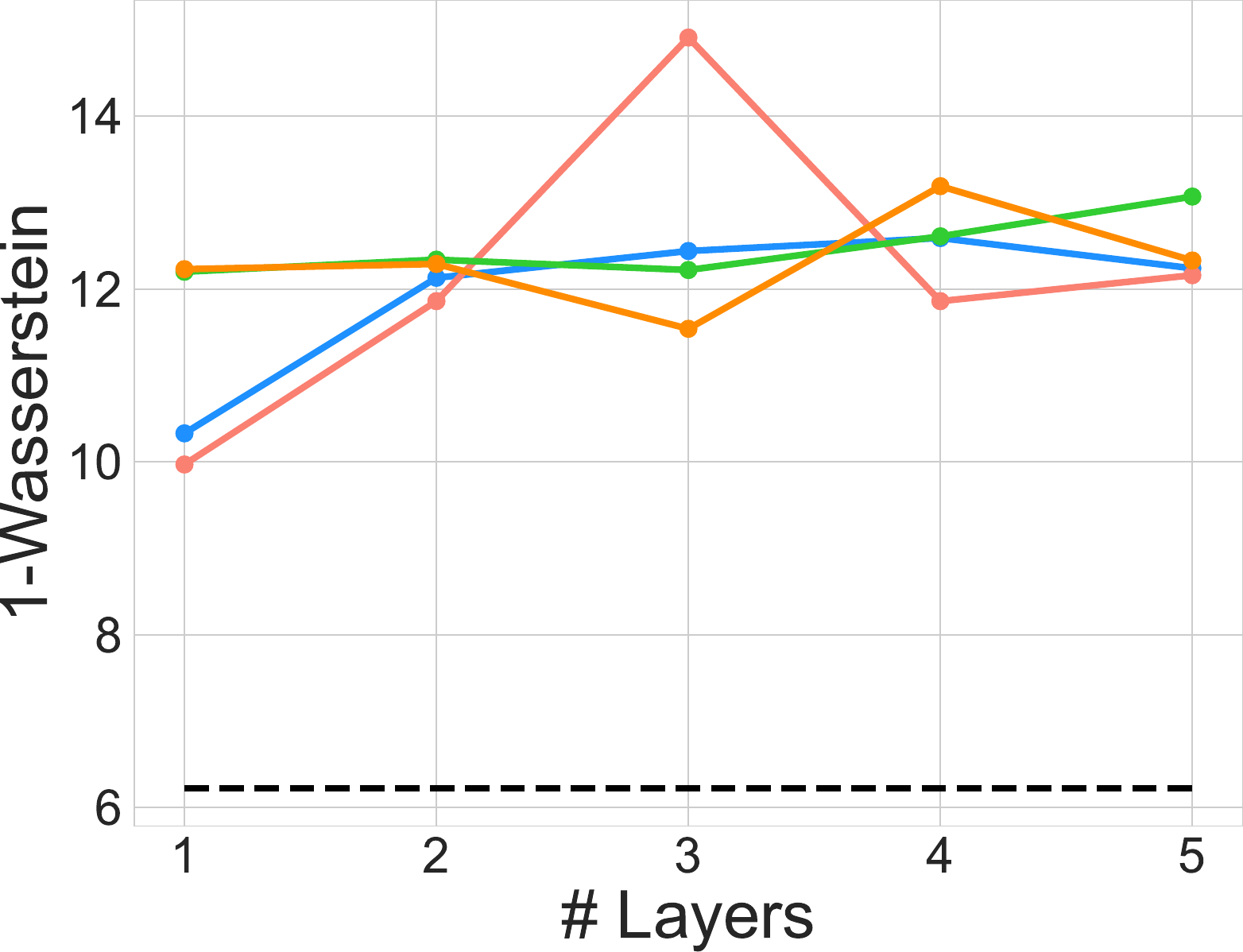} \\
\includegraphics[width=0.33\textwidth]{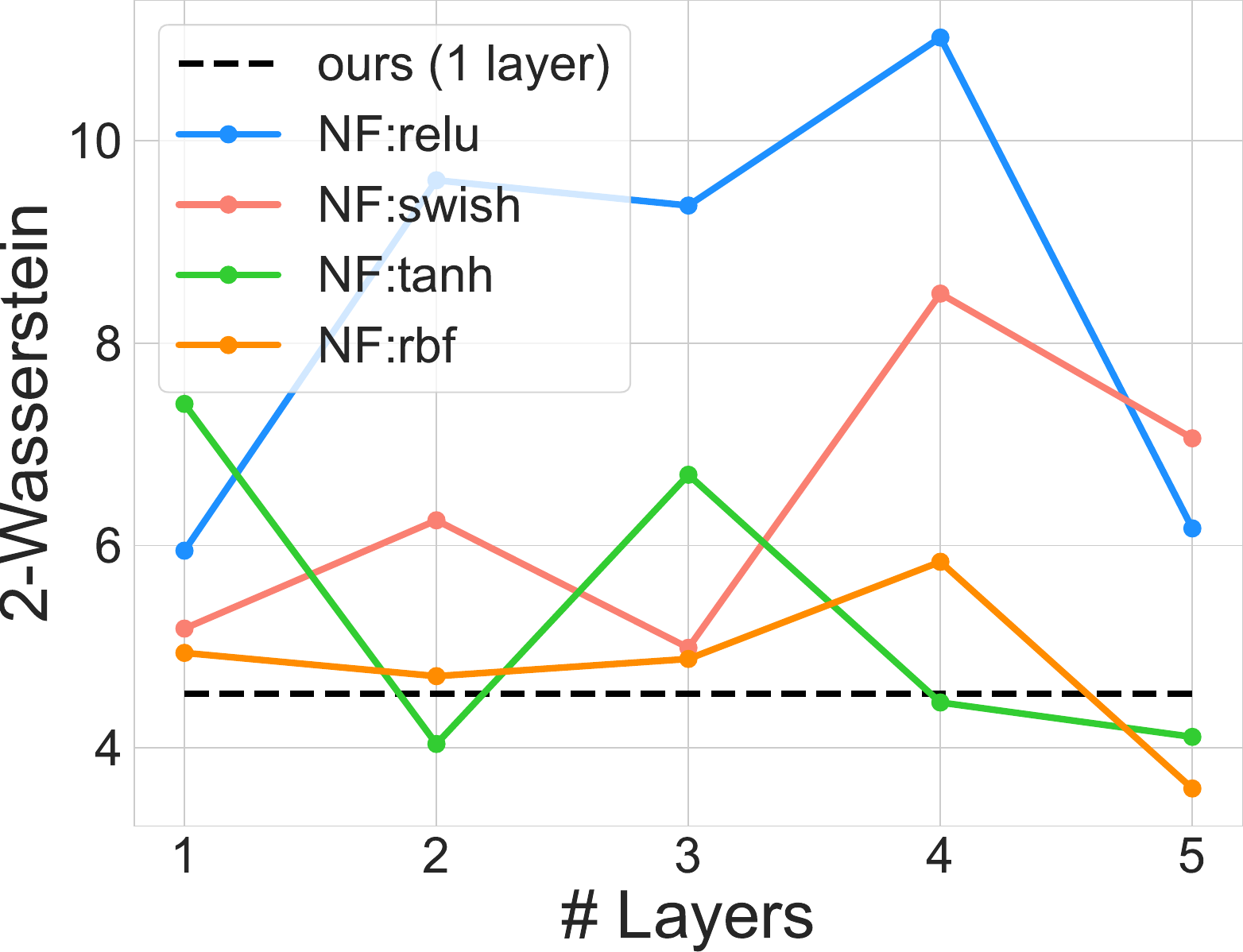} 
\includegraphics[width=0.33\textwidth]{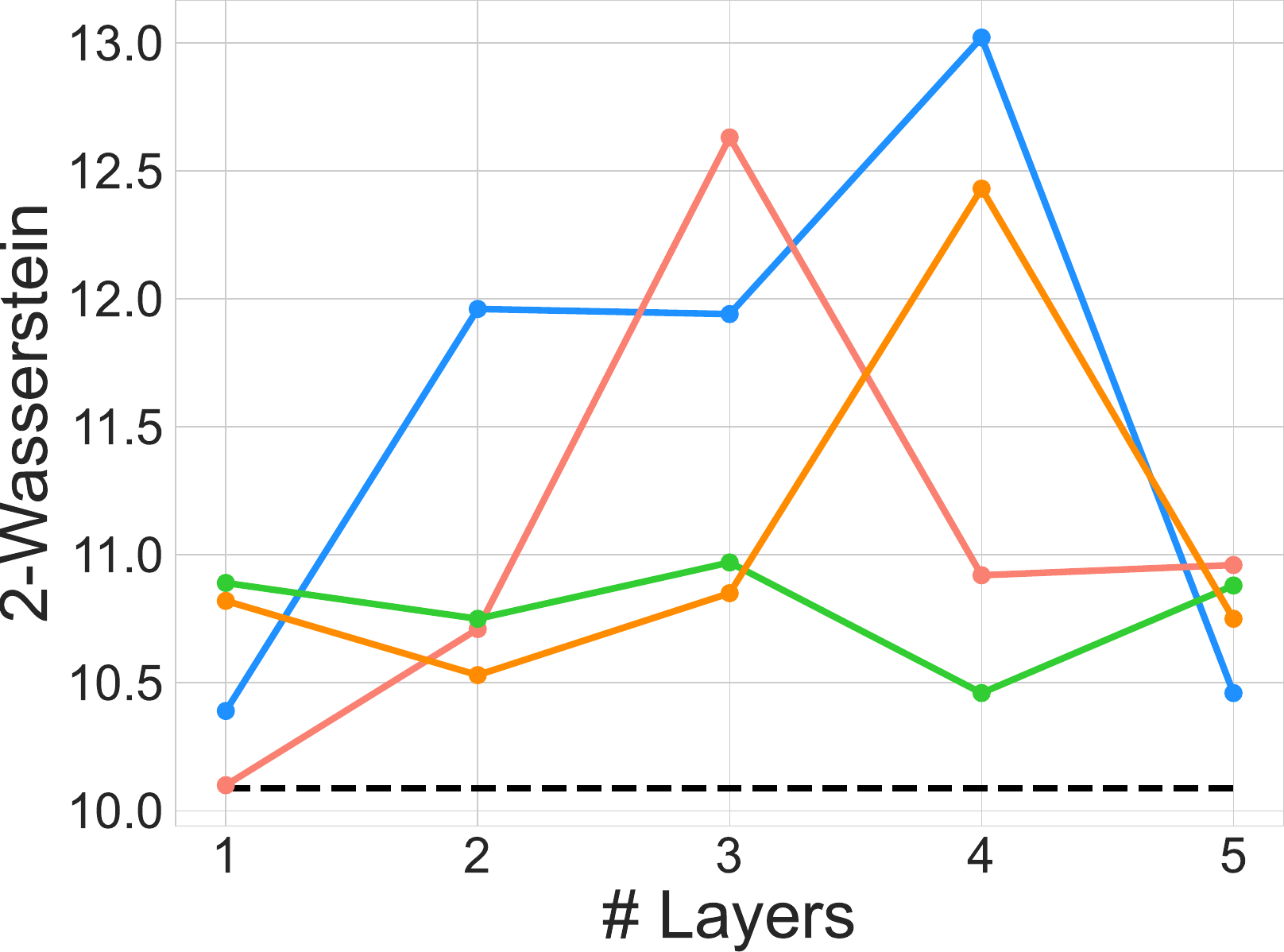} 
\includegraphics[width=0.33\textwidth]{fig/new_pdf_figures_appendix/near_blobs_norm_flow_prior_w2_no_legend.pdf} \\
\caption{
Comparison of \cite{tran2022all} method with a Normalizing Flow prior and different number of hidden layers. In this plot, we present $\mathcal{W}_1^1$ (\textbf{top row}) and $\mathcal{W}_2^2$ (\textbf{bottom row}) for the regions \textbf{R3}, \textbf{R4} $\bigcup$ \textbf{R2}, and \textbf{R1} $\bigcup$ \textbf{R5} from left to right side. The appropriate regions might be found in Fig.~\ref{fig:evaluation_ranges}.}
\label{fig:ood_eval_nf}
\end{figure}

\begin{figure}[h!]
\centering
\centering
\includegraphics[width=0.245\textwidth]{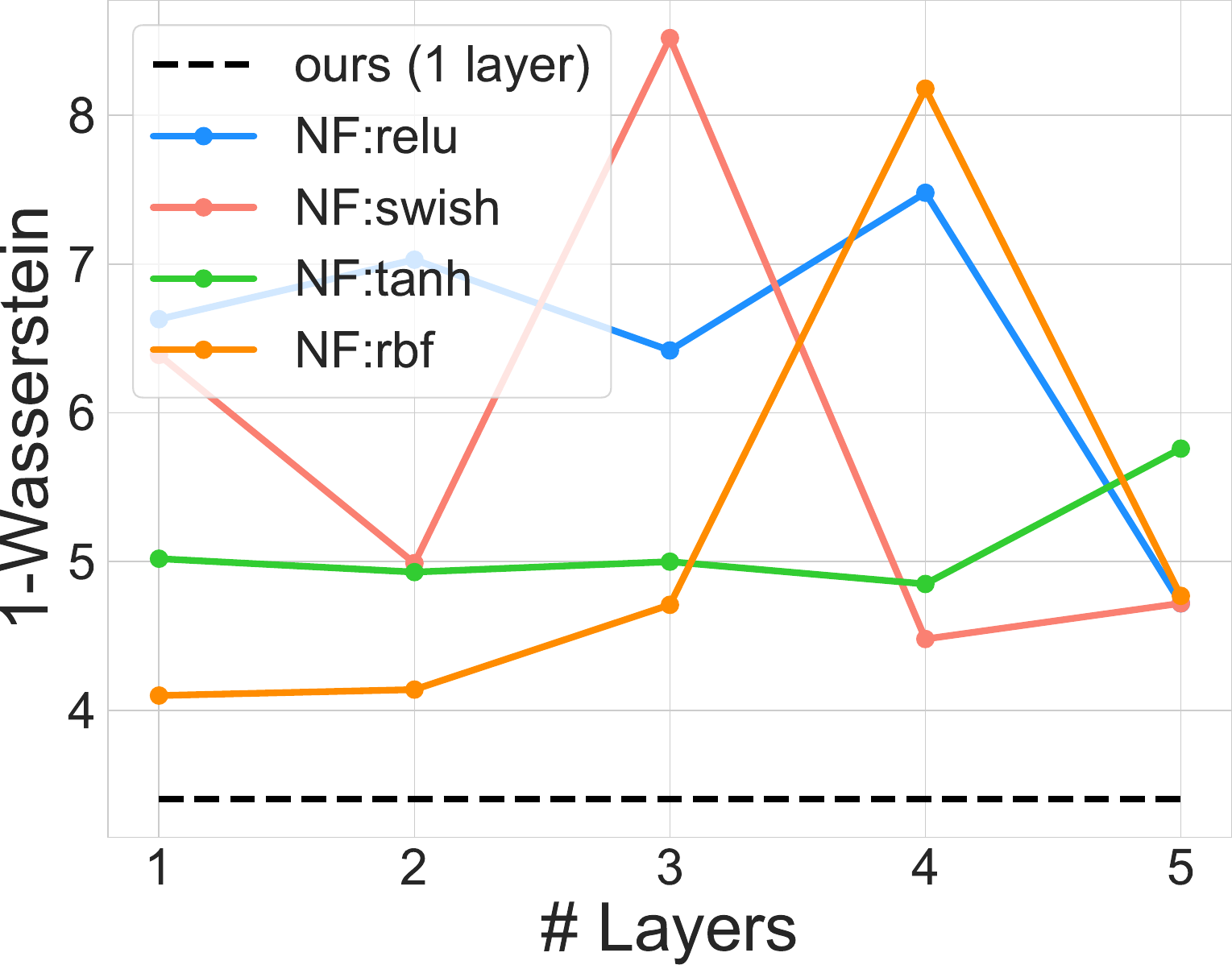} 
\includegraphics[width=0.245\textwidth]{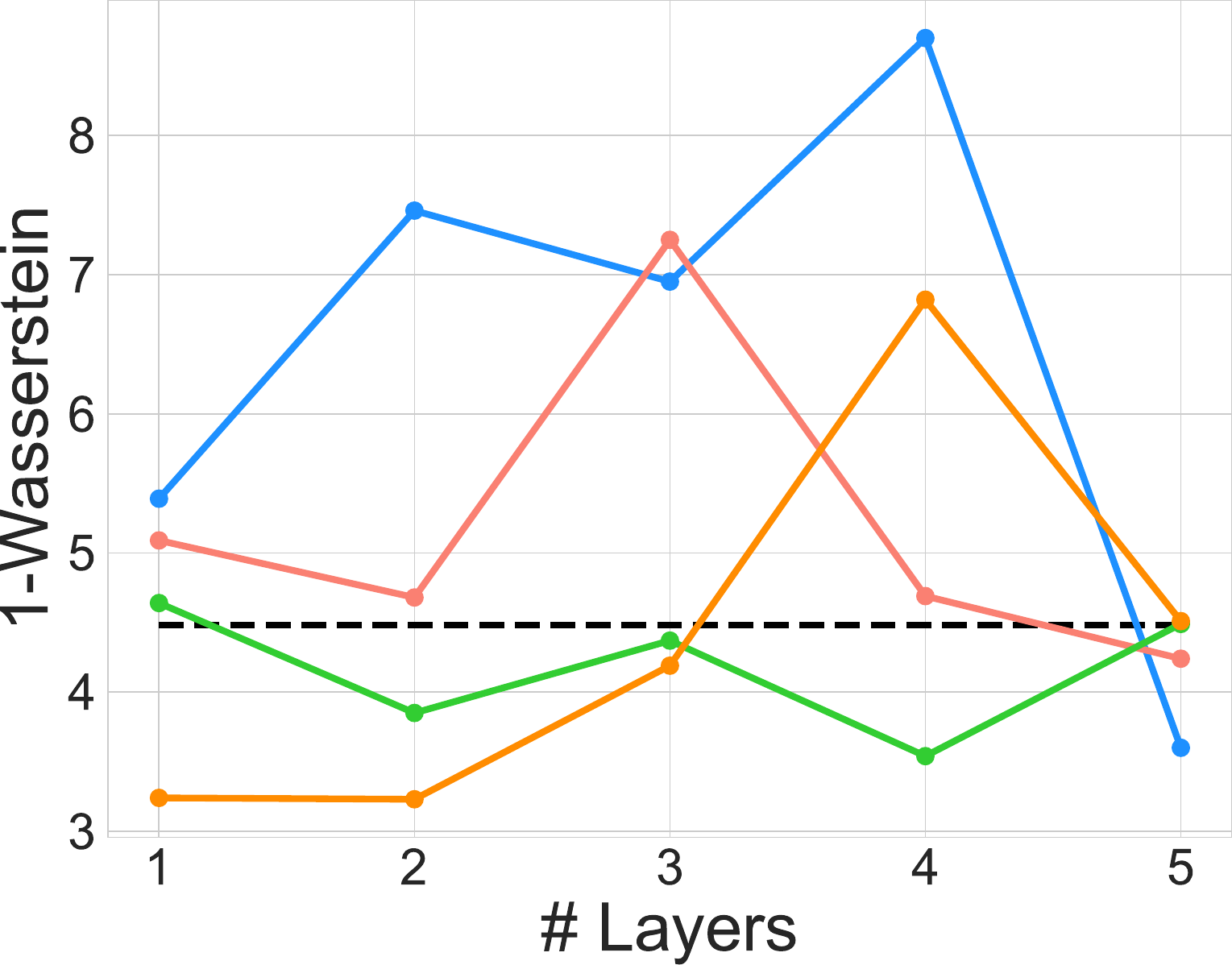} 
\includegraphics[width=0.245\textwidth]{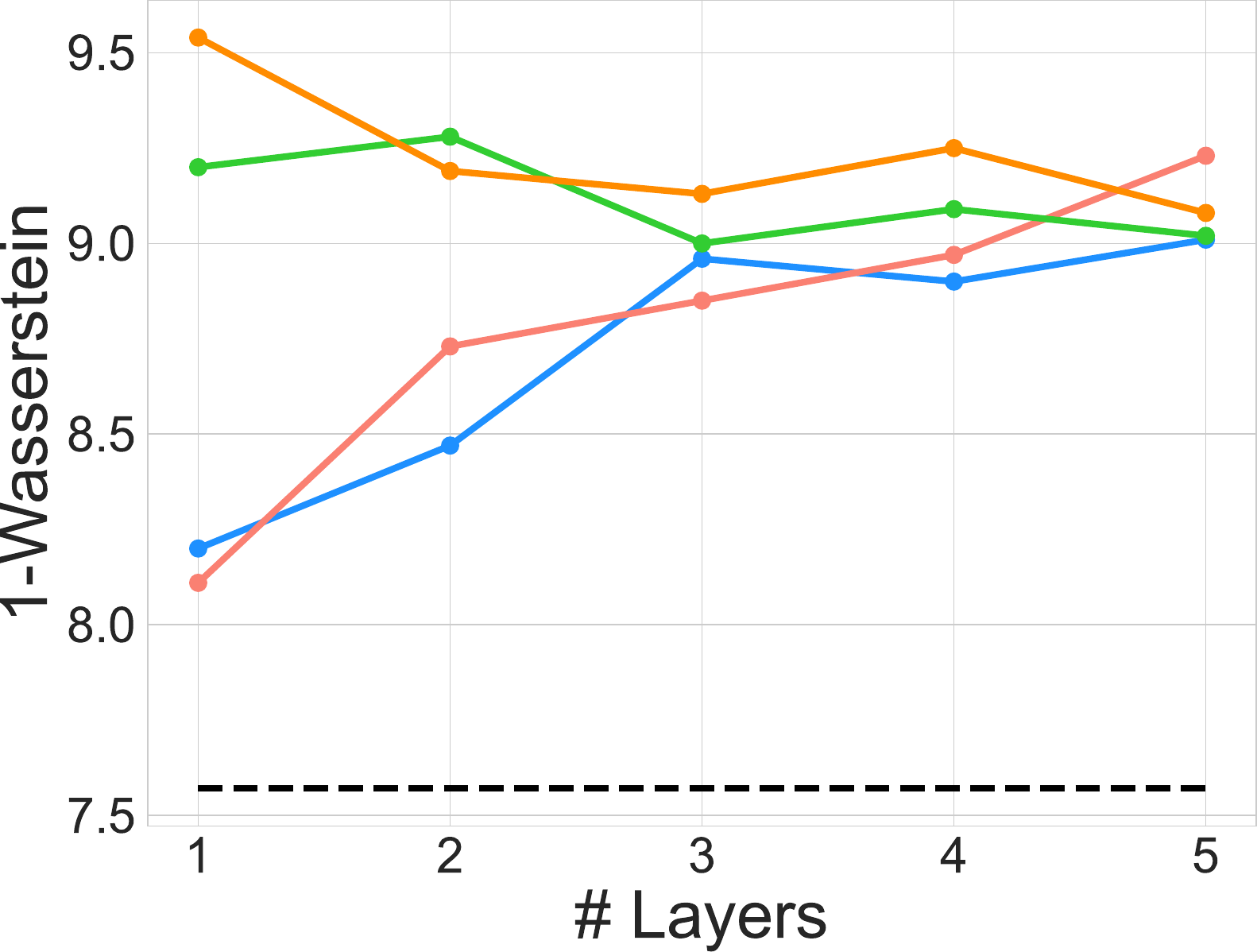}
\includegraphics[width=0.245\textwidth]{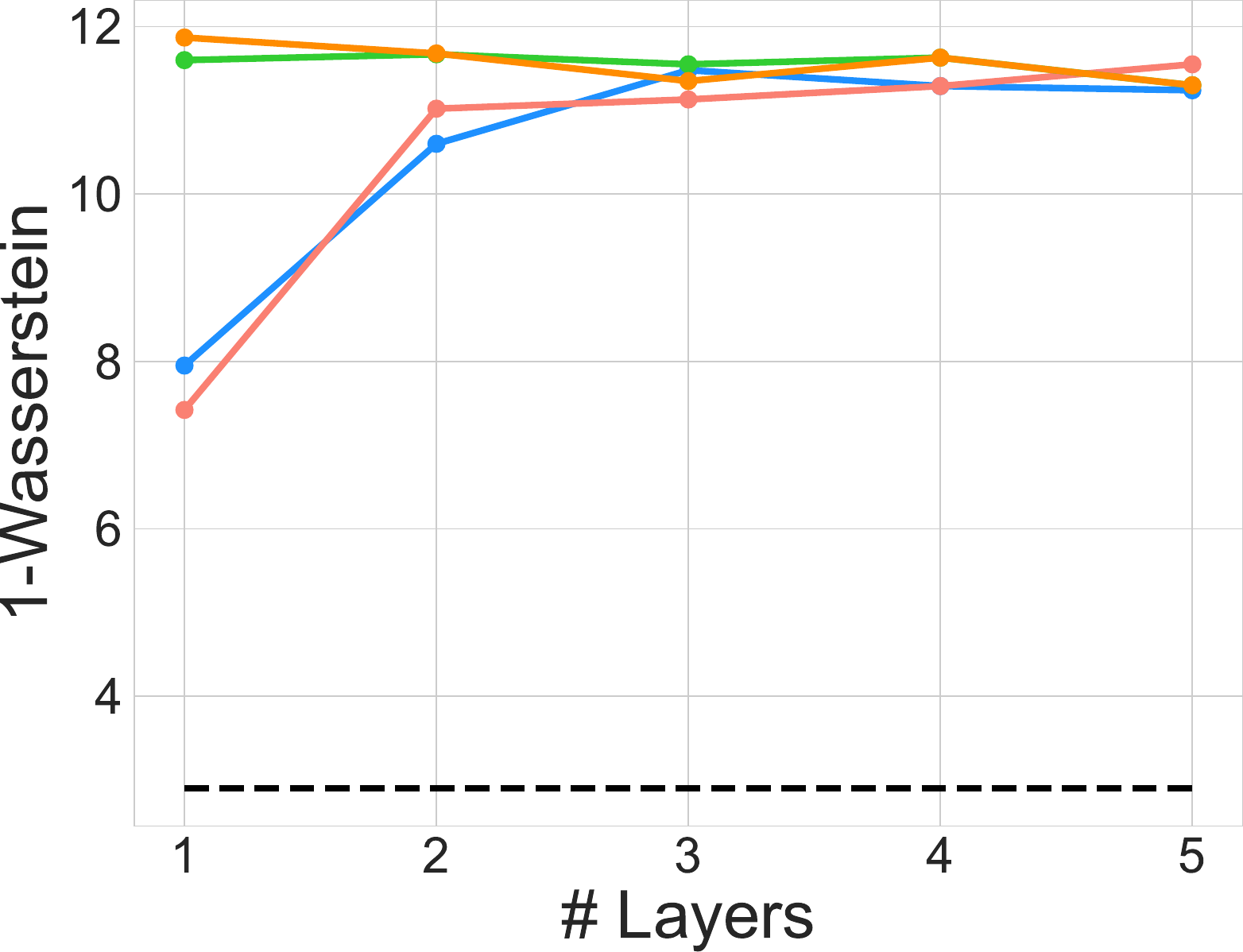}\\
\includegraphics[width=0.245\textwidth]{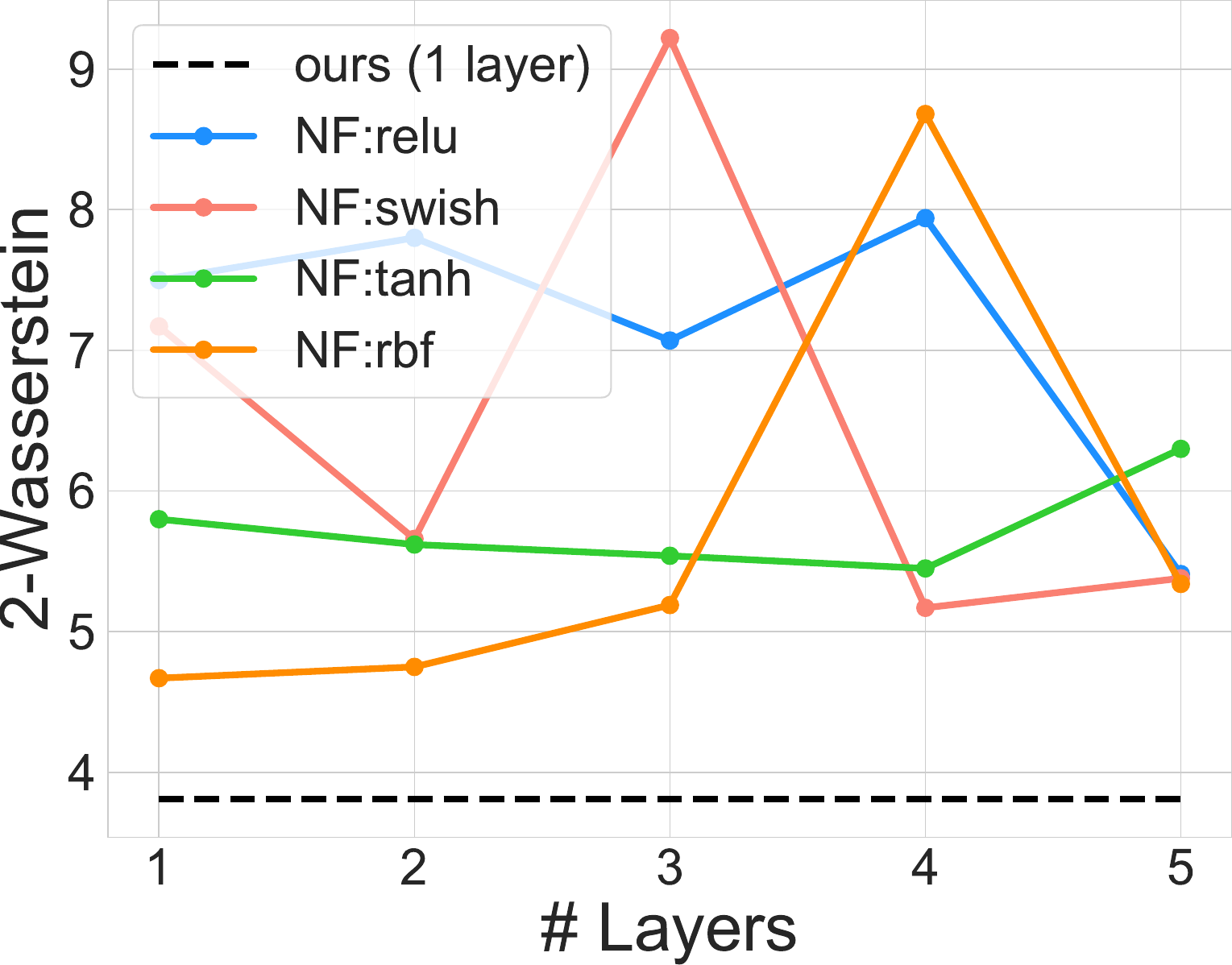} 
\includegraphics[width=0.245\textwidth]{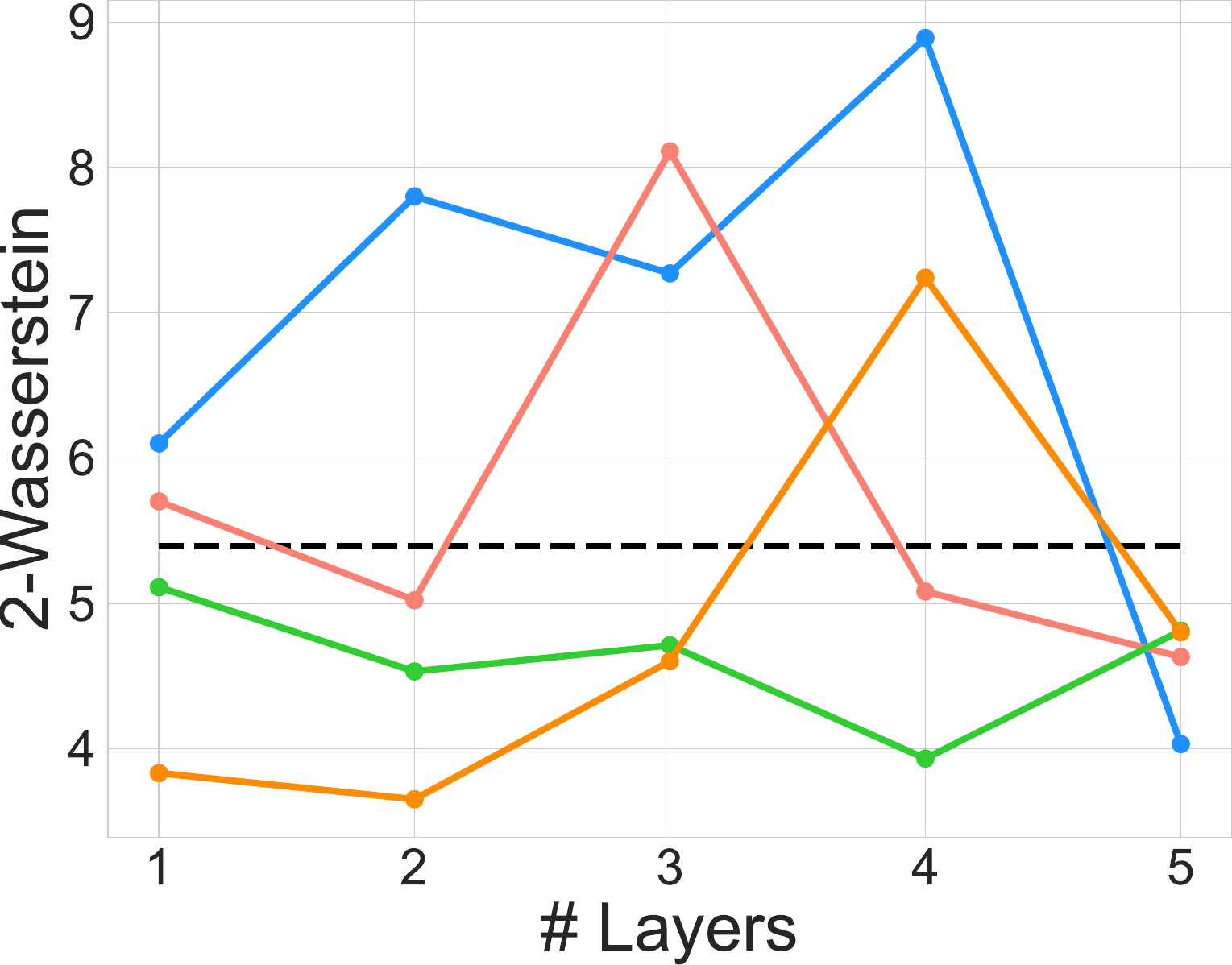} 
\includegraphics[width=0.245\textwidth]{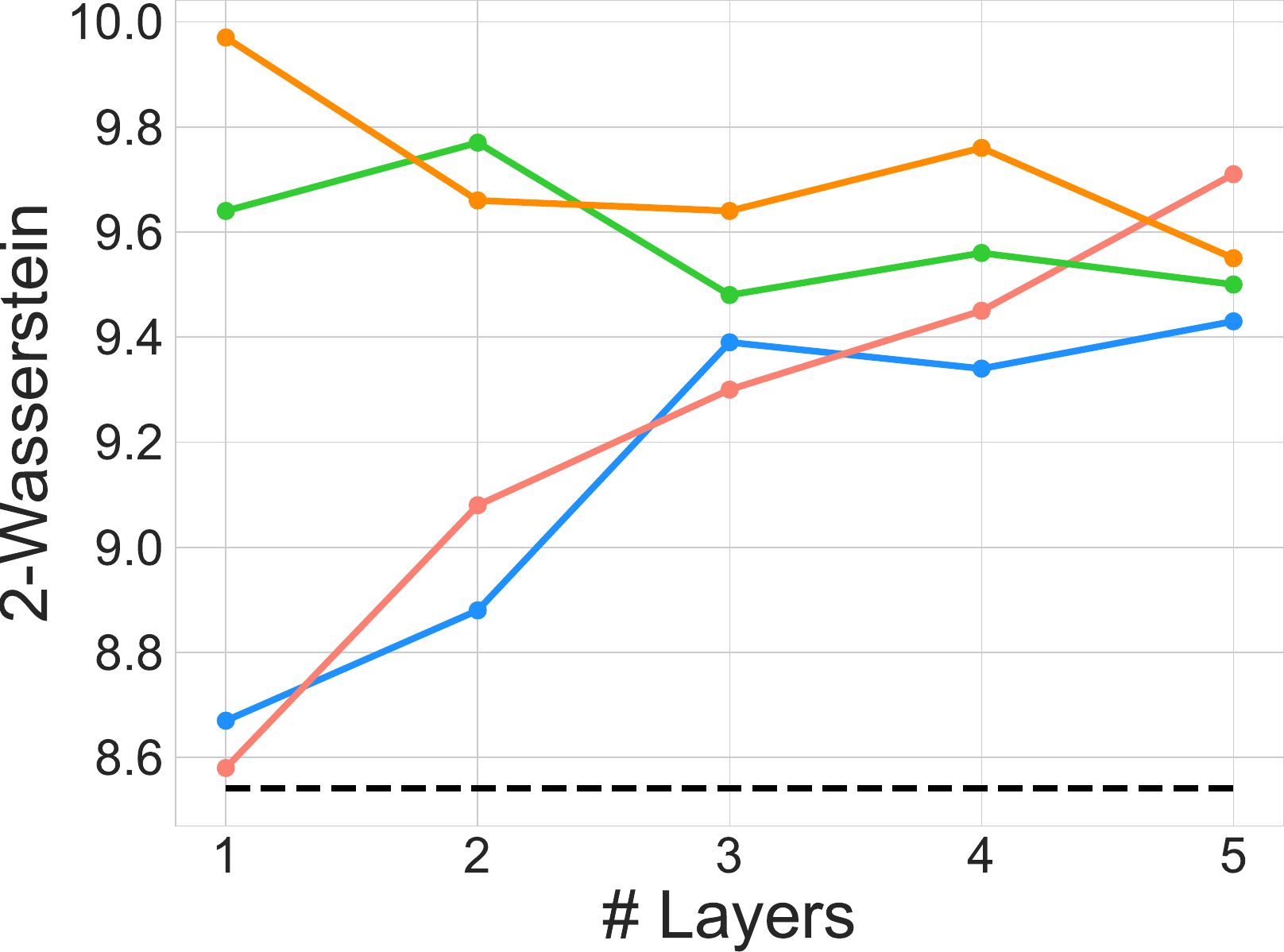} 
\includegraphics[width=0.245\textwidth]{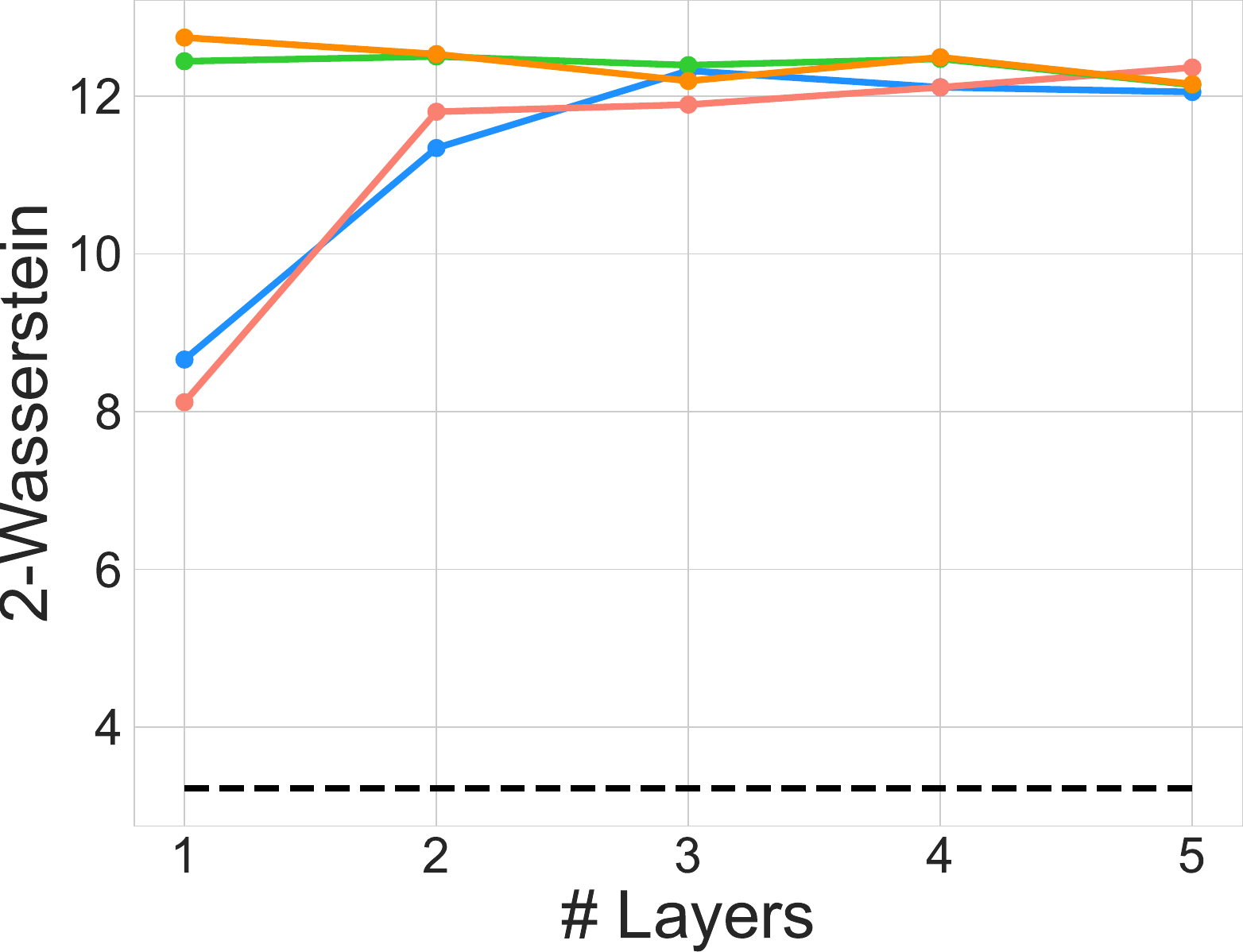}\\
\caption{
Comparison of \cite{tran2022all} method with a Normalizing Flow prior and different number of hidden layers. In this plot, we present $\mathcal{W}_1^1$ (\textbf{top row}) and $\mathcal{W}_2^2$ (\textbf{bottom row}) for the regions \textbf{R1}, \textbf{R2}, \textbf{R4}, and \textbf{R5} from left to right side. The appropriate regions might be found in Fig.~\ref{fig:evaluation_ranges}.}
\label{fig:ood_eval_nf_2}
\end{figure}

The results, presented in Fig.~\ref{fig:ood_eval_gaussian}, \ref{fig:ood_eval_gaussian_2}, \ref{fig:ood_eval_nf}, and \ref{fig:ood_eval_nf_2}, lead to several key observations regarding the performance on OOD tasks:
\begin{itemize}
    \item \textbf{Gaussian prior:} A greater number of hidden layers generally improves performance when using appropriate activation functions (\textit{TanH, RBF}) but degrades it for others (\textit{ReLU, Swish}).
    \item \textbf{Normalizing Flow prior:} In many cases, the best results are achieved with single- or two-hidden-layer BNNs, while deeper architectures significantly degrade performance.
    \item \textbf{Sensitivity to depth:} The Normalizing Flow prior is highly sensitive to the number of hidden layers, providing clear evidence of the heuristic's instability.
    \item \textbf{Our method:} Due to its grounding in theoretical properties, our approach consistently outperforms all configurations of the~\citep{tran2022all} method.
\end{itemize}

These findings highlight the limitations of heuristic-based approaches and reinforce the importance of theoretical guarantees in achieving robust generalization.

\end{document}